%% file: main.tex
\PassOptionsToPackage{table}{xcolor}
\documentclass[10pt]{article} 
\usepackage[preprint]{tmlr}

\input{math_commands.tex}

 \usepackage{graphicx}
\usepackage{url}
\usepackage{xcolor}
\usepackage[utf8]{inputenc}
\usepackage[TS1,T1]{fontenc}
\usepackage{array, booktabs, amssymb}
\usepackage{graphicx}
\usepackage{xcolor}
\usepackage{colortbl}
\usepackage{caption}
\usepackage{mathtools}
\definecolor{baseD}{HTML}{7CAFC2}

\usepackage{longtable}
\usepackage{ragged2e}
\usepackage{makecell}

\definecolor{lightblue}{rgb}{0.68, 0.85, 0.9} 
\definecolor{lightorange}{rgb}{1.0, 0.8, 0.6}
\definecolor{refblue}{HTML}{215F8A}

\usepackage{hyperref}
\hypersetup{
  colorlinks=true,
  linkcolor=black,
  citecolor=refblue,
  urlcolor=refblue,
}

\title{Discrete Diffusion Models: A Unified Framework from Tokenization to Generation}

\author{%
Ye Yuan\textsuperscript{1, 2}\thanks{Corresponding to: \href{mailto:ye.yuan3@mail.mcgill.ca}{ye.yuan3@mail.mcgill.ca}.}~, 
  Weien Li\textsuperscript{1},
  Rui Song\textsuperscript{1},
  Zeyu Li\textsuperscript{1},
  Haochen Liu\textsuperscript{3},\\
  Xiangyu Kong\textsuperscript{1, 2},
  Zixuan Dong\textsuperscript{4},
  Linfeng Du\textsuperscript{1, 2},
  Zipeng Sun\textsuperscript{1, 2},
  Weixu Zhang\textsuperscript{1, 2},\\
  Jiaxin Huang\textsuperscript{5},
  Changjiang Han\textsuperscript{5},
  Yonghan Yang\textsuperscript{5},
  Zichen Zhao\textsuperscript{5},
  Xiuyuan Hu\textsuperscript{6},\\
  Haolun Wu\textsuperscript{1, 2},
  Yankai Chen\textsuperscript{5},
  Fengran Mo\textsuperscript{7},
  Jikun Kang\textsuperscript{8},
  Bowei He\textsuperscript{5},\\
  Dawn Song\textsuperscript{9},
  Philip S. Yu\textsuperscript{10},
  Xue Liu\textsuperscript{5, 1, 2}\\
  \\
  \textsuperscript{1}McGill University, 
  \textsuperscript{2}Mila - Quebec AI Institute,
  \textsuperscript{3}University of Cambridge,\\
  \textsuperscript{4}University of Toronto,
  \textsuperscript{5}MBZUAI - Mohamed bin Zayed University of Artificial Intelligence,\\
  \textsuperscript{6}Tsinghua University,
  \textsuperscript{7}Rochester Institute of Technology ,
  \textsuperscript{8}Salesforce,\\
 \textsuperscript{9}University of California, Berkeley,
 \textsuperscript{10}University of Illinois Chicago
}

\def\month{MM}  
\def\year{YYYY} 
\def\openreview{\url{https://openreview.net/forum?id=XXXX}} 

\begin{document}

\maketitle

\begin{abstract}
\input{sections/00-abstract}
\end{abstract}


\input{sections/01-introduction}
\input{sections/02-related-work}
\input{sections/03-background}
\input{sections/04-tokenization}
\input{sections/05-formulation}
\input{sections/06-training}
\input{sections/07-inference}
\input{sections/08-scaling}
\input{sections/09-applications}
\input{sections/10-evaluation}
\input{sections/11-discussion}
\input{sections/12-future-directions}
\input{sections/13-conclusion}
\input{sections/14-broader-impact}

\newpage
\bibliography{main}
\bibliographystyle{tmlr}


\end{document}

%% file: math_commands.tex
\usepackage{amsmath,amsfonts,bm}

\def\eqref#1{equation~\ref{#1}}

\def\1{\bm{1}}

\DeclareMathAlphabet{\mathsfit}{\encodingdefault}{\sfdefault}{m}{sl}
\SetMathAlphabet{\mathsfit}{bold}{\encodingdefault}{\sfdefault}{bx}{n}



%% file: sections/00-abstract.tex
\noindent
Discrete denoising diffusion models (DDMs) have recently emerged as a compelling alternative to autoregressive (AR) modeling for discrete data, offering parallel generation and iterative global refinement capabilities. 
Unlike continuous diffusion, where the state space is fixed, DDMs are fundamentally shaped by how the discrete state space is constructed: the tokenization scheme, the vocabulary topology, and domain-specific structural alphabets. 
This work introduces a unified conceptual framework that views discrete diffusion models through the construction of the underlying discrete state space.
Within this framework, existing formulations, including transition-matrix, masking/absorbing-state, and score/ratio-based approaches, emerge as different instantiations of a common design space.
The framework further exposes common design trade-offs across training objectives, inference algorithms, scaling behavior, systems optimization, and evaluation protocols, suggesting several promising directions for future research.

%% file: sections/01-introduction.tex
\section{Introduction}

%
Autoregressive (AR) models have become the standard for generating discrete sequences.
By factoring the joint distribution into a product of left-to-right conditionals, AR models enjoy a clean maximum-likelihood objective, stable training, and mature decoding infrastructure~\citep{ghazvininejad2019maskpredict}, properties that have powered the scaling of large language models (LLMs) from early billion-parameter systems to frontier models with hundreds of billions of parameters.
Yet the AR paradigm carries fundamental limitations that grow more pressing as models are deployed at scale.
First, generation is inherently \emph{sequential}: each token must be produced before the next can begin, imposing $\mathcal{O}(L)$ serial decoding steps regardless of available parallelism, a bottleneck that becomes increasingly costly as sequence lengths grow into the tens or hundreds of thousands~\citep{unknown2025survparatext}.
Second, AR decoding is \emph{inherently left-to-right and irrevocable}: once a token is emitted it cannot be revised in light of later context unless an explicit post-hoc correction mechanism is applied.
This single-pass commitment limits the model's ability to perform global planning, satisfy non-local constraints, or refine earlier decisions after observing downstream structure.
Such capabilities are critical for tasks such as infilling, constrained editing, controllable generation, and long-horizon reasoning~\citep{ye2025beyondautore, ye2024diffusion}.

Diffusion models offer a fundamentally different generation paradigm: instead of building a sequence token by token, they start from a fully corrupted input and iteratively \emph{refine all positions simultaneously} over multiple denoising steps, with each step accessing the entire current state and thus enabling bidirectional context and global revision at every stage of generation.
This iterative-refinement view is attractive for several reasons.
First, parallel updates amortize wall-clock latency across positions, making generation time largely independent of sequence length given sufficient compute~\citep{sahoo2024simpleeffect, sahoo2025scalingup, nie2025largelanguag}.
Second, global context at every step allows the model to plan holistically, correcting early mistakes, coordinating long-range dependencies, and satisfying constraints that span the entire output~\citep{ye2025beyondautore, ye2024diffusion, deng2025thinkwhile}.
Third, the corruption-denoising framework also provides a natural interface for \emph{editing and infilling}: by selectively re-corrupting only a subset of positions, the model can revise or complete partial outputs while conditioning on the surrounding context, without architectural changes or task-specific fine-tuning~\citep{lee2025editcontcoar, nie2024dicediscrete, unknown2025flextextinfi}.
These properties make diffusion a compelling complement for AR generation whenever global coherence, parallel decoding, or fine-grained controllability is required.

Applying diffusion to discrete data, however, is not a straightforward extension of continuous diffusion.
In continuous spaces like images and waveforms, the forward process adds Gaussian noise and the reverse process predicts a score or denoised signal in the same $\mathbb{R}^d$ space.
Discrete sequences instead live in a categorical space $\{1,\dots,K\}^L$ where Euclidean notions of "small perturbations" and gradient-based score functions do not directly apply.
One might consider embedding discrete tokens into a continuous space, applying standard Gaussian diffusion, and rounding back to the nearest token.
However, such embed-then-diffuse approaches introduce a geometry mismatch between the embedding manifold and the discrete vocabulary, often resulting in rounding errors, representation collapse, and poor sample quality~\citep{li2022diffusionlm, gong2023diffuseq, he2023diffusionber}.
The alternative, which is the focus of this paper, is to define diffusion \emph{natively in discrete space}, which requires designing a \emph{discrete corruption process} together with a matching \emph{reverse denoiser} that predicts clean tokens from corrupted categorical inputs~\citep{austin2021structured, hoogeboom2021argmaxflows, sahoo2024simpleeffect, lou2024discrete}.
The choice of corruption process is not merely an implementation detail: it determines the topology of perturbations in token space, the difficulty of the denoising task at each timestep, and the resulting generation order.
The discrete setting further introduces unique challenges related to exact likelihood computation, noise-schedule design for categorical variables, and the interaction between vocabulary structure and forward-process semantics~\citep{shi2024likelihoodba, shi2024simplified, campbell2024discrete}.
These design choices constitute a rich and rapidly expanding design space that distinguishes discrete diffusion from both its continuous counterpart and from standard AR modeling.

The central premise of this paper is that this design space is best understood through a single organizing lens: the construction of the discrete state space, which we broadly refer to as \emph{tokenization}.
We argue that tokenization is not a preprocessing detail but a first-class design axis that shapes virtually every subsequent modeling decision.
In text, the choice of subword vocabulary determines the granularity of masking, the semantics of substitution noise, and the effective sequence length seen by the denoiser.
In tokenized multimodal generation, the codebook topology of a VQ-VAE or residual quantizer defines a notion of "distance" between tokens that the corruption process can either exploit or ignore.
In scientific domains, the natural alphabet, such as amino acids, nucleotides, atom types, bond types, already carries rich structural and chemical semantics that interact non-trivially with noise schedules and validity constraints.
By foregrounding tokenization, we unify these seemingly disparate domains under a single lens: the interaction between \emph{state-space structure} and \emph{corruption-denoising dynamics}.
This perspective reveals recurring design patterns and recurring failure modes that are largely invisible when tokenization is treated as a fixed upstream choice.

Building on this premise, this paper develops a unified framework for discrete diffusion that spans the full pipeline from tokenization to generation, and instantiates it across three broad application families.
The first domain is \emph{text and code generation via diffusion language models} (dLLMs): masked, absorbing-state, and flow-matching approaches that have recently been scaled to billion-parameter regimes for open-ended generation, instruction following, reasoning, and code synthesis~\citep{nie2025largelanguag, sahoo2025scalingup, ye2025scalingdiffu, dream2025dream7bdiffu, gong2026diffucoder}.
The second is \emph{tokenized multimodal generation}, where continuous signals, such as images, audio, and video, are first discretized using vector quantization or codec models and then modeled with discrete diffusion in the resulting token space, often jointly with text tokens in unified architectures~\citep{gu2022vectorquanti, chang2022maskgitmaske, xie2024showoone, kou2025mmadamultimo}.
The third encompasses \emph{domain-intrinsic discrete structures in science and engineering}: protein sequences over amino-acid alphabets, DNA/RNA over nucleotide vocabularies, molecular graphs with categorical node and edge attributes, and combinatorial objects such as graph layouts and scheduling solutions~\citep{gruver2023proteindesig, avdeyev2023dirichlet, vignac2023digressdiscr, sun2023difuscograph}.
We additionally cover an emerging direction, involving \emph{agents, planning, and tool use}, where discrete diffusion serves as a planner or structured-output generator for decision-making pipelines, including vision-language-action models and combinatorial optimization~\citep{ye2025beyondautore, deng2025thinkwhile, kaplan2025implicit}.

\begin{figure}[t]
  \centering
  \includegraphics[width=0.97\linewidth]{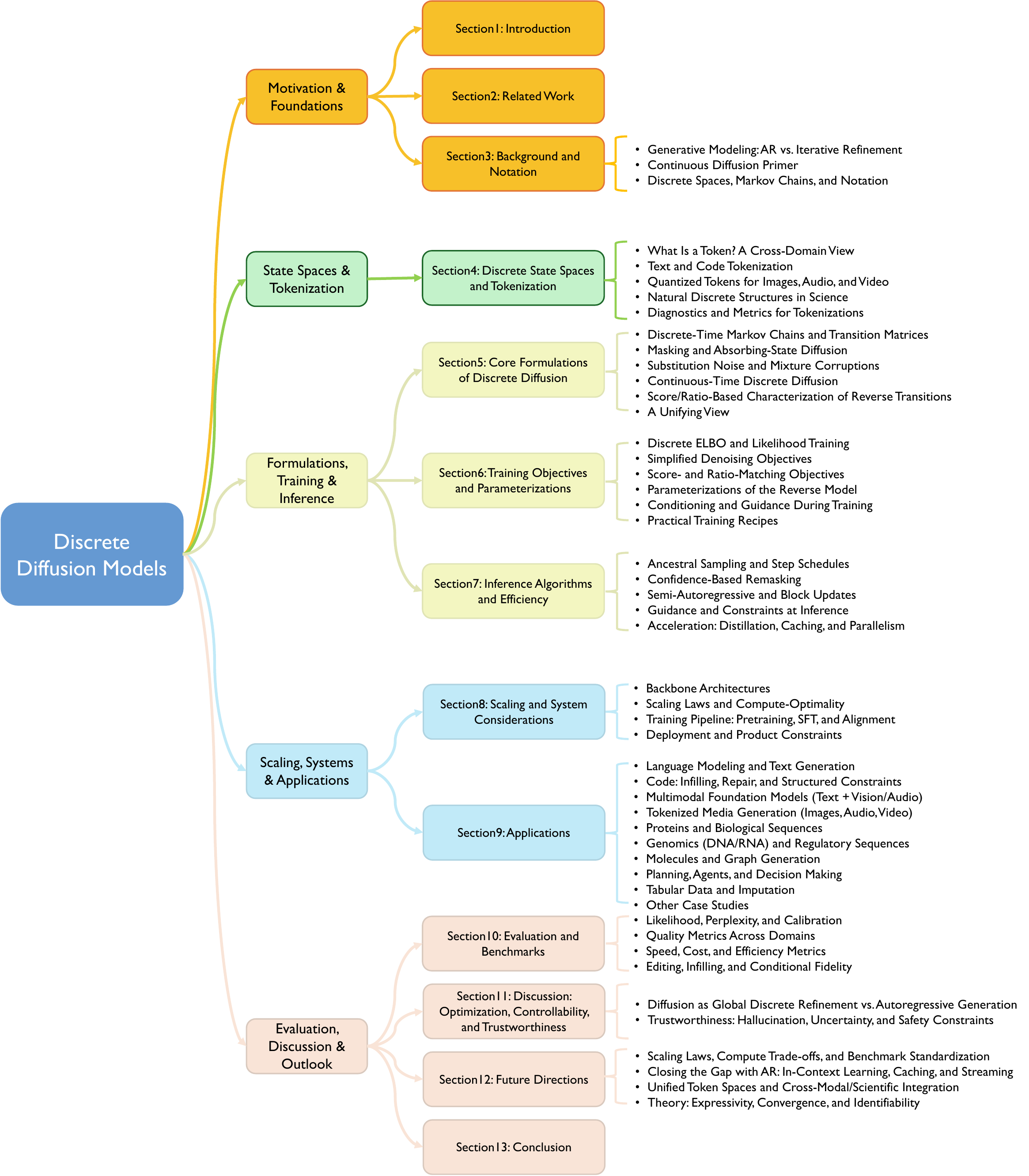}
  \caption{Paper overview.}
  \label{fig:overview-taxonomy}
\end{figure}

%
Our contributions are as follows.
\begin{enumerate}
    \item \textbf{Tokenization-centric lens.}
    We elevate discrete state-space construction from a peripheral implementation choice to a primary organizing principle, analyzing how vocabulary design, codebook topology, and natural alphabets shape corruption processes, denoising difficulty, and downstream controllability, while introducing diffusion-facing diagnostics for tokenization quality (Section~\ref{sec:tokenization}).

    \item \textbf{Unified formulation and training/inference framework.}
    We decompose every discrete diffusion model into four components --- the corruption operator, denoiser parameterization, training objective, and sampler --- and show how the major formulation families instantiate this shared structure (Section~\ref{sec:formulations}), then organize training objectives and parameterizations (Section~\ref{sec:training}) and inference algorithms (Section~\ref{sec:inference}) around the same decomposition.

    \item \textbf{Cross-domain instantiation.}
    We map the framework across text and code, tokenized multimodal generation, proteins, genomics, molecules, and graphs, as well as planning and agents, surfacing shared design patterns and domain-specific constraints (Section~\ref{sec:applications}).

    \item \textbf{Scaling, systems, and evaluation.}
    We treat scaling behavior, training pipelines, inference acceleration, and evaluation protocols as integral parts of the design space rather than afterthoughts (Sections~\ref{sec:scaling} and~\ref{sec:evaluation}).

    \item \textbf{Open problems.}
    We identify concrete open challenges: scaling behavior, the in-context-learning gap, KV-cache analogues, streaming generation, unified token spaces, and theoretical foundations, and frame these challenges as actionable research questions (Section~\ref{sec:future}).
\end{enumerate}

The remainder of the paper is organized as follows.
Section~\ref{sec:related} positions our framework relative to prior overviews of diffusion-based generation.
Section~\ref{sec:background} establishes notation and provides minimal background on AR generation, continuous diffusion, and discrete Markov chains.
Section~\ref{sec:tokenization} examines discrete state spaces and tokenization across text, code, quantized media, and scientific domains.
Section~\ref{sec:formulations} presents the core formulations: transition-matrix models, absorbing-state masking, substitution and mixture corruptions, continuous-time processes, score/ratio-based methods, and a unifying view.
Section~\ref{sec:training} systematizes training objectives, reverse parameterizations, conditioning and guidance, and practical training recipes.
Section~\ref{sec:inference} covers inference algorithms: ancestral sampling, confidence-based remasking, semi-autoregressive and block updates, guidance and constraints, and acceleration techniques.
Section~\ref{sec:scaling} discusses scaling and systems considerations.
Section~\ref{sec:applications} instantiates the framework across application domains.
Section~\ref{sec:evaluation} reviews evaluation and benchmarks.
Section~\ref{sec:discussion} discusses discrete diffusion as global refinement, connections to optimization, trustworthiness, and when AR or hybrid approaches are preferable.
Section~\ref{sec:future} outlines future directions, and Section~\ref{sec:conclusion} concludes.
We release a companion repository containing taxonomies and curated resources at \href{https://github.com/AAAAA-Academia-Attractions/Discrete-Diffusion}{https://github.com/AAAAA-Academia-Attractions/Discrete-Diffusion}.

%% file: sections/02-related-work.tex
\section{Related Work} \label{sec:related}
A number of recent overviews touch on diffusion-based generation. Here we position our framework relative to them and make explicit what a tokenization-centric, cross-domain treatment adds.


%
%
Recent surveys have substantially improved coverage of diffusion-based generation, especially in text and multimodal language modeling.
Text-centered reviews have organized the literature around non-autoregressive generation, continuous-versus-discrete formulations, or the emergence of diffusion language models and diffusion-based large language models~\citep{unknown2023diffmodenona, unknown2025overdiffmode, unknown2025survdiffmode, unknown2025diffmodetext, unknown2025survdifflang, unknown2025difflarglang, unknown2025discdifflarg, unknown2025top10open}.
Other surveys emphasize efficiency, including caching, parallel decoding, and deployment bottlenecks~\citep{unknown2025effidifflang, unknown2025survcachmeth, unknown2025survparatext}.
Application-specific reviews in graphs, molecules, biomolecules, drug design, and recommender systems document the expansion of diffusion methods into structured scientific and industrial domains~\citep{unknown2023genediffmode, yang2023graphdiffusi, unknown2025diffmodemole, unknown2025siftthronois, unknown2025survgeneai, unknown2025survdiffmodea}.
Broad diffusion surveys provide DDPM/SDE foundations and a field-wide taxonomy~\citep{yang2023diffusion, unknown2025survgenediff}.
These works collectively provide strong coverage of model families, applications, and recent progress.
However, several issues remain under-emphasized for a survey centered on \emph{discrete diffusion}.
Most notably, prior surveys rarely treat \emph{discrete state spaces and tokenization} as a first-class organizing axis.
Even when tokenization is discussed, it is usually framed as a local modeling choice, a future challenge, or a subtopic within a broader taxonomy rather than as a central principle that shapes corruption design, reverse parameterization, controllability, and validity~\citep{unknown2025discdifflarg, unknown2025top10open}.
A second under-emphasized aspect is the lack of a unified treatment of \emph{domain-intrinsic discrete structures}.
Language tokens, code symbols, quantized multimodal tokens, biological alphabets, molecular components, and graph primitives are usually discussed within separate application areas rather than within a shared design space~\citep{unknown2025diffmodemole, unknown2023genediffmode, yang2023graphdiffusi, unknown2025siftthronois, unknown2025survdiffmodea}.
A third gap is that formulations, objectives, inference, systems considerations, and evaluation are often reviewed separately, even though in practice they are coupled through the underlying discrete representation.
As a result, the field still lacks a survey that treats discrete diffusion not simply as a collection of models, but as an end-to-end design space.

This paper is organized around precisely these missing links.
Rather than centering only model families, we take \emph{tokenization and discrete state-space diagnostics} as a primary organizing principle.
This reflects our main thesis: in discrete diffusion, many of the most consequential choices arise before and beyond the denoiser architecture itself.
Accordingly, this paper differs from prior work in three main ways.
First, it treats tokenization, state-space factorization, and representation diagnostics as core methodological questions rather than peripheral implementation details.
Second, it builds a cross-domain map spanning language, tokenized multimodal generation, and scientific discrete structures such as proteins, genomics, molecules, and graphs~\citep{unknown2025discdifflarg, unknown2025difflarglang, unknown2025diffmodemole, unknown2023genediffmode, yang2023graphdiffusi, unknown2025siftthronois}.
Third, it foregrounds scaling, systems, and evaluation standardization in the same framework, bringing together deployment constraints, parallelism, caching, efficiency metrics, controllability, calibration, and structural validity rather than leaving them fragmented across separate surveys~\citep{unknown2025effidifflang, unknown2025survcachmeth, unknown2025survparatext}.
Table~\ref{tab:survey_summary} compresses this comparison into a coverage checklist: each column is an organizing axis of this paper, each row is a class of prior surveys, and the bottom row is the present survey.

Beyond taxonomy, we aim to make the paper directly reusable as a practical design reference.
To this end, we distill cross-domain checklists for choosing tokenization schemes, corruption operators, training objectives, reverse parameterizations, sampling procedures, constraints, and evaluation protocols.
These checklists are motivated by recurring questions that appear across domains.
What discrete state best captures the underlying structure.
Which corruption mechanism preserves enough semantics or validity to support effective denoising.
Which objective best matches the intended reverse process.
When iterative remasking, block updates, or semi-autoregressive decoding are preferable.
How hard and soft constraints should be imposed.
And which metrics meaningfully capture not only quality, but also efficiency, controllability, calibration, and structural correctness.
By making these decisions explicit and comparable across domains, this paper is intended not only to summarize the literature, but also to provide a reusable framework for designing, analyzing, and evaluating discrete diffusion systems in a principled way.

\paragraph{Scope and selection.} 
This is a narrative survey. 
We assembled the literature by tracking the citation graph around the foundational discrete-diffusion formulations (D3PM, multinomial diffusion, MDLM/MD4, SEDD, discrete flow matching) and the major venues and preprint servers through 2021 to early 2026, prioritizing works that (i) introduce a distinct corruption/parameterization/objective/sampler choice, (ii) scale discrete diffusion to a new modality or size regime, or (iii) report a reusable evaluation or systems technique. 
We do not claim exhaustive coverage of every application paper. The marks in Table~\ref{tab:survey_summary} reflect our qualitative reading of each survey class relative to this paper's tokenization-centric scope rather than a quantitative coding rubric.
The most closely related prior survey is \citet{unknown2025discdifflarg}, grouped with the text and DLM surveys in Table~\ref{tab:survey_summary}: relative to it, we (a) make discrete state-space construction (tokenization) the primary organizing axis rather than one topic among many, (b) add the domain-intrinsic scientific alphabets (proteins, genomics, molecules/graphs) and the planning/agent and tabular settings under one shared four-component framework, and (c) fold scaling, systems, and evaluation standardization into the same framework; conversely, that survey offers complementary depth on multimodal quantization that we treat more briefly.

\input{tables/01-related-work}

%% file: tables/01-related-work.tex
\begin{table}[t]
\centering
\caption{Prior surveys grouped by primary lens and compared along the organizing axes of this paper.
Each row is a class of related surveys rather than an individual paper.
$\checkmark$: treated as a central topic;
$\triangle$: discussed but not an organizing focus;
--: not a focus.
Only this paper covers all six axes.}
\label{tab:survey_summary}
\small
\setlength{\tabcolsep}{2.5pt}
\renewcommand{\arraystretch}{1.18}
\begin{tabular}{@{}
  >{\RaggedRight\arraybackslash}p{0.28\linewidth}
  *{6}{>{\centering\arraybackslash}p{\dimexpr(\linewidth-0.28\linewidth-12\tabcolsep)/6\relax}}
  @{}}
\toprule
\textbf{Survey class}
& \textbf{Tokenization as axis}
& \textbf{Unified formulation}
& \textbf{Text \& code}
& \textbf{Tokenized multimodal}
& \textbf{Scientific discrete}
& \textbf{Scaling, systems, eval.} \\
\midrule
Text and DLM surveys\newline
{\footnotesize\citep[e.g.][]{unknown2025discdifflarg, unknown2025survdifflang, unknown2025difflarglang}}
& $\triangle$ & $\checkmark$ & $\checkmark$ & $\triangle$ & -- & $\triangle$ \\
\addlinespace
Efficiency and systems surveys\newline
{\footnotesize\citep{unknown2025effidifflang, unknown2025survcachmeth, unknown2025survparatext}}
& -- & $\triangle$ & $\checkmark$ & $\triangle$ & -- & $\checkmark$ \\
\addlinespace
Domain-specific surveys\newline
{\footnotesize\citep[e.g.][]{yang2023graphdiffusi, unknown2025diffmodemole, unknown2025siftthronois}}
& -- & $\triangle$ & -- & -- & $\checkmark$ & $\triangle$ \\
\addlinespace
Broad diffusion surveys\newline
{\footnotesize\citep{yang2023diffusion, unknown2025survgenediff}}
& -- & $\triangle$ & $\triangle$ & $\triangle$ & $\triangle$ & -- \\
\midrule
\rowcolor{gray!12}
\textbf{This paper}
& $\checkmark$ & $\checkmark$ & $\checkmark$ & $\checkmark$ & $\checkmark$ & $\checkmark$ \\
\bottomrule
\end{tabular}
\end{table}

%% file: sections/03-background.tex
\section{Background and Notation} \label{sec:background}
\noindent
This section synthesizes two lines of prior work that underpin discrete diffusion: the probabilistic foundations of diffusion models~\citep{sohldickstein2015deepunsuperv, yang2023diffusion, yang2023graphdiffusi}, and earlier non-autoregressive and order-agnostic generation methods whose iterative-refinement view discrete diffusion inherits and formalizes~\citep{ghazvininejad2019maskpredict, uria2022training, hoogeboom2022autoregressi}.

\subsection{Generative Modeling: AR vs.\ Iterative Refinement}

\paragraph{Autoregressive factorization.}
Let $\bm{x} = (x_1, x_2, \dots, x_L)$ denote a sequence of discrete tokens, where each $x_i$ takes values in a finite vocabulary $\mathcal{V} = \{1, \dots, K\}$.
An autoregressive (AR) model decomposes the joint distribution via the chain rule:
\begin{equation}\label{eq:ar}
  p_\theta(\bm{x}) \;=\; \prod_{i=1}^{L} p_\theta(x_i \mid x_{<i}),
\end{equation}
where $x_{<i} = (x_1, \dots, x_{i-1})$ denotes the left context.
Each conditional $p_\theta(x_i \mid x_{<i})$ is typically parameterized by a causal (unidirectional) Transformer that outputs a categorical distribution over $\mathcal{V}$ at position~$i$.
Training maximizes the log-likelihood under \emph{teacher forcing}: at each step the model receives the ground-truth prefix $x_{<i}$ and is trained to predict $x_i$, yielding a simple per-token cross-entropy loss.

At inference time, tokens are sampled sequentially: $x_1 \sim p_\theta(x_1)$, then $x_2 \sim p_\theta(x_2 \mid x_1)$, and so on.
This sequential procedure requires $L$ serial forward passes through the model, each generating exactly one token.
While techniques such as KV caching (which avoids redundant recomputation of attention over the prefix) and speculative decoding (which drafts multiple tokens in parallel and verifies them against the target model) substantially reduce per-step cost and amortize latency, they do not eliminate the fundamental $\mathcal{O}(L)$ serial dependency: the $i$-th token cannot be sampled until the $(i{-}1)$-th token has been committed.
For long sequences, this sequential bottleneck becomes the dominant wall-clock cost, especially on modern accelerators where parallel compute is abundant but serial throughput is constrained~\citep{unknown2025survparatext}.

A second structural limitation is that AR decoding is \emph{left-to-right committed}: once $x_i$ is emitted, it becomes part of the conditioning context for all subsequent tokens and is never revisited.
The model cannot correct an early mistake in light of later evidence, satisfy constraints that span the full sequence, or revise global structure after local details have been fixed.
This one-pass commitment makes AR models a poor fit for tasks that require bidirectional context, such as infilling (generating text conditioned on both a prefix and a suffix), constrained editing (modifying a passage while preserving specific spans), and planning (producing a sequence whose global coherence depends on long-range coordination)~\citep{ye2025beyondautore, ye2024diffusion}.

\paragraph{Iterative refinement as an alternative.}
Non-autoregressive (NAR) and iterative-refinement approaches break the sequential dependency by generating or refining all positions simultaneously.
The idea dates back at least to iterative conditional modes in structured prediction and was popularized for neural sequence generation by Mask-Predict~\citep{ghazvininejad2019maskpredict}, which alternates between predicting all masked tokens in parallel and re-masking the least confident predictions.
This iterative loop (predict, evaluate confidence, re-corrupt, repeat) can be viewed as a fixed number of refinement steps $T \ll L$, where each step updates many positions at once.

Diffusion models generalize this idea by defining a principled probabilistic framework for the corruption and refinement process.
Rather than relying on a hand-designed re-masking heuristic, diffusion models specify a \emph{forward process} that gradually corrupts data over $T$ steps (or continuously in time) and learn a \emph{reverse process} that progressively denoises the corrupted input back to a clean sample.
The number of denoising steps $T$ is a free parameter that trades off sample quality against generation speed, and is typically much smaller than the sequence length ($T \ll L$).
At each reverse step, the model observes the \emph{entire} current state and updates them jointly, providing bidirectional context and the ability to revise earlier decisions.
This parallel, globally informed refinement is the key structural advantage of diffusion over AR generation, and it motivates the substantial body of work reviewed in this survey.

\subsection{Continuous Diffusion Primer}

\paragraph{Forward process, reverse denoising, and the score.}
To motivate discrete diffusion, we briefly recall the continuous-diffusion framework that has driven progress in image and audio generation~\citep{sohldickstein2015deepunsuperv, yang2023diffusion}.
Let $\bm{z}_0 \in \mathbb{R}^d$ be a data point.
A \emph{forward} (noising) process gradually perturbs $\bm{z}_0$ by adding Gaussian noise over a sequence of timesteps $t = 1, \dots, T$:
\begin{equation}\label{eq:cont_forward}
  q(\bm{z}_t \mid \bm{z}_{t-1}) \;=\; \mathcal{N}\!\bigl(\bm{z}_t;\, \sqrt{1 - \beta_t}\,\bm{z}_{t-1},\; \beta_t \bm{I}\bigr),
\end{equation}
where $\{\beta_t\}_{t=1}^T$ is a noise schedule controlling the rate of corruption.
This process admits a closed-form marginal $q(\bm{z}_t \mid \bm{z}_0) = \mathcal{N}(\bm{z}_t;\, \sqrt{\bar\alpha_t}\,\bm{z}_0,\, (1 - \bar\alpha_t)\bm{I})$ with $\bar\alpha_t = \prod_{s=1}^t (1 - \beta_s)$, so that for large $T$ the noisy sample $\bm{z}_T$ becomes approximately standard Gaussian.
A neural network $\bm{\epsilon}_\theta(\bm{z}_t, t)$ is trained to predict the noise (or, equivalently, the score $\nabla_{\bm{z}_t} \log q(\bm{z}_t)$) at each step, yielding a learned \emph{reverse process} that iteratively denoises $\bm{z}_T \sim \mathcal{N}(\bm{0}, \bm{I})$ back toward the data distribution.
The training objective, typically a reweighted mean-squared error between the predicted and true noise, can be derived as a simplified variational bound on $\log p_\theta(\bm{z}_0)$.

Two features of this framework are worth highlighting.
First, the forward process is \emph{fixed} and analytic: Gaussian noise in $\mathbb{R}^d$ has well-understood geometry, and the posterior $q(\bm{z}_{t-1} \mid \bm{z}_t, \bm{z}_0)$ is available in closed form.
Second, the score function $\nabla_{\bm{z}_t} \log q(\bm{z}_t)$ is a continuous vector field, and the denoiser operates via gradient-like updates in the same Euclidean space as the data.

\paragraph{Why "embed then diffuse" is problematic for discrete data.}
A natural attempt to apply continuous diffusion to discrete sequences is to embed tokens into a continuous space, run Gaussian diffusion, and then round back to the nearest token, an approach explored by several early works~\citep{li2022diffusionlm, gong2023diffuseq, he2023diffusionber, mahabadi2024tesstexttote, han2023ssdlmsemiaut}.
As discussed in the introduction, this introduces a geometry mismatch between the continuous embedding manifold and the discrete vocabulary, leading to rounding errors and embedding collapse.
The technical root of the problem in the present notation is that the continuous score $\nabla_{\bm{z}_t} \log q(\bm{z}_t)$ does not respect the discrete structure of the vocabulary: it can point in directions that interpolate between token embeddings rather than toward any valid token, so the gradient-based reverse update of Eq.~\ref{eq:cont_forward} has no faithful categorical analogue.

These observations motivate defining diffusion \emph{natively in discrete space}, where the forward process corrupts tokens by categorical operations (masking, substitution, or structured transitions) and the reverse process directly predicts categorical distributions over the vocabulary.
This is the approach taken by the discrete diffusion models that are the focus of this survey, and we introduce the necessary formalism next.

\subsection{Discrete Spaces, Markov Chains, and Notation}

\paragraph{State space.}
Throughout this paper, we consider data consisting of sequences of discrete tokens.
Let $K$ denote the vocabulary size and $\mathcal{V} = \{1, 2, \dots, K\}$ the vocabulary.
A sequence of length $L$ is written $\bm{x} = (x_1, x_2, \dots, x_L) \in \mathcal{V}^L$.
Each token $x_i$ can be represented as a one-hot vector $\bm{e}_{x_i} \in \{0,1\}^K$; we will use this representation whenever it simplifies notation.
In some domains the "sequence" has richer structure: for example, a molecular graph can be represented as a tuple of node-type and edge-type vectors, each taking values in a domain-specific categorical alphabet.
We generally write $\bm{x}$ for the full discrete object and $x_i$ for its $i$-th component, noting that the framework extends to non-sequential structures by indexing over nodes, edges, or other discrete components.
We use $K+1$ to refer to the extended vocabulary $\mathcal{V}_{\texttt{m}} = \{1, \dots, K, \texttt{m}\}$ when an absorbing (mask) token $\texttt{m}$ is introduced.
The data distribution is denoted $q_{\text{data}}(\bm{x})$.

\paragraph{Forward process: discrete corruption via transition matrices.}
A discrete diffusion model defines a \emph{forward process} that progressively corrupts a clean data point $\bm{x}_0 \sim q_{\text{data}}$ over $T$ timesteps, producing a sequence of increasingly noisy states $\bm{x}_1, \bm{x}_2, \dots, \bm{x}_T$.
The corruption at each step is specified by a categorical Markov chain.
For a single token $x_i$ (we drop the position subscript $i$ when the per-position structure is clear), the one-step transition is defined by a \emph{transition matrix} $\bm{Q}_t \in [0,1]^{K \times K}$ (or $\mathbb{R}^{(K+1) \times (K+1)}$ when a mask token is included):
\begin{equation}\label{eq:forward_onestep}
  q(x_t \mid x_{t-1}) \;=\; \mathrm{Cat}\!\bigl(x_t;\; \bm{e}_{x_{t-1}}^\top \bm{Q}_t\bigr),
\end{equation}
where $[\bm{Q}_t]_{jk} = q(x_t{=}k \mid x_{t-1}{=}j)$ gives the probability of transitioning from category $j$ to category $k$ at step~$t$, and each row of $\bm{Q}_t$ sums to one.
Because the chain is Markov, the marginal at any step $t$ given the clean token $x_0$ is obtained by composing transition matrices:
\begin{equation}\label{eq:forward_marginal}
  q(x_t \mid x_0) \;=\; \mathrm{Cat}\!\bigl(x_t;\; \bm{e}_{x_0}^\top \bar{\bm{Q}}_t\bigr), \qquad \bar{\bm{Q}}_t \;=\; \bm{Q}_1 \bm{Q}_2 \cdots \bm{Q}_t.
\end{equation}
The cumulative matrix $\bar{\bm{Q}}_t$ summarizes the total corruption from $x_0$ to $x_t$ and is the key object for training, since it allows direct sampling of $x_t$ given $x_0$ without simulating the chain step by step.
The schedule is designed so that $q(x_T \mid x_0)$ approaches a known stationary distribution, typically uniform over $\mathcal{V}$ or concentrated on $\texttt{m}$, that serves as the prior of the generative model.

The choice of $\bm{Q}_t$ encodes strong inductive biases about what "corruption" means in a given domain.
Three canonical designs recur: \emph{uniform substitution} (any token may be replaced by any other with equal probability), \emph{absorbing/masking} (tokens are progressively replaced by a special token $\texttt{m}$), and \emph{structured/embedding-aware} transitions (substitution probabilities reflect token similarity)~\citep{austin2021structured, hoogeboom2021argmaxflows}. 
We give the explicit matrices and analyze their trade-offs in Section~\ref{sec:formulations}; for the present purposes it suffices to note that absorbing-state masking dominates modern large-scale models~\citep{sahoo2024simpleeffect, shi2024simplified, nie2025largelanguag}.

\input{tables/02-notation}

\paragraph{Reverse process: learning to denoise.}
Generation proceeds by running the forward chain in reverse.
Starting from a sample $\bm{x}_T$ drawn from the stationary distribution (e.g., all-\texttt{m} or uniform), the model iteratively denoises by sampling $\bm{x}_{t-1} \sim p_\theta(\bm{x}_{t-1} \mid \bm{x}_t)$ for $t = T, T{-}1, \dots, 1$, ultimately producing a clean sample $\bm{x}_0$.
The reverse transition for each token is parameterized as:
\begin{equation}\label{eq:reverse}
  p_\theta(x_{t-1} \mid \bm{x}_t) \;=\; \mathrm{Cat}\!\bigl(x_{t-1};\; \bm{\pi}_\theta(\bm{x}_t, t)\bigr),
\end{equation}
where $\bm{\pi}_\theta(\bm{x}_t, t) \in \Delta^{K-1}$ (or $\Delta^K$ with the mask token) is a categorical distribution predicted by a neural network that takes the entire noisy sequence $\bm{x}_t$ and the timestep $t$ as input.
Note that the reverse model conditions on the \emph{full} noisy sequence $\bm{x}_t$, not just the single token at the current position, allowing it to exploit bidirectional context across all positions.

In practice, the network more commonly predicts a distribution over the \emph{clean} token, written $p_\theta(x_0 \mid \bm{x}_t, t)$, from which the reverse transition is recovered analytically using the known forward-process posterior. 
We refer to this dominant choice as the "$x_0$-parameterization" and defer its closed form, together with alternatives such as predicting $x_{t-1}$ directly, log-probability ratios, or a score analog, to Sections~\ref{sec:formulations} and~\ref{sec:training}.
We write $\bm{\pi}_\theta(\bm{x}_t,t)$ for this predicted clean-token distribution throughout; the equivalent symbols $\mu_\theta(\bm{x}_t,t)$ (used in the masked-diffusion objectives of Section~\ref{sec:training}) and $\hat{p}(x_0\mid \bm{x}_t)$ (used in the score/ratio discussion of Section~\ref{sec:formulations}) denote the same object, and we flag each on first use.

\paragraph{Conditioning.}
In conditional generation, the reverse process is augmented with a conditioning signal $\bm{c}$ (e.g., a text prompt, a class label, a structural constraint, or a partial sequence to be infilled):
\begin{equation}
  p_\theta(x_{t-1} \mid \bm{x}_t, \bm{c}) \;=\; \mathrm{Cat}\!\bigl(x_{t-1};\; \bm{\pi}_\theta(\bm{x}_t, t, \bm{c})\bigr).
\end{equation}
The conditioning signal is typically incorporated via cross-attention, prefix concatenation, or channel-wise concatenation within the denoiser architecture.
Training with classifier-free guidance, randomly dropping the conditioning signal during training so that the model learns both the conditional and unconditional distributions, enables guidance-style inference, where the predicted logits are interpolated between the conditional and unconditional predictions to amplify the influence of $\bm{c}$.
We return to conditioning and guidance in Sections~\ref{sec:training} and~\ref{sec:inference}.

\begin{figure}[t]
  \centering
  \includegraphics[width=\linewidth]{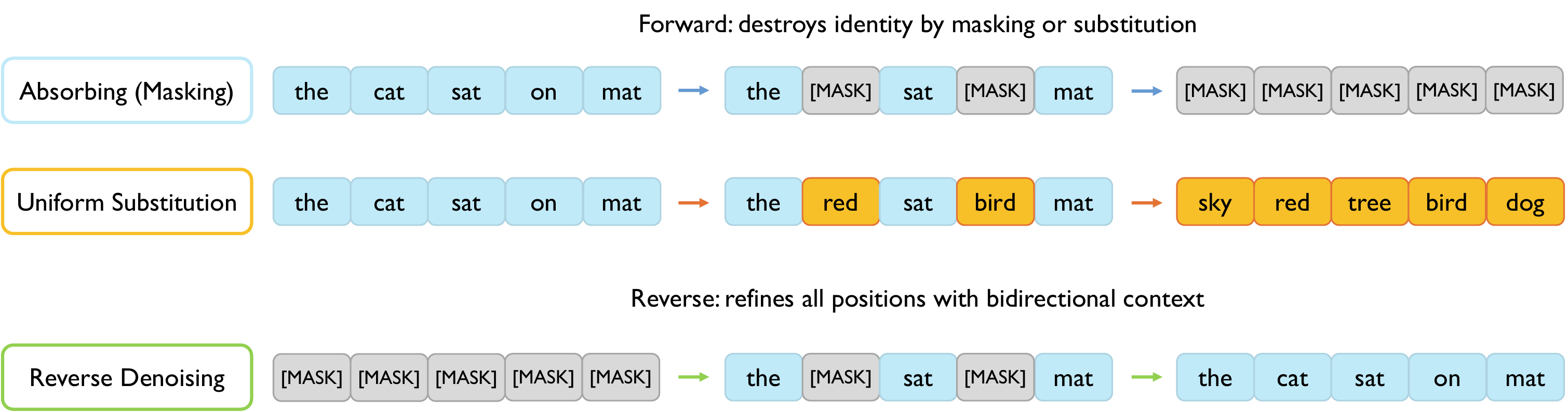}
  \caption{Discrete forward corruption and learned reverse denoising. Absorbing-state masking removes token identity, uniform substitution replaces tokens with categorical alternatives, and the reverse model denoises using full-sequence bidirectional context.}
  \label{fig:forward-reverse}
\end{figure}

\paragraph{Notation summary.}
Table~\ref{tab:notation} collects the core notation used throughout the paper for quick reference.

%% file: tables/02-notation.tex
 \begin{table}[t]
\centering
\caption{Core notation used throughout this paper.}
\label{tab:notation}
\small
\begin{tabular}{@{}ll@{}}
\toprule
\textbf{Symbol} & \textbf{Meaning} \\
\midrule
$K$ & Vocabulary size (number of categories) \\
$\mathcal{V} = \{1,\dots,K\}$ & Vocabulary; $\mathcal{V}_{\texttt{m}} = \mathcal{V} \cup \{\texttt{m}\}$ includes the mask token \\
$L$ & Sequence length \\
$\bm{x} = (x_1,\dots,x_L)$ & Discrete sequence; $\bm{x}_0$ is the clean data, $\bm{x}_t$ at corruption level $t$ \\
$\bm{e}_{k} \in \{0,1\}^K$ & One-hot vector for category $k$ \\
$T$ & Number of diffusion steps (discrete time) \\
$\bm{Q}_t \in [0,1]^{K\times K}$ & One-step transition matrix at step $t$; $[\bm{Q}_t]_{jk} = q(x_t{=}k\mid x_{t-1}{=}j)$ \\
$\bar{\bm{Q}}_t = \bm{Q}_1\cdots\bm{Q}_t$ & Cumulative transition matrix; gives $q(x_t\mid x_0)$ \\
$q(x_t \mid x_0)$ & Forward marginal at step $t$ \\
$p_\theta(x_{t-1}\mid \bm{x}_t)$ & Learned reverse transition \\
$\bm{\pi}_\theta(\bm{x}_t, t)$ & Predicted categorical distribution (reverse-process output) \\
$\bm{c}$ & Conditioning signal (prompt, label, constraint, etc.) \\
$\beta_t$ & Per-step noise rate (schedule parameter) \\
$\alpha_t = \prod_{s\le t}(1-\beta_s)$ & Cumulative survival probability of a token at step $t$ (masking schedule) \\
$\bm{R}_t$ & Continuous-time rate matrix (CTMC formulation; Section~\ref{sec:formulations}) \\
$s_\theta$ & Score / ratio network (score-based parameterization) \\
$\Delta^{K-1}$ & Probability simplex over $K$ categories \\
$\texttt{m}$ & Absorbing / mask token \\
\bottomrule
\end{tabular}
\end{table}

%% file: sections/04-tokenization.tex
\section{Discrete State Spaces and Tokenization} \label{sec:tokenization}

%
This section develops the central thesis of the paper: that the construction of the discrete state space, tokenization, is a first-class design axis for diffusion rather than a fixed preprocessing step. 
Section~\ref{subsec:tok_text} treats semantic tokens for text and code, Section~\ref{subsec:tok_quantized} treats quantized tokens for images, audio, and video, and Section~\ref{subsec:tok_science} treats the natural discrete alphabets of scientific domains. 
We begin in Section~\ref{subsec:tok_crossdomain} with a cross-domain definition of a token and an argument for why tokenization shapes corruption, denoising difficulty, controllability, and cost, and close in Section~\ref{subsec:tok_diagnostics} with diagnostics for evaluating tokenization quality from a diffusion-facing perspective.

\input{tables/03-tokenization}

\subsection{What Is a Token? A Cross-Domain View} \label{subsec:tok_crossdomain}
\noindent

\paragraph{A unifying definition.}
At the most abstract level, a \emph{token} is a discrete symbol drawn from a finite vocabulary $\mathcal{V}=\{1,\dots,K\}$ that serves as the atomic unit of categorical modeling.
Every discrete diffusion model operates over sequences of such symbols: the forward process corrupts tokens by replacing them with other tokens or with a distinguished absorbing state, and the reverse process predicts a categorical distribution over $\mathcal{V}$ at each position.
Tokenization is the process by which raw data (text, pixels, waveforms, amino-acid chains, molecular graphs) is mapped into this categorical representation.
Although the resulting symbols are always discrete integers, the \emph{origin} of the vocabulary varies fundamentally across domains, and this variation has deep consequences for how diffusion should be designed.
We distinguish three families of tokens that recur throughout this survey.

\emph{Semantic tokens} arise in language and code, where subword segmentation algorithms (BPE, WordPiece, Unigram, byte-level) partition a character stream into a fixed vocabulary of variable-length pieces.
These tokens are defined by frequency-based compression: they carry no intrinsic metric structure, and two tokens that are adjacent in vocabulary index may be semantically unrelated.
This lack of topology is why most text-based discrete diffusion models rely on absorbing-state (masking) corruption, which does not require any notion of inter-token distance~\citep{sahoo2024simpleeffect, shi2024simplified, nie2025largelanguag}.

\emph{Quantized tokens} arise in image, audio, and video generation, where continuous signals are discretized via vector quantization (VQ-VAE, VQ-GAN) or related codec architectures into codebook indices~\citep{chang2022maskgitmaske, yu2023magvitmasked, gu2022vectorquanti}.
Unlike text tokens, quantized codes \emph{do} possess geometric structure inherited from the codebook: codes whose embedding vectors are close in $\ell_2$ distance typically reconstruct perceptually similar patches, frames, or spectral segments.
This latent geometry creates an opportunity, largely underexploited in current practice, for structured corruption processes that treat nearby codes as more likely substitution targets, rather than treating all non-identity transitions as equally unlikely~\citep{austin2021structured, gu2024rethobjevect}.

\emph{Natural discrete alphabets} arise in scientific domains where the data are intrinsically categorical.
Proteins are sequences over the 20 standard amino-acids; DNA and RNA are sequences over 4-letter nucleotide alphabets; molecules is represented as graphs whose nodes and edges take categorical atom-type and bond-type labels.
These alphabets carry rich domain-specific semantics: amino-acid substitution rates have been estimated from evolutionary data, nucleotide mutations follow biochemical biases, and chemical valence rules impose hard constraints on which atom--bond combinations are valid.
In contrast to text tokens (no metric) and quantized codes (learned metric), natural alphabets come with \emph{external, domain-grounded notions of similarity and validity} that can inform the design of corruption processes, guide constrained sampling, and provide interpretable evaluation metrics~\citep{gruver2023proteindesig, avdeyev2023dirichlet, vignac2023digressdiscr}.

\begin{figure}[t]
  \centering
  \includegraphics[width=\linewidth]{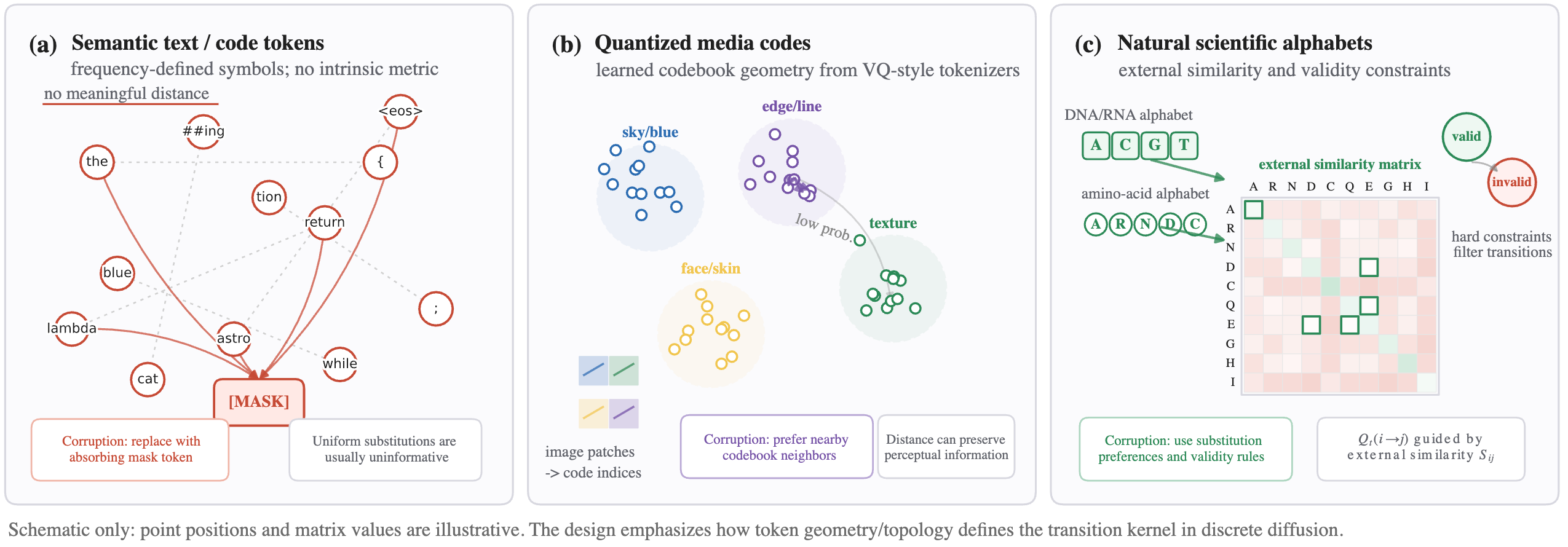}
  \caption{Token families in discrete diffusion.
(a) Semantic text and code tokens are frequency-defined categorical symbols with no intrinsic metric, so corruption is commonly represented as absorbing masking.
(b) Quantized media tokens inherit a learned codebook geometry, enabling structured corruption toward nearby codes.
(c) Natural scientific alphabets, such as nucleotide or amino-acid tokens, can use external similarity matrices and validity constraints to guide transition kernels.
}
  \label{fig:token-families}
\end{figure}

\paragraph{Why tokenization matters.}
The choice of tokenization is not a preprocessing step that can be separated from the diffusion model: it shapes the entire generative pipeline in at least four interrelated ways.

\emph{First, tokenization defines the topology of perturbations.}
A forward process that replaces one token with another is implicitly making a statement about which corruptions are "small".
In continuous diffusion, this is handled by Gaussian noise, which has a natural notion of locality in $\mathbb{R}^d$ (Section~\ref{sec:background}).
In discrete diffusion, there is no default metric: the transition matrix $\bm{Q}_t$ \emph{is} the corruption topology.
When $\bm{Q}_t$ is uniform (any token can be replaced by any other with equal probability), the model treats all errors as equally severe.
When $\bm{Q}_t$ reflects codebook geometry or substitution-matrix semantics, the model can exploit the fact that some corruptions are more informative than others.
And when $\bm{Q}_t$ is absorbing (tokens are only ever replaced by $\texttt{m}$), the model sidesteps the topology question entirely by reducing denoising to a fill-in-the-blank task.
Each of these choices leads to a qualitatively different denoising problem and generation behavior.

\emph{Second, tokenization determines the denoising difficulty curve.}
The difficulty of predicting a corrupted token depends on how much information the corruption has destroyed, which in turn depends on the relationship between the vocabulary structure and the noise process.
In a vocabulary where semantically similar tokens cluster, a small amount of substitution noise can be "undone" by exploiting local context; in an unstructured vocabulary the same noise level may be catastrophic.
Similarly, the ratio of vocabulary size $K$ to sequence length $L$ affects whether the model's capacity bottleneck is in per-token prediction (large $K$) or in modeling inter-position dependencies (large $L$).
These interactions between tokenization and noise schedule are often discovered empirically through expensive hyperparameter sweeps, but they are in principle predictable from properties of the token space (Section~\ref{subsec:tok_diagnostics}).

\emph{Third, tokenization governs downstream controllability and validity.}
Constrained generation, ensuring that outputs satisfy syntactic, chemical, or structural requirements, is fundamentally a question about which regions of $\mathcal{V}^L$ are valid.
A tokenization that aligns with domain constraints makes it easier to enforce validity during sampling: for example, if each token directly corresponds to a chemically meaningful fragment, valence rules can be checked locally at each position.
Conversely, a tokenization that scatters structural information across many tokens makes constraint satisfaction a global, combinatorial problem.

\emph{Fourth, tokenization sets the effective sequence length and thus the computational cost.}
Finer-grained tokenizations (byte-level text, pixel-level images) produce longer sequences, increasing the cost of bidirectional attention at each denoising step.
Coarser tokenizations (large-vocabulary subwords, low-resolution codebooks) produce shorter sequences but require the model to make more complex per-token predictions and may lose fine-grained detail.
This length-complexity tradeoff is particularly acute for diffusion models, which perform multiple forward passes over the full sequence during generation.

Taken together, these four axes explain why tokenization is not merely a practical choice but a first-class design axis for discrete diffusion.
The following subsections examine how each domain instantiates this tradeoff: text and code (Section~\ref{subsec:tok_text}), quantized media (Section~\ref{subsec:tok_quantized}), natural scientific alphabets (Section~\ref{subsec:tok_science}), and diagnostics for evaluating tokenization quality (Section~\ref{subsec:tok_diagnostics}).

\subsection{Text and Code Tokenization} \label{subsec:tok_text}
In natural language processing, subword segmentation methods such as Byte Pair Encoding (BPE), WordPiece, and Unigram are fundamental to large-scale autoregressive (AR) models. 
However, applying these tokenization schemes directly to discrete diffusion models introduces significant computational challenges. 
Modern language models often use vocabularies with more than 100,000 tokens. 
While AR models can accommodate this with a larger final softmax layer, discrete diffusion models require operations over the entire vocabulary space at each step. 
As a result, constructing and manipulating such large transition spaces leads to combinatorial explosion and numerical instability~\citep{chang2022maskgitmaske, yu2023magvitmasked}.
Beyond computational concerns, tokenization also affects sequence structure and semantic consistency. 
Subword methods frequently split a single semantic unit into multiple tokens of varying lengths.
When uniform diffusion noise is applied to these fragments, longer or more complex words are disproportionately corrupted, leading to semantic degradation~\citep{huang2022maskgeneadve}. 
Byte-level tokenization avoids this uneven fragmentation but significantly increases sequence length, resulting in higher memory and computational costs during diffusion.

These challenges reflect a deeper issue: text tokens are discrete categorical symbols without inherent continuous structure or meaningful distance metric. 
To avoid transitions between unrelated tokens during the diffusion process, recent approaches adopt a designated \texttt{m} token for corruption~\citep{wang2023binalatediff, gu2024rethobjevect}. 
Within this mask-based framework, vocabulary granularity plays a critical role. 
Coarser vocabularies allow the model to generate higher-level semantic units in fewer steps, but they also require more accurate predictions at each step. 
More importantly, token granularity shapes the refinement dynamics of diffusion. 
The noise schedule must be aligned with token structure to determine whether generation proceeds in a coarse-to-fine manner or follows a fine-to-coarse strategy~\citep{kim2025effigenemode, lezama2022imprmaskimag}.

The limitations of standard tokenization become even more pronounced when extending discrete diffusion models to programming languages. 
Unlike natural language, which can tolerate minor grammatical errors, code is highly sensitive to structural correctness. 
A single incorrect token such as a missing parenthesis or improper indentation, can invalidate the entire Abstract Syntax Tree (AST). 
In this setting, discrete diffusion models offer distinct advantages. 
Unlike autoregressive models, which generate tokens sequentially from left to right, diffusion models operate over the entire sequence in a bidirectional manner. 
This enables iterative refinement of global structure and improves the handling of long-range dependencies and strict formatting constraints, as also observed in multimodal token alignment tasks~\citep{chen2026aligvisufoun, wang2026scalspeetoke}.
Furthermore, the iterative denoising process enables new forms of constrained decoding. 
If an invalid token is produced during an intermediate step, external tools such as static analyzers or compilers can provide corrective feedback. 
This allows the generation process to be dynamically constrained within the space of syntactically valid programs, offering a level of grammar-aware control that is difficult to achieve with unidirectional autoregressive models.

\subsection{Quantized Tokens for Images, Audio, and Video} \label{subsec:tok_quantized}

\paragraph{Vector quantization.} 
Unlike text, continuous signals such as images, audio, and video do not have a natural discrete alphabet. 
For these modalities, quantized tokenization converts continuous signals into discrete integer indices drawn from a finite codebook.

Vector Quantized Variational Autoencoder (VQ-VAE)~\citep{vandenoord2017neurdiscrepr} is the dominant approach to media tokenization. 
It follows an encoder-codebook-decoder blueprint. The encoder compresses the input into a spatial grid of continuous latent vectors, each of which is snapped to its nearest entry in a learned codebook. 
The corresponding embedding vectors are then passed to the decoder for reconstruction, while the discrete indices are consumed by downstream generative models for generation tasks~\citep{chang2022maskgitmaske, lezama2022imprmaskimag}.
The training objective governs both reconstruction fidelity and codebook utilization. 
VQ-VAE uses a pixel-level L2 reconstruction loss~\citep{vandenoord2017neurdiscrepr}, which tends to produce blurry outputs and encourages codebook collapse, where most codebook entries go unused and the effective vocabulary shrinks. 
VQ-GAN~\citep{esser2021tamitranhigh} addresses both issues by replacing the pixel-wise L2 loss with a composite of a perceptual loss penalizing differences in intermediate VGG feature activations, and a patch-based adversarial loss from a discriminator that pushes reconstructed patches to be locally indistinguishable from real data. 
Together, these losses sharpen reconstructions and improve codebook utilization. 
Subsequent work continued to refine codebook training objectives: \citet{zhang2023reguvectquan} proposes regularization on both the prior-predicted token distribution gap and the stochastic mask during inference to mitigate codebook collapse, while SeQ-GAN~\citep{gu2024rethobjevect} and AlignTok~\citep{chen2026aligvisufoun} incorporate semantic-enhanced perceptual losses to better capture linguistic structure in the codebook. 
MAGVIT \citet{yu2023magvitmasked} extends the spatial codebook to 3D, enabling spatial-temporal tokenization for video sequences.

\paragraph{Advanced quantization.} Beyond standard vector quantization, two alternative quantization families have been explored, each trading off reconstruction fidelity, token efficiency, and training stability differently.

Residual Vector Quantization (RVQ) \citet{zeghidour2021sounstreend, defossez2022highfideneur} extends VQ by iteratively quantizing the residual error of each previous quantization stage, producing a hierarchy of codebook indices that together represent the input with increasing fidelity. 
This improves reconstruction quality without increasing sequence length, but introduces a deeper token hierarchy that complicates generative modelling. 
\citet{kim2025effigenemode} addresses this by directly predicting the cumulative vector embedding of all RVQ levels at each position simultaneously, decoupling inference steps from both sequence length and depth. 
For audio signals, the challenge is especially acute: the high information density of speech requires multiple RVQ codebook levels to maintain fidelity, inflating the token budget and undermining language modelling efficiency. 
SiTok~\citep{wang2026scalspeetoke} addresses this by replacing the deterministic RVQ reconstruction objective with a diffusion autoencoder that explicitly models quantization uncertainty, achieving high-fidelity reconstruction at an extremely low bit rate.

At the other end of the spectrum, scalar quantization (SQ)~\citep{wang2023binalatediff} independently maps each dimension of a continuous latent vector to the nearest level on a fixed axis-aligned grid, requiring no learned codebook and avoiding the collapse and cascade instabilities of VQ and RVQ. 
However, because its boundaries cannot adapt to the geometry of the learned representation, SQ ignores inter-dimensional correlations and can produce systematic reconstruction artifacts when semantically meaningful directions in latent space are not axis-aligned with the quantization grid. 
Together, RVQ and SQ illustrate the fundamental tradeoff in advanced quantization: richer hierarchical structure improves fidelity but complicates generation, while simpler fixed-grid schemes are stable but geometrically inflexible.

\paragraph{Codebook topology.} 
Beyond quantization fidelity, the topology of the codebook has important consequences for generative modeling. 
The semantic distance between two codebook entries measures the perceptual or linguistic difference between the inputs they represent. 
In standard VQ training, geometric proximity between embedding vectors is decoupled from semantic similarity, since the reconstruction objective encourages acoustic fidelity rather than linguistic organisation. 
An ideally structured codebook would satisfy the property that semantically similar inputs map to geometrically nearby entries, making geometric distance a reliable proxy for semantic distance. 
Approaches such as semantic distillation~\citep{ye2024codedoesmatt} and CTC-based supervision~\citep{wang2026scalspeetoke} explicitly reshape the codebook topology toward this goal, producing more regular and linguistically meaningful discrete spaces.
This regularity can be directly exploited in discrete diffusion by designing structured, non-uniform corruption processes. 
Standard discrete diffusion assumes a uniform transition distribution, where each token is equally likely to be replaced by any other, and this discards all codebook geometry. 
In contrast, a structured transition matrix that assigns higher corruption probability to semantically nearby codes allows intermediate noisy tokens to retain information about their origins~\citep{austin2021structured}. 
This produces better-conditioned denoising targets and stronger training signal at intermediate corruption levels, motivating joint design of the tokenizer and the generative model so that a semantically regular codebook can be fully exploited by a structured discrete diffusion process~\citep{kim2025effigenemode, wang2026scalspeetoke}.

\subsection{Natural Discrete Structures in Science} \label{subsec:tok_science}
Biological sequences provide the most direct instances of natural discrete alphabets for diffusion models. 
Proteins are strings over a 20-letter amino-acid vocabulary, while DNA and RNA are composed of 4-letter nucleotide alphabets. 
These alphabets are compact enough to define tractable categorical state spaces, yet the sequences they form exhibit long-range dependencies that pose a distinct challenge for iterative denoising.
In proteins, residues that are distant in primary sequence can be proximal in three-dimensional structure, so a diffusion model must learn co-evolutionary couplings that span hundreds of positions to produce stable and functional folds~\citep{gruver2023proteindesig, wang2024diffusion}. 
Analogous long-range constraints arise in genomic sequences, where regulatory elements such as promoters, enhancers, and splice sites can modulate gene expression across kilobase-scale distances~\citep{avdeyev2023dirichlet, stark2024dirichlet}. 
Because these dependencies are non-local and often non-contiguous, absorbing-state (masking) diffusion has been the dominant corruption strategy: it avoids introducing biologically implausible substitutions during the forward process and allows the reverse process to reconstruct correlated positions jointly rather than independently~\citep{sahoo2024simpleeffect, campbell2024allatominver}. 
Domain-specific priors can further inform the noise process; for example, evolutionary substitution matrices such as BLOSUM encode empirical amino-acid exchange abilities and could serve as structured transition kernels, though most current models still rely on uniform or absorbing transitions~\citep{austin2021structured}.

Molecules and materials present a complementary tokenization challenge because they may be represented either as linear molecular strings or as graphs.
String-based representations, including SMILES, SELFIES, and fragment-sequence representations such as SAFE, make molecules compatible with sequence models and discrete diffusion over token sequences.
Recent systems such as GenMol apply masked discrete diffusion to SAFE sequences, using fragments as basic building blocks and enabling fragment remasking for molecular optimization~\citep{unknown2025genmdrugdisc}.
Graph-based representations instead map atoms to categorical node types and bonds to categorical edge types, making the state space a discrete combinatorial object with hard validity constraints: each atom must satisfy valence rules, the resulting graph should be connected, and stereochemical configurations should be self-consistent.
DiGress~\citep{vignac2023digressdiscr} operates directly on the joint node-edge categorical space, applying discrete noise to both adjacency and feature matrices, while flow-matching alternatives such as discrete flow matching~\citep{campbell2024generative} parameterize interpolating paths between noise and data distributions over graphs.
A persistent difficulty is that even small violations of chemical validity render the generated molecule meaningless, motivating guidance and projection mechanisms at inference time (discussed in Section~\ref{sec:inference}).
Beyond small molecules, crystalline materials introduce additional discrete variables such as space-group symmetries and Wyckoff positions~\citep{zeni2024mattergen}, further expanding the categorical design space that tokenization must accommodate.

An important design axis for scientific domains is the granularity at which the object is discretized.
For molecules, atom-level graph representations preserve full chemical expressiveness but expose the denoiser to tight local constraints, increasing the computational burden of multi-step denoising.
Coarser fragment-, scaffold-, or motif-level alternatives decompose molecules into chemically meaningful substructures and treat each substructure as a generation unit, reducing effective sequence or graph size and often encoding local chemical validity by construction~\citep{campbell2024fragfmeffici, maziarz2022learning}.
For proteins, the analogous granularity choice appears in residue-level amino-acid tokens, structure-aware residue vocabularies, structural alphabets, and learned structure tokens.
For example, Foldseek-style structural alphabets and SaProt-style structure-aware vocabularies augment each residue with local geometric information, while VQ-VAE-based protein structure tokenizers discretize continuous backbone or tertiary-structure geometry into finite codebooks~\citep{su2024saprot, yuan2025proteinstructuretokenization, gao2024foldtoken}.
These representations trade off reconstruction fidelity, structural locality, sequence length, and compatibility with sequence-based language or diffusion models.
The choice among these representations determines the effective state-space size, the severity of validity constraints the model must learn or enforce, and the length of the sequence the diffusion process must denoise, echoing the vocabulary-granularity tradeoffs discussed for text in Section~\ref{subsec:tok_text}.

\subsection{Diagnostics and Metrics for Tokenizations} \label{subsec:tok_diagnostics}

\noindent
Evaluating a tokenization scheme requires metrics that go beyond downstream task performance to characterize how well the discrete representation preserves information, respects semantic structure, and supports iterative denoising. 
We organize tokenization diagnostics into three complementary families: information-theoretic and compression metrics, geometric and topological probes, and diffusion-facing diagnostics that measure the interaction between the token space and the generative process.
Table~\ref{tab:tokenization_diagnostics} turns these diagnostics into a reporting checklist. 
We distinguish established tokenizer metrics from proposed or synthesized diffusion-facing probes; the latter should be read as actionable recommendations for tokenizer--diffusion co-design rather than as validated predictors of generation quality.
We flag the status of each family explicitly: the information/compression and reconstruction metrics below are established and widely reported, whereas the geometric/topological probes and the diffusion-facing diagnostics are largely \emph{our own proposals} synthesized from scattered practice rather than standardized tools; where a proposed diagnostic has, to our knowledge, not yet been systematically measured, we say so at that point. 
This subsection is therefore part survey and part position; we have not validated the proposed diagnostics with new experiments, and we present them as recommendations for future tokenizer-diffusion co-design.

\input{tables/03b-tokenization-metrics}

\paragraph{Information and compression metrics.}
For text and code, where tokenization is typically deterministic (e.g., BPE or byte-level), the relevant information-theoretic measure is the token-level entropy of the data distribution under the chosen vocabulary. 
Larger vocabularies compress more information per token but increase the categorical state space the diffusion model must navigate. 
Bits-per-character (BPC) or bits-per-byte metrics normalize by input length rather than token count, enabling comparison across tokenizers with different vocabulary sizes~\citep{lou2024discrete}.

For quantized media tokenizers, reconstruction fidelity is the most direct indicator of information retention. 
Standard pixel-level metrics such as peak signal-to-noise ratio (PSNR) and structural similarity index (SSIM) quantify low-level fidelity but correlate poorly with human perception; learned perceptual metrics such as LPIPS, which measures the distance between intermediate feature activations of a pretrained network, capture semantic preservation more faithfully and are widely used both as training losses and evaluation metrics in VQ-GAN-style tokenizers~\citep{esser2021tamitranhigh}. 
Reconstruction Fr\'{e}chet Inception Distance (rFID), i.e., FID computed between original and reconstructed images, has become the de facto aggregate metric for image tokenizer quality~\citep{yu2023magvitmasked, chang2022maskgitmaske}. 
For audio and video, analogous distributional metrics in domain-specific feature spaces serve a similar purpose.
Beyond reconstruction, it is essential to assess how effectively the codebook is utilized.
Codebook perplexity (defined as the exponentiated entropy of the empirical code usage distribution, $\mathrm{Perp} = \exp\bigl(-\sum_{k=1}^{K} p(k) \log p(k)\bigr)$, where $p(k)$ is the fraction of encoded tokens assigned to code $k$) measures the effective vocabulary size~\citep{vandenoord2017neurdiscrepr}. 
A codebook with $K$ entries but perplexity much smaller than $K$ suffers from codebook collapse. 
A complementary diagnostic is the \emph{dead code ratio}: the fraction of entries with zero or near-zero usage, which directly quantifies wasted capacity. 
Regularization strategies that penalize low-entropy usage distributions~\citep{zhang2023reguvectquan} or semantic-enhanced objectives~\citep{gu2024rethobjevect, chen2026aligvisufoun} improve utilization, and monitoring perplexity throughout training serves as an early warning for collapse.

\paragraph{Geometric and topological diagnostics.}
As discussed in Section~\ref{subsec:tok_quantized}, the topology of the codebook, whether geometric proximity among code embeddings reflects semantic similarity, has direct consequences for structured corruption in discrete diffusion. 
Here we focus on \emph{how to measure} this alignment, a question that remains largely open, likely because tokenizers and diffusion models are typically developed and evaluated separately, with end-to-end generation quality serving as the only joint assessment.

We suggest that a natural diagnostic is \emph{neighborhood consistency}: given a distance metric in the embedding space and a semantic similarity measure (e.g., perceptual distance between the inputs reconstructed by each code), one can compute the rank correlation between the two. 
A well-structured codebook exhibits high correlation, meaning that $k$-nearest neighbors in embedding space also reconstruct semantically similar inputs. 
Approaches that explicitly reshape codebook geometry via semantic distillation~\citep{ye2024codedoesmatt} or supervision from pretrained models~\citep{chen2026aligvisufoun, wang2026scalspeetoke} could be evaluated by tracking this correlation before and after the intervention, though to our knowledge such systematic measurements have not yet been reported.

For scientific domains with natural discrete alphabets, external references for inter-token similarity already exist. 
For example, evolutionary substitution matrices such as BLOSUM for amino acids, or atom-type similarity based on chemical properties for molecular graphs. 
We note that these could serve as ground-truth references against which to calibrate the corruption process of domain-specific discrete diffusion models, providing a principled alternative to the unstructured transitions (uniform or absorbing) that current models employ~\citep{austin2021structured}. 
Domain-specific generation metrics (validity, uniqueness, and novelty (V.U.N.) for molecules~\citep{vignac2023digressdiscr}, or structure-prediction confidence (e.g., pLDDT) for proteins~\citep{wang2024diffusion}) also serve as indirect tokenization diagnostics: if a token space cannot support the generation of valid structures, the representation itself may be deficient regardless of reconstruction fidelity.

\paragraph{Diffusion-facing diagnostics.}
The ultimate test of a tokenization scheme is how it affects the learnability and quality of the denoising process. 
We highlight three diagnostics that directly probe this interaction.

\emph{The first is the reconstruction-generation gap.} 
Comparing reconstruction quality (rFID, which isolates tokenizer fidelity) with generation quality (generation FID, which reflects the full tokenizer-plus-diffusion pipeline) reveals how much performance is lost in the generative modeling stage. 
A large gap indicates that the token space is difficult for the diffusion model to learn, even when the tokenizer itself reconstructs well. 
While both metrics are routinely reported, explicitly tracking their gap across tokenizer variants would directly isolate the effect of tokenization on generative performance~\citep{chang2022maskgitmaske, yu2023magvitmasked}.

\emph{The second is the denoising loss curves across timesteps.} 
Plotting the per-timestep denoising loss $\mathcal{L}(t)$ as a function of the corruption level $t$ reveals the difficulty profile of the diffusion task. 
For absorbing-state (masking) diffusion, the loss at each $t$ reflects how difficult it is to predict masked tokens given the fraction $1-\alpha_t$ that have been masked~\citep{sahoo2024simpleeffect, shi2024simplified}. 
Comparing these curves across tokenizers for the same data and architecture would isolate the effect of tokenization on denoising difficulty. 
Tokenizations that spread difficulty relatively uniformly across timesteps are generally preferable, as they provide informative training signal at all noise levels.

\emph{The third is the sensitivity to noise schedule.} 
For a given tokenization, sweeping over schedule families (linear, cosine, log-linear) and measuring final sample quality reveals the degree of coupling between tokenizer design and schedule tuning~\citep{sahoo2024simpleeffect}. 
Tokenizations that are robust across schedules simplify model development. 
Recent theoretical work on masked diffusion clarifies this interaction, showing that the optimal unmasking order depends on the conditional entropy structure of token predictions, and that random-order masking during training implicitly prepares the model for worst-case orderings~\citep{unknown2025traiworsplan}.

Together, these three families of diagnostics provide a toolkit for evaluating and comparing tokenization schemes. 
The geometric and diffusion-facing diagnostics remain underexplored relative to reconstruction metrics: at present, there is no established way to predict how well a tokenizer will support diffusion-based generation without training a generative model end-to-end. 
Closing this gap, developing lightweight and model-free proxies for diffusion-readiness, is an important open problem. 
In the meantime, we encourage practitioners to report metrics from each family when proposing new tokenizers for discrete diffusion.

%% file: tables/03-tokenization.tex
\begin{table}[t]
\centering
\caption{Tokenization families across domains and their implications for discrete diffusion. "Metric" denotes whether the token space carries a usable notion of inter-token similarity; "Typical corruption" denotes the forward process most commonly paired with each family.}
\label{tab:tokenization_families}
\small
\setlength{\tabcolsep}{4pt}
\renewcommand{\arraystretch}{1.15}
\begin{tabular}{@{}p{3cm}p{2.6cm}p{2.0cm}p{3.5cm}p{3.5cm}@{}}
\toprule
\textbf{Token family} & \textbf{Examples} & \textbf{Metric} & \textbf{Typical corruption} & \textbf{Key design concern} \\
\midrule
Semantic \\(text, code) & BPE, WordPiece, byte-level & None (index-arbitrary) & Absorbing / masking & Vocabulary granularity vs.\ sequence length \\
\midrule
Quantized \\(image, audio, video) & VQ-VAE, VQ-GAN, RVQ codes & Learned (codebook geometry) & Masking; structured (underused) & Codebook collapse; topology--semantics alignment \\
\midrule
Natural alphabets \\(proteins, DNA/RNA) & Amino acids, nucleotides & External (BLOSUM, biochemical) & Absorbing; structured possible & Long-range dependencies; biological validity \\
\midrule
Graph/combinatorial \\(molecules, layouts) & Atom/bond types, element classes & Partial (chemical similarity) & Joint node--edge masking & Hard validity (valence, connectivity) \\
\bottomrule
\end{tabular}
\end{table}

%% file: tables/03b-tokenization-metrics.tex
\begin{table}[t]
\centering
\caption{Tokenizer diagnostics for discrete diffusion. "Status" indicates whether the diagnostic is already standard in tokenizer evaluation or is proposed/synthesized here as a diffusion-facing reporting recommendation. "Requires DDM" indicates whether the diagnostic requires training or running a discrete diffusion model rather than evaluating the tokenizer alone.}
\label{tab:tokenization_diagnostics}
\small
\setlength{\tabcolsep}{3pt}
\renewcommand{\arraystretch}{1.15}
\begin{tabular}{@{}p{3.1cm}p{2.45cm}p{1.8cm}p{4cm}p{4.3cm}@{}}
\toprule
\textbf{Diagnostic} & \textbf{Status} & \textbf{Requires DDM} & \textbf{What it detects} & \textbf{Recommended reporting} \\
\midrule
Token entropy / BPC & Established & No & Compression--granularity tradeoff for text/code tokenizers; large vocabularies shorten sequences but enlarge the categorical state space & Report vocabulary size, average sequence length, token entropy, and BPC/BPB under the same corpus preprocessing \\
\midrule
Reconstruction fidelity & Established & No & Information loss introduced by quantization or codec tokenization before diffusion is trained & Report domain-appropriate reconstruction metrics, e.g., PSNR/SSIM/LPIPS/rFID for images or analogous audio/video metrics \\
\midrule
Codebook utilization & Established & No & Codebook collapse or wasted capacity when only a small subset of codes is used frequently & Report codebook perplexity, perplexity-to-size ratio, and dead-code ratio with the dead-code threshold specified \\
\midrule
Neighborhood consistency & Proposed / underexplored & No, if a semantic metric is available & Whether geometric neighbors in the codebook correspond to semantically similar reconstructions or domain objects & Report rank correlation between code-space nearest neighbors and semantic/perceptual similarity; specify the distance and similarity metrics \\
\midrule
Domain-prior alignment & Proposed / domain-dependent & No & Whether token similarity or transition structure agrees with external priors such as biochemical substitution matrices or chemical similarity & Report correlation or divergence between the proposed transition kernel and the domain prior, when such a prior exists \\
\midrule
Reconstruction--generation gap & Synthesized / not standardized & Yes & Cases where a tokenizer reconstructs well but produces a token space that is difficult for the diffusion model to learn & Report reconstruction quality and generation quality side by side, e.g., rFID versus generation FID, under matched architecture and training budget \\
\midrule
Denoising loss curve & Proposed / not standardized & Yes & Timestep-specific denoising difficulty induced by a tokenizer and corruption schedule & Plot or tabulate $\mathcal{L}(t)$ across timesteps for fixed data, architecture, objective, and schedule \\
\midrule
Schedule sensitivity & Proposed / not standardized & Yes & Coupling between tokenizer design and noise-schedule tuning; high sensitivity suggests fragile tokenizer--diffusion interaction & Sweep standard schedules, e.g., linear, cosine, and log-linear, and report mean quality plus variance or range across schedules \\
\bottomrule
\end{tabular}
\end{table}

%% file: sections/05-formulation.tex
\section{Core Formulations of Discrete Diffusion} \label{sec:formulations}

\noindent
This section presents the major formulation families for discrete diffusion in detail.
Building on the notation introduced in Section~\ref{sec:background}, we move from general transition-matrix models to the specific corruption families that dominate practice, then to continuous-time and score-based perspectives, and finally to a unifying view that reveals the common structure underlying all variants.


\subsection{Discrete-Time Markov Chains and Transition Matrices} \label{subsec:form_d3pm}
\noindent

\paragraph{The D3PM framework.}
The Discrete Denoising Diffusion Probabilistic Model (D3PM)~\citep{austin2021structured} provides the foundational framework for discrete diffusion.
Recall from Section~\ref{sec:background} that each token $x_i$ (we again drop the position subscript when the per-position structure is clear) evolves under a forward Markov chain parameterized by a transition matrix $\bm{Q}_t \in [0,1]^{K \times K}$, with cumulative corruption matrix $\bar{\bm{Q}}_t = \bm{Q}_1 \cdots \bm{Q}_t$ giving the marginal $q(x_t \mid x_0) = \mathrm{Cat}(x_t;\, \bm{e}_{x_0}^\top \bar{\bm{Q}}_t)$.
The key technical insight of D3PM is that the \emph{reverse posterior}, the distribution of the previous state given the current noisy state and the clean data, is available in closed form:
\begin{equation}\label{eq:reverse_posterior}
  q(x_{t-1} \mid x_t, x_0) \;=\; \frac{q(x_t \mid x_{t-1})\, q(x_{t-1} \mid x_0)}{q(x_t \mid x_0)} \;\propto\; \bigl[\bm{e}_{x_t}^\top \bm{Q}_t^\top\bigr]_{x_{t-1}} \cdot \bigl[\bm{e}_{x_0}^\top \bar{\bm{Q}}_{t-1}\bigr]_{x_{t-1}},
\end{equation}
where the denominator $q(x_t \mid x_0)$ serves as a normalizing constant.
Here the two bracketed terms are the $x_{t-1}$-th entries of the row vectors $\bm{e}_{x_t}^\top \bm{Q}_t^\top$ and $\bm{e}_{x_0}^\top \bar{\bm{Q}}_{t-1}$, and the product is taken elementwise over the $K$ candidate values of $x_{t-1}$; normalizing this length-$K$ vector by $q(x_t \mid x_0) = \sum_{x_{t-1}} (\cdot)$ recovers the categorical posterior.
This posterior plays a central role: since the reverse process $p_\theta(x_{t-1} \mid x_t)$ must approximate $q(x_{t-1} \mid x_t, x_0)$ averaged over $x_0$, one can parameterize the model to predict $\hat{x}_0 = f_\theta(\bm{x}_t, t)$ and then plug the prediction into Eq.~\ref{eq:reverse_posterior} to obtain the reverse transition analytically.
This \emph{predict-$x_0$-then-posterior} strategy avoids learning the reverse distribution directly and is the dominant parameterization across most discrete diffusion models.

\paragraph{Design space of the transition matrix.}
The choice of $\bm{Q}_t$ is the primary degree of freedom in the D3PM framework, encoding the model designer's prior about what kinds of corruption are meaningful.
D3PM~\citep{austin2021structured} introduced and compared several designs.
 
\emph{The first is the uniform substitution.} 
Each token is independently replaced by a uniformly random token with probability $\beta_t$:
\begin{equation}\label{eq:q_uniform}
  \bm{Q}_t^{\text{uniform}} \;=\; (1 - \beta_t)\,\bm{I} \;+\; \frac{\beta_t}{K}\,\bm{1}\bm{1}^\top.
\end{equation}
This treats all token substitutions as equally plausible regardless of identity.
The stationary distribution is uniform over $\mathcal{V}$, and the cumulative corruption $\bar{\bm{Q}}_t$ has a simple closed form.
The uniform distribution is stationary because $\bm{Q}_t^{\text{uniform}}$ is doubly stochastic, so it leaves the uniform vector $\tfrac{1}{K}\bm{1}$ invariant; iterating drives any initial distribution toward $\tfrac{1}{K}\bm{1}$ as $\prod_s(1-\beta_s)\to 0$.
While conceptually clean, uniform substitution makes denoising difficult in practice: the model must distinguish the true token from $K-1$ equally likely alternatives at each corrupted position, without any structural signal about which alternatives are "close."
 
\emph{The second is the absorbing (masking).} 
Each token is replaced by a special absorbing token $\texttt{m} \notin \mathcal{V}$ with probability $\beta_t$ and remains unchanged otherwise; once masked, a token stays masked (so $\bm{Q}_t^{\text{absorb}}$ acts on the extended vocabulary $\mathcal{V}_{\texttt{m}}=\{1,\dots,K,\texttt{m}\}$ of Table~\ref{tab:notation} and is therefore $(K{+}1)\times(K{+}1)$):
\begin{equation}\label{eq:q_absorbing}
  [\bm{Q}_t^{\text{absorb}}]_{jk} \;=\;
  \begin{cases}
    1 - \beta_t & \text{if } j = k \neq \texttt{m}, \\
    \beta_t     & \text{if } k = \texttt{m},\; j \neq \texttt{m}, \\
    1           & \text{if } j = k = \texttt{m}, \\
    0           & \text{otherwise}.
  \end{cases}
\end{equation}
The stationary distribution concentrates all mass on $\texttt{m}$.
This follows because $\texttt{m}$ is absorbing (its row is $\bm{e}_{\texttt{m}}^\top$) and every non-mask state has probability $\beta_t>0$ of transitioning to it, so the only invariant distribution is the point mass $\bm{e}_{\texttt{m}}$.
The denoising task reduces to fill-in-the-blank prediction at masked positions, a problem for which bidirectional Transformers are particularly well suited.
This is the design that has scaled most successfully; we discuss it in detail in Section~\ref{subsec:form_masking}.
 
\emph{The third is the embedding-aware / structured transitions.} 
The transition probabilities depend on token identity via an external similarity measure.
For example, one can define $[\bm{Q}_t]_{jk} \propto \exp(-\|\bm{v}_j - \bm{v}_k\|^2 / \sigma_t^2)$ for token embeddings $\bm{v}_j, \bm{v}_k$, so that tokens with similar embeddings are more likely substitution targets.
This encodes the inductive bias that "nearby" tokens should be confused before "distant" ones, producing a coarse-to-fine corruption trajectory.
D3PM explored such designs but found them harder to tune and less stable than absorbing-state corruption in practice~\citep{austin2021structured}.
Subsequent work has revisited structured transitions for domains with natural similarity metrics (e.g., amino-acid substitution matrices for proteins or edge-type similarities for molecular graphs), where the external metric provides a well-grounded notion of proximity~\citep{gruver2023proteindesig, avdeyev2023dirichlet}.

Multinomial diffusion~\citep{hoogeboom2021argmaxflows} independently developed a closely related framework, defining the forward process as interpolation toward a uniform categorical distribution and connecting the reverse process to a discrete analogue of the score.
Together, D3PM and multinomial diffusion established the transition-matrix view as the canonical starting point for discrete diffusion, and nearly all subsequent formulations can be understood as specializations, continuous-time limits, or reparameterizations of this general framework.

\subsection{Masking and Absorbing-State Diffusion} \label{subsec:form_masking}

\paragraph{Why absorbing-state diffusion dominates in practice.}
Among the transition-matrix designs introduced in Section~\ref{subsec:form_d3pm}, absorbing-state (masking) corruption has emerged as the most dominant choice for large-scale discrete diffusion, particularly in language modeling.
Three properties explain this dominance.

\emph{Simplicity.}
The forward process has a single free parameter per step (the masking rate $\beta_t$), and the cumulative masking probability $\alpha_t = \prod_{s=1}^t (1 - \beta_s)$ gives the probability that a token survives to step $t$ in closed form.
The marginal at any step is a mixture: with probability $\alpha_t$ the token equals $x_0$, and with probability $1 - \alpha_t$ it equals $\texttt{m}$.
Training reduces to a reweighted masked-language-modeling (MLM) loss, where the model predicts the identity of masked tokens given the unmasked context~\citep{sahoo2024simpleeffect, shi2024simplified}.

\emph{Training stability.}
Because the denoising task at each timestep is a classification problem over $\mathcal{V}$ (predict the original token at a masked position), the loss landscape is well behaved: cross-entropy over a softmax output, with no need for the complex reweighting schemes or auxiliary losses that can arise with substitution-based corruption.
The MDLM~\citep{sahoo2024simpleeffect} and MD4~\citep{shi2024simplified} frameworks showed that the continuous-time ELBO for masked diffusion collapses to a simple integral of reweighted cross-entropy terms, making the connection to MLM losses explicit and facilitating stable large-scale training.

\emph{Compatibility with bidirectional architectures.}
Masked diffusion naturally pairs with bidirectional (full-attention) Transformers: at each denoising step, the model sees a partially masked sequence and must fill in the blanks using context from both directions.
This is architecturally identical to BERT-style pre-training, allowing direct reuse of well-understood Transformer architectures, positional encodings, and training infrastructure~\citep{sahoo2025scalingup, nie2025largelanguag}.
The bidirectional context also provides a natural advantage for infilling and editing tasks, where the model must condition on both a prefix and a suffix.

\paragraph{Masking schedules and variants.}
Within the absorbing-state framework, the masking schedule $\{\beta_t\}_{t=1}^T$ (equivalently, the survival probability $\alpha_t$) controls the rate at which tokens are corrupted and thus the difficulty profile of the denoising task across timesteps.
Common choices include linear schedules ($\alpha_t$ decreasing linearly from 1 to 0), cosine schedules (following a shifted cosine curve), and log-linear schedules.
Recent theoretical work has shown that the cosine schedule is Fisher-Rao optimal for masked discrete diffusion under certain assumptions, providing a principled justification for what was originally an empirical heuristic~\citep{unknown2025cosischefish}.

Several variants extend basic per-token masking:
\emph{Span masking} corrupts contiguous spans of tokens rather than individual positions, encouraging the model to learn phrase-level coherence and reducing the effective number of independent predictions per step.
\emph{Structured masking} for images masks spatial patches or frequency bands rather than individual codes, inducing coarse-to-fine generation that respects the spatial structure of the token grid~\citep{chang2022maskgitmaske}.
\emph{Partial masking}~\citep{unknown2025beyomaskunma} generalizes absorbing-state diffusion by allowing tokens to transition to states other than $\texttt{m}$, for example, to a semantically similar token, while retaining the simplicity of a single-step masking rate.
\emph{Any-order masking}~\citep{unknown2025anyogptmask, unknown2025unifmaskdiff} decouples the masking (corruption) process from the unmasking (generation) order, showing that the same trained model can support diverse generation orders at inference time, bridging masked diffusion and autoregressive generation within a single framework.

\subsection{Substitution Noise and Mixture Corruptions}

\paragraph{Uniform substitution as denoising under label noise.}
When the forward process replaces a token with a uniformly random category (Eq.~\ref{eq:q_uniform}), the denoising task at each corrupted position can be interpreted as classification under heavy \emph{label noise}: the model observes a "label" (the corrupted token $x_t$) that has been randomized with probability $1-\alpha_t$ and must recover the true class $x_0$.
This connection to the noise-robust classification literature~\citep{austin2021structured} has two practical consequences.
First, the difficulty of denoising scales with the vocabulary size $K$: for large $K$, even a moderate corruption rate renders the corrupted token nearly uninformative about $x_0$, collapsing the task to unconditional prediction from context alone.
Second, unlike absorbing-state corruption, the model must learn to distinguish two cases, whether the observed token is the original (with probability $\alpha_t$) or a random substitute (with probability $1-\alpha_t$), which adds a source of estimation difficulty that does not arise with masking.

\paragraph{Hybrid and mixture corruptions.}
Pure masking and pure substitution represent two extremes of a design spectrum.
Several works have explored \emph{hybrid} corruption processes that combine both mechanisms, typically by partitioning the corruption probability at each step into a masking component and a substitution component.
The motivation is twofold.
First, pure masking can lead to \emph{degenerate dynamics}: at intermediate timesteps, the model only ever sees tokens that are either perfectly clean or fully masked, with no partially corrupted intermediate states.
This means the model never practices recovering from plausible-but-wrong tokens, a scenario that arises naturally during iterative sampling when the model's own predictions (which may be incorrect) are fed back as input at the next step.
Adding substitution noise forces the model to handle corrupted-but-non-masked tokens, improving robustness to its own errors during generation~\citep{unknown2025beyomaskunma}.
Second, substitution noise provides a richer training signal at intermediate corruption levels.
With pure masking, the loss at a given timestep $t$ depends only on the fraction of tokens that are masked; all information about "how close" a corruption is to the clean data is lost.
With substitution noise, corrupted tokens retain partial information (a random token is occasionally the correct one by chance), and the model can learn to exploit this signal.

Concrete instantiations include D3PM's hybrid matrices~\citep{austin2021structured}, which interpolate between absorbing and uniform transitions, and more recent approaches that learn or anneal the mixing ratio during training.
The general finding is that small amounts of substitution noise improve generation quality, but the optimal mixing ratio is domain- and scale-dependent, and pure masking remains competitive at large scale when combined with modern training recipes~\citep{sahoo2024simpleeffect, sahoo2025scalingup}.

\subsection{Continuous-Time Discrete Diffusion} \label{subsec:continuous_time_discrete_diffusion}

\paragraph{The CTMC formulation.}
An alternative to discrete-time Markov chains is to define the forward process as a \emph{continuous-time Markov chain} (CTMC) over the categorical state space $\mathcal{V}$.
Corruption is governed by a \emph{rate matrix} $\bm{R}_t \in \mathbb{R}^{K \times K}$ (with off-diagonal entries non-negative and rows summing to zero), so that the probability of a transition from state $j$ to state $k \neq j$ in an infinitesimal interval $[t, t+dt)$ is $[\bm{R}_t]_{jk}\,dt$.
The transition probabilities over a finite interval are obtained by solving the Kolmogorov forward equation:
\begin{equation}\label{eq:ctmc_forward}
  \frac{d}{dt}\bm{P}_t \;=\; \bm{P}_t\,\bm{R}_t, \qquad \bm{P}_0 = \bm{I},
\end{equation}
where $[\bm{P}_t]_{jk} = q(x_t{=}k \mid x_0{=}j)$ is the transition probability matrix from $0$ to $t$.
For time-homogeneous rates ($\bm{R}_t = \bm{R}$ constant), this simplifies to the matrix exponential $\bm{P}_t = \exp(t\bm{R})$.
The continuous-time framework subsumes the discrete-time setting: a discrete-time chain with step matrices $\bm{Q}_1, \dots, \bm{Q}_T$ can be recovered by choosing piecewise-constant rates $\bm{R}_t = \log \bm{Q}_t / \Delta t$ over intervals of length $\Delta t = 1$.

Score Entropy Discrete Diffusion (SEDD)~\citet{lou2024discrete} developed this perspective systematically, defining rate matrices for uniform, absorbing, and general transition processes and deriving training objectives directly in continuous time.
The continuous-time viewpoint was also advanced independently by~\citet{sun2023scorebased} and~\citet{campbell2024discrete}, with the latter connecting CTMCs to discrete flow matching.
 
\paragraph{Advantages and trade-offs.}
The CTMC formulation offers several theoretical and practical advantages.

\emph{Unified objective derivation.}
Working in continuous time allows the ELBO (or its equivalent) to be derived as an integral rather than a sum, which often leads to simpler and more interpretable expressions.
For masked diffusion, this integral form reveals that the training objective is invariant to the functional form of the noise schedule beyond its endpoints~\citep{sahoo2024simpleeffect, shi2024simplified}, a result that is less transparent in the discrete-time derivation.

\emph{Flexible step discretization at inference.}
Because the model is trained in continuous time, it is not tied to a fixed number of steps $T$.
At inference time, one can choose any discretization of $[0, 1]$ into $T'$ steps (where $T'$ may be much smaller than the effective $T$ used during training), trading off quality against speed without retraining.
This decoupling of training and inference granularity is a key practical benefit, enabling adaptive step schedules and accelerated sampling~\citep{lou2024discrete, campbell2024discrete}.

\emph{Connection to flows and optimal transport.}
Continuous-time discrete processes connect naturally to discrete flow matching~\citep{campbell2024discrete, davis2024categorical} and optimal-transport formulations~\citep{wu2023formulating, unknown2025cateschrbrid}, enabling the transfer of tools and theory from these neighboring fields.
The Ehrenfest process~\citet{cataldo2024bridging} provides a concrete bridge between discrete and continuous state spaces by modeling binary diffusion through a physical jump process.

The main trade-off is \emph{implementation complexity}: matrix exponentials are expensive for large $K$, and numerical integration of the Kolmogorov equation requires care.
In practice, most CTMC-based models sidestep this by working with rate matrices that have special structure for which $\exp(t\bm{R})$ is available in closed form~\citep{lou2024discrete, campbell2024discrete}.
For absorbing-state rates, the matrix exponential reduces to the same simple mixture formula as in the discrete-time setting, eliminating any computational overhead.

\subsection{Score/Ratio-Based Characterization of Reverse Transitions}

\paragraph{From explicit posteriors to ratio functions.}
The formulations in Sections~\ref{subsec:form_d3pm}-\ref{subsec:continuous_time_discrete_diffusion} all ultimately require specifying the reverse transition $p_\theta(x_{t-1} \mid x_t)$, typically via the predict-$x_0$-then-posterior strategy.
An alternative approach, developed by \citet{meng2022concrete} and \citet{lou2024discrete}, characterizes the reverse process not by predicting $x_0$ explicitly but by learning \emph{ratio} or \emph{score} functions that capture the relative probabilities of reverse transitions directly.

For a CTMC with rate matrix $\bm{R}_t$, the time-reversed process has rate matrix:
\begin{equation}\label{eq:reverse_rate}
  [\overleftarrow{\bm{R}}_t]_{jk} \;=\; \frac{q_t(k)}{q_t(j)}\,[\bm{R}_t]_{kj}, \qquad j \neq k,
\end{equation}
where $q_t(j) = q(x_t = j)$ is the marginal probability of state $j$ at time $t$.
The ratio $q_t(k)/q_t(j)$ is the \emph{concrete score}~\citep{meng2022concrete}: it plays the same role for categorical diffusion that the Stein score $\nabla_{\bm{z}} \log q_t(\bm{z})$ plays for continuous diffusion, characterizing how the data distribution shapes the reverse dynamics.
SEDD~\citep{lou2024discrete} defines a \emph{score entropy} loss that trains a network $s_\theta(x_t, t)$ to estimate these ratios for all pairs of states, and shows that the resulting model admits an exact likelihood via a change-of-variables identity for CTMCs.
The concrete score matching framework~\citep{meng2022concrete} arrives at a similar objective from a different derivation, defining a score function on the simplex of categorical distributions and training it via a denoising score-matching loss.
 
\paragraph{Connection to transition-matrix and CTMC views.}
The ratio-based perspective is not a separate model family but an alternative \emph{parameterization} of the same reverse process.
To see this, note that the predict-$x_0$ parameterization implicitly computes the ratios in Eq.~\ref{eq:reverse_rate}: given a predicted clean distribution $\hat{p}(x_0 \mid x_t)$, one can compute the marginal $\hat{q}_t(k) = \sum_{x_0} q(x_t{=}k \mid x_0)\,\hat{p}(x_0 \mid x_t)$ and form the ratio.
Conversely, the ratio function can be used to recover the reverse transition probabilities in discrete time via Eq.~\ref{eq:reverse_posterior}.
The practical implication is that the choice between predict-$x_0$, predict-logits, and predict-ratios is a question of \emph{parameterization convenience} rather than fundamental model expressivity.
This equivalence should be read as holding for a sufficiently expressive denoiser under the same factorization assumptions: the parameterizations are interconvertible at the level of the exact reverse transition, but with finite capacity, a fixed per-position factorization, and approximate optimization, the choice does affect the induced inductive bias, the tightness of the variational bound, and optimization stability (Section~\ref{subsec:train_score}), as the SEDD and MD4 results below illustrate.
Different parameterizations have different inductive biases: predicting $x_0$ directly ties the model to a classification-style output, while predicting ratios allows the model to express relative preferences between states without committing to an absolute distribution.
SEDD~\citet{lou2024discrete} found that the ratio parameterization yields tighter likelihood bounds than the $x_0$ parameterization for the same model capacity, particularly under uniform (non-absorbing) corruption.
Target concrete score matching~\citep{unknown2025targconcscor} further refines this approach by providing a holistic framework that unifies several prior score-matching objectives under a single target formulation.

\input{tables/04-unified-view}

\subsection{A Unifying View}

\paragraph{Common structure.}
Despite the diversity of formulations reviewed above, every discrete diffusion model instantiates the same four-component structure.
\emph{(1) Corruption operator}: a forward process that maps clean data $\bm{x}_0$ to noisy data $\bm{x}_t$. 
This is specified by a transition matrix $\bm{Q}_t$ (discrete time) or a rate matrix $\bm{R}_t$ (continuous time), with masking, uniform substitution, structured transitions, and hybrids as specific instantiations.
\emph{(2) Denoiser parameterization}: a neural network that, given $(\bm{x}_t, t)$ and optionally a conditioning signal $\bm{c}$, predicts information about the clean data. 
The output may be a distribution over $x_0$, a set of logits, or a vector of log-ratios, depending on the parameterization choice.
\emph{(3) Training objective}: a loss function derived from a variational bound, a score-matching identity, or a simplified surrogate. 
The objective determines how the denoiser is trained and what aspects of the reverse process it is optimized to approximate.
\emph{(4) Sampler}: an inference algorithm that, starting from the stationary distribution, iteratively applies the learned reverse transitions to produce a clean sample. 
The sampler may follow the reverse chain exactly, use confidence-based remasking, apply guidance, or employ acceleration techniques.
 
\paragraph{Mapping across formulations.}
Table~\ref{tab:formulation_map} summarizes how the major formulation families instantiate each component.
The key takeaway is that the distinctions between formulations lie primarily in the corruption operator and the denoiser parameterization; the training objectives and samplers are often interchangeable or can be mixed across families.
For example, a model trained with a score-entropy loss (ratio parameterization) can be sampled using the same ancestral-sampling or $\tau$-leaping algorithms as a model trained with a cross-entropy loss ($x_0$ parameterization), and vice versa~\citep{lou2024discrete, campbell2024discrete}.

This unifying perspective has several practical implications.
First, it clarifies that the choice of corruption operator (masking vs.\ substitution vs.\ structured) is \emph{orthogonal} to the choice of parameterization (predict $x_0$ vs.\ predict ratios): any combination is in principle valid, and the optimal pairing is an empirical question that depends on the domain, vocabulary size, and computational budget.
Second, it reveals that many apparently distinct models differ only in one or two components: for example, MDLM and SEDD use the same absorbing corruption but differ in parameterization (predict $x_0$ vs.\ predict ratios) and objective (reweighted MLM vs.\ score entropy).
Third, it suggests that future work should explore the \emph{off-diagonal} cells of the table, combinations that have not yet been tried, as a systematic way to discover improved designs.
We defer the detailed discussion of training objectives (component 3) to Section~\ref{sec:training} and inference algorithms (component 4) to Section~\ref{sec:inference}.

%% file: tables/04-unified-view.tex
\begin{table}[t]
\centering
\caption{Mapping of major discrete diffusion formulation families to the four-component structure. "Predict $x_0$" means the network outputs a distribution over clean tokens; "predict ratios" means it outputs log-probability ratios $\log[q_t(k)/q_t(j)]$ for pairs of states.}
\label{tab:formulation_map}
\small
\renewcommand{\arraystretch}{1.15}
\begin{tabular}{@{}p{4.0cm}p{2.5cm}p{2.5cm}p{2.8cm}p{2.8cm}@{}}
\toprule
\textbf{Formulation} & \textbf{Corruption} & \textbf{Parameterization} & \textbf{Typical objective} & \textbf{Typical sampler} \\
\midrule
D3PM \citep{austin2021structured}
  & General $\bm{Q}_t$ (uniform, absorb, structured)
  & Predict $x_0$
  & Discrete ELBO (KL per step)
  & Ancestral sampling \\
\midrule
Multinomial Diffusion \citep{hoogeboom2021argmaxflows}
  & Uniform interpolation toward uniform prior
  & Predict $x_0$ / logits
  & Reweighted VLB
  & Ancestral sampling \\
\midrule
MDLM / MD4 \citep{sahoo2024simpleeffect, shi2024simplified}
  & Absorbing (masking)
  & Predict $x_0$
  & Continuous-time ELBO $\to$ reweighted MLM
  & Ancestral / confidence remasking \\
\midrule
SEDD \citep{lou2024discrete}
  & CTMC (uniform or absorb rate matrix)
  & Predict ratios / scores
  & Score entropy
  & $\tau$-leaping / analytic reverse \\
\midrule
Discrete Flow Matching \citep{campbell2024discrete}
  & CTMC / interpolant
  & Predict velocity / $x_0$
  & Flow-matching loss
  & Euler / midpoint integration \\
\midrule
Categorical FM \citep{davis2024categorical}
  & Geodesic on statistical manifold
  & Predict $x_0$ / velocity
  & Fisher--Rao flow matching
  & Euler integration on simplex \\
\midrule
Discrete Interpolants \citep{campbell2025discrete}
  & Interpolation between $x_0$ and noise
  & Predict $x_0$
  & Interpolant matching
  & Ancestral / $\tau$-leaping \\
\bottomrule
\end{tabular}
\end{table}

%% file: sections/06-training.tex
\section{Training Objectives and Parameterizations} \label{sec:training}

%
This section systematizes how discrete diffusion models are trained, proceeding from the most principled objective to the most practical. 
Section~\ref{subsec:train_elbo} develops the discrete ELBO and likelihood-based training; Section~\ref{subsec:train_simplified} shows how it simplifies to the reweighted denoising losses used in practice; and Section~\ref{subsec:train_score} covers the complementary score- and ratio-matching objectives. 
Section~\ref{subsec:train_param} then discusses what the reverse network should predict, Section~\ref{subsec:train_cond} treats conditioning and guidance introduced at training time, and Section~\ref{subsec:train_recipes} collects practical recipes that determine whether these objectives train stably at scale.

\input{tables/05-training-objectives}

\subsection{Discrete ELBO and Likelihood Training} \label{subsec:train_elbo}
\noindent

%
Likelihood-based training provides the most principled starting point for discrete diffusion models. 
For a discrete-time forward Markov chain, the standard variational decomposition can be written, equivalently, as a per-sample lower bound on $\log p_\theta(x_0)$:
\begin{equation}
\log p_\theta(x_0)
\ge
\mathcal{L}_{\mathrm{rec}}
-
\sum_{t=2}^{T}\mathcal{L}_t
-
\mathcal{L}_{\mathrm{prior}},
\end{equation}
where
\begin{equation}
\mathcal{L}_{\mathrm{rec}}
:=
\mathbb{E}_{q(x_1\mid x_0)}
\!\left[\log p_\theta(x_0\mid x_1)\right],
\end{equation}
\begin{equation}
\mathcal{L}_t
:=
\mathbb{E}_{q(x_t\mid x_0)}
\!\left[
\mathrm{KL}\!\left(
q(x_{t-1}\mid x_t,x_0)\,\|\,p_\theta(x_{t-1}\mid x_t)
\right)
\right],
\end{equation}
\begin{equation}
\mathcal{L}_{\mathrm{prior}}
:=
\mathrm{KL}\!\left(q(x_T\mid x_0)\,\|\,p(x_T)\right).
\end{equation}
This form makes two points explicit. 
First, the training problem is governed by a sequence of reverse-time matching terms, together with a reconstruction term and a terminal prior-matching term. 
Second, tractable marginals $q(x_t\mid x_0)$ are indispensable in practice: once they are available in closed form, the posterior $q(x_{t-1}\mid x_t,x_0)$ often becomes tractable as well, which turns the objective into a denoising problem with analytically specified weights.

For masked diffusion, the above structure simplifies dramatically. 
In MD4, the continuous-time objective is written exactly as
\begin{equation}
\mathcal{L}_\infty
=
\int_{t(1)}^{1}
\frac{\alpha_t'}{1-\alpha_t}\,
\mathbb{E}_{q(x_t\mid x_0)}
\!\left[
\delta_{x_t,\texttt{m}}\,x_0^\top \log \mu_\theta(x_t,t)
\right]dt,
\end{equation}
where $\mu_\theta(x_t,t)$ denotes the model's predicted categorical distribution over the clean token (the $x_0$-prediction $\bm{\pi}_\theta$ of Section~\ref{sec:background}), and $\delta_{x_t,\texttt{m}}$ restricts the loss to masked positions. 
Here $t(1)$ is the lower endpoint of the time grid (the smallest noise level at which the loss is evaluated), so the integral runs over the corruption interval $[t(1),1]$; $\alpha_t'=\mathrm{d}\alpha_t/\mathrm{d}t<0$ since the survival probability $\alpha_t$ decreases in $t$, which makes the weight $\alpha_t'/(1-\alpha_t)$ negative and $\mathcal{L}_\infty$ itself negative (it is an expected log-likelihood).
After taking $t(1)\to 0$, the ELBO reduces to $-\mathcal{L}_\infty$ up to the convention adopted in the paper~\citep{shi2024simplified}; equivalently, $-\mathcal{L}_\infty \le \log p_\theta(x_0)$ is a variational lower bound, and training minimizes $\mathcal{L}_\infty$ (i.e.\ maximizes the bound). 
This formulation sharply clarifies why masked diffusion is so closely tied to reweighted cross-entropy training: only masked positions contribute, and the schedule enters through the scalar factor $\alpha_t'/(1-\alpha_t)$. 
MDLM reaches a closely related conclusion from a different derivational route, showing that the continuous-time ELBO collapses to a mixture of masked-language-modeling losses and is invariant to the functional form of the noise schedule beyond its endpoints~\citep{sahoo2024simpleeffect}.
Taken together, these papers make it clear that the ELBO in masked discrete diffusion is not merely tractable; in important cases it is substantially simpler than its general Markov-chain presentation suggests.

Plaid~\citep{shi2024likelihoodba} provides the strongest case for taking likelihood training seriously as an end in itself. 
Rather than abandoning the full variational objective, it optimizes a VLB throughout training and evaluation, uses a learned noise schedule, and shows that diffusion language models exhibit clean scaling trends under maximum-likelihood training. 
At the same time, it also makes the cost of this commitment plain: principled likelihood optimization is feasible, but much more compute-hungry than standard autoregressive training. 
This is precisely why many discrete diffusion papers prefer simplified denoising losses in practice, even when the ELBO remains the conceptual starting point.

Recent work has therefore focused on narrowing the gap between variational rigor and optimization convenience rather than treating them as mutually exclusive. 
\citet{unknown2025demydiffobje} derive a family of increasingly tight time-dependent lower bounds and show that common reweighted diffusion losses can be interpreted as weighted sums of improved variational bounds rather than as purely heuristic surrogates. 
\citet{unknown2025effiperpboun} extend the theory further for discrete CTMCs, deriving cross-entropy/KL identities that lead to a tighter and cheaper perplexity upper bound $J_2$, thereby reducing one of the main practical barriers to strict likelihood-style evaluation in large-vocabulary settings.

A separate line of work enriches the ELBO itself by introducing latent variables. 
\citet{unknown2025varimaskdiff} augment masked diffusion with a global latent variable $z$ and obtain a variational NELBO with an additional KL regularizer, using the latent to capture inter-token dependencies that factorized token predictions cannot express directly. 
\citet{unknown2026variautodisc} extend this idea into a hierarchical variational construction, proposing a "Double ELBO" together with KL annealing to stabilize learning and reduce posterior collapse. 
These models preserve the likelihood-based perspective, but relax the assumption that the reverse model should be purely token-factorized.

At the same time, the literature also contains a clear counterpoint to strict ELBO optimization. 
\citet{unknown2025simpdenodiff} show that, for Uniform State Diffusion Models, a highly simplified denoising objective can scale more favorably and yield better generation quality even when the formal variational bound becomes looser. 
The resulting picture is therefore not one of replacement, but of tension: ELBO-based training remains the cleanest route to likelihood estimation and principled comparison, while simplified objectives often provide a better optimization interface when generation quality and scalability are the primary concerns.

\subsection{Simplified Denoising Objectives} \label{subsec:train_simplified}
In practice, the dominant training view is to sample a corruption level, corrupt the clean sequence, and train the model to recover the original tokens from the noisy input. 
This perspective is broad enough to cover masked diffusion, uniform-state corruption, and several hybrid designs that operate in continuous latent spaces but decode with categorical targets. 
Its appeal is obvious: compared with the full ELBO, it exposes a single supervised signal that is easy to implement, easy to batch, and often better aligned with the reverse model one actually uses at inference time.

MDLM~\citep{sahoo2024simpleeffect} is the clearest example of how such simplification can still remain theoretically grounded. 
It shows that the continuous-time ELBO of masked diffusion is equivalent to a weighted masked-language-modeling objective, thereby turning a generative diffusion objective into a randomized denoising problem that can be trained with standard cross-entropy machinery.
\citet{unknown2025demydiffobje} then systematize this viewpoint by deriving explicit reweighting schemes and proving that monotonically increasing weights still correspond to valid variational bounds. 
Their analysis is useful because it reframes weighting not as an afterthought, but as a design axis with both theoretical and empirical consequences.

Several works simplify the denoising target even further by restricting supervision to genuinely corrupted positions. 
\citet{unknown2025repadiscdiff} derive, from a route-and-denoise perspective, a reweighted cross-entropy objective that is evaluated only on noisy token positions, and they find linear reweighting to work particularly well in practice. 
\citet{unknown2025simpdenodiff} push this principle to an extreme in Uniform State Diffusion Models: the loss is applied only to noise-replaced tokens, while an additional anti-uniform sharpening term is introduced to prevent the predictions from drifting toward an overly flat distribution. 
The conceptual shift here is small but important. 
Once the model is only penalized where corruption actually destroyed information, the objective becomes much closer in spirit to masked language modeling than to a generic per-position reconstruction loss.

A related but distinct simplification appears in \citet{unknown2023cheabettdiff}. 
Instead of keeping a continuous diffusion loss and then rounding back to discrete tokens, they learn a projection to vocabulary space and optimize a token-level cross-entropy directly. 
This removes the mismatch between a continuous reconstruction target and a discrete generation goal, and it also eliminates the need for heuristic rounding procedures. 
In effect, it is a simplified denoising objective designed not for masked diffusion proper, but for continuous latent diffusion models that ultimately serve discrete data.

One of the most striking conceptual simplifications comes from \citet{shi2025maskeddiffus}. 
They show that, for masked diffusion, the NELBO can be rewritten purely in terms of the number of masked tokens $n$, together with a natural $1/n$ weighting. 
Under this view, the scalar time variable is not fundamental: what matters is the masking pattern itself and, in particular, how many tokens remain unresolved. 
This is a strong result because it explains why time conditioning can sometimes be weakened or even omitted without destroying the training signal.

Even so, noise-level conditioning remains useful in many practical settings. 
When a single denoiser must operate across corruption levels ranging from almost clean to nearly fully destroyed, explicit time embeddings often make optimization easier and improve calibration across steps. 
The key point is not that time conditioning is always necessary, but that its role is contingent.
In some models it genuinely helps the network adapt across noise regimes; in others, especially masked models with sufficiently informative corruption patterns, the dominant variable is not continuous time but the combinatorial structure of which tokens are missing. 
This is precisely why simplified denoising objectives have become the default engineering choice across much of the literature: they preserve the central denoising semantics while discarding parts of the variational machinery that are not always needed for good optimization.

\subsection{Score- and Ratio-Matching Objectives} \label{subsec:train_score}
A second family of objectives learns reverse dynamics through scores or probability ratios rather than directly through a categorical denoising distribution. 
In discrete state spaces, this amounts to modeling relative plausibilities of local transitions, which can be more natural for CTMC-style sampling rules than predicting a normalized clean-token distribution outright. 
Conceptually, this differs from the denoising objectives of Section~\ref{subsec:train_simplified}, even though the two views can coincide under specific parameterizations.

For masked diffusion, MD4 gives an exact bridge between the two. In the notation of \citet{shi2024simplified}, for a mask state and any non-mask class $j$,
\begin{equation}
s_\theta(m,t)_j
=
\frac{\alpha_t}{1-\alpha_t}\,\mu_\theta(\texttt{m},t)_j,
\qquad j\neq m.
\end{equation}
This identity is more than a cosmetic reparameterization. 
It shows that, in masked diffusion, score-based training can collapse back to a mean-parameterized denoising objective once the reverse model is constrained to be compatible with the forward process. 
The same paper also points out the corresponding failure mode: if the score model is left unconstrained, as in some earlier formulations, the learned reverse process need not remain consistent with the forward dynamics, which can lead to instability and poorer behavior in practice.

This observation explains why score- and ratio-based methods have often looked more attractive in theory than in implementation. 
They are closely tied to likelihood and continuous-time sampling, but a naive parameterization can be both expensive and fragile when the vocabulary is large. 
\citet{unknown2025effiperpboun} address this bottleneck by showing that ratio matching can be implemented efficiently through a denoising cross-entropy objective in CEDD. 
In that sense, their contribution is not just a faster estimator; it is a practical reconciliation of two training philosophies that had previously seemed to pull in opposite directions.

The practical trade-off therefore becomes clearer. 
Compared with simplified denoising, score- and ratio-matching can provide a tighter link to continuous-time likelihood theory and reverse-process construction. 
Compared with direct ELBO optimization, they can be more operational once expressed in a scalable parameterization. 
The remaining difficulty is not the idea itself, but the form in which it is trained: unconstrained scores are elegant but brittle, whereas denoising-compatible ratio parameterizations are less free-form but much easier to scale.

\subsection{Parameterizations of the Reverse Model} \label{subsec:train_param}
A central design choice in discrete diffusion is what the reverse network should predict. 
One option is to predict one-step reverse transitions directly, i.e., to model $p_\theta(x_{t-1}\mid x_t)$ as the primitive object. 
The alternative, which has become much more common, is to predict the clean sample or its categorical distribution and then obtain the reverse step from the known structure of the forward process. 
In discrete settings this latter choice is especially appealing, because the corruption mechanism is often simple enough that much of the reverse transition can be reconstructed analytically once a prediction of $x_0$ is available.

MDLM~\citep{sahoo2024simpleeffect} exemplifies this predict-$x_0$ perspective through its SUBS parameterization. 
The construction is deliberately structured: already revealed tokens are carried forward, remasking is forbidden, and the model is only asked to make meaningful predictions where information is still missing. 
These constraints simplify the objective analytically and, just as importantly, make the reverse process easier to cache and reuse at inference time. 
The paper further shows that explicit time conditioning can be removed in this setup, which is a particularly strong statement about how much of the reverse dynamics is already encoded by the masking pattern itself.

Other papers broaden the parameterization space rather than narrowing it. 
\citet{unknown2025repadiscdiff} expose a hidden route-and-denoise decomposition, in which each token is first routed, kept noisy or selected for denoising, and only then decoded through a predict-$x_0$ mechanism. 
This introduces an additional latent decision into the reverse model, and it also opens the door to using different routing behavior at training and sampling time. 
\citet{unknown2025varimaskdiff} and \citet{unknown2026variautodisc} add yet another layer by injecting a global continuous latent variable $z$ through adaptive layer normalization. 
In both cases, the local token predictor remains factorized conditional on $z$, while marginalization over $z$ restores some degree of global dependence across positions. 
The result is a reverse model that remains operationally simple at the token level but is no longer purely local in its representational capacity.

A related question is whether the network should predict continuous vectors or categorical logits. 
Plaid~\citep{shi2024likelihoodba} argues strongly for the latter. 
Rather than forcing the model to reproduce precise embedding vectors and then infer token identities indirectly, it reparameterizes the denoiser to output a softmax over the vocabulary.
This uses model capacity more efficiently and improves both likelihood and generation quality.
\citet{unknown2023cheabettdiff} reach a similar conclusion in a different setting: replacing continuous $\ell_2$ reconstruction and nearest-neighbor rounding with direct logit prediction substantially improves training stability and allows pretrained masked language models to be incorporated more naturally. 
In discrete diffusion, then, direct categorical modeling is not merely convenient; it is often the cleaner parameterization.

The same predict-$x_0$ output can also be repurposed into other reverse-time quantities. 
\citet{unknown2025effiperpboun} keep the model in the predict-$x_0$ regime during training, but convert the resulting predictions into scores or ratios at sampling time using known conditional ratios from the matrix exponential. 
This is an instructive design choice: instead of learning a harder object directly, the model learns a simpler one and lets the analytic structure of the forward process do the rest.

Finally, self-conditioning has emerged as an effective auxiliary parameterization trick rather than a separate modeling family. 
In Plaid, the model feeds its own current estimate back into later denoising evaluations, which improves both generation quality and held-out likelihood~\citep{shi2024likelihoodba}. 
Conceptually, this gives the reverse model access to its own evolving hypothesis about the clean sample, allowing refinement without changing the basic sampling framework. 
Latent-augmented denoisers such as VMD and VADD play a related role at a larger scale: they supplement token-level predictions with an auxiliary channel that carries global information across steps. 
Seen this way, reverse-model parameterization is less about choosing a single target and more about deciding which parts of the reverse process should be learned explicitly and which should be provided by structure.

\subsection{Conditioning and Guidance During Training} \label{subsec:train_cond}
Conditioning in discrete diffusion can be introduced either directly in the training architecture or indirectly through guidance mechanisms applied at inference time. 
The surveyed literature is notable in that it obtains a large amount of controllability without relying exclusively on dedicated conditional pretraining. 
This distinguishes diffusion from many conditional autoregressive pipelines, where the architecture often encodes the condition more explicitly from the outset.

Plaid~\citep{shi2024likelihoodba} is the clearest example of this inference-time view. 
It is trained unconditionally, yet supports a wide range of zero-shot control settings through token guidance. 
The key idea is to manipulate the denoiser's own token-level predictions during sampling so as to realize span constraints, lexical conditions, negation, and other compositional requirements. 
This is attractive for survey purposes because it shows that conditioning in discrete diffusion need not be reduced to a choice between prefix prompting and classifier guidance; the denoising model itself can expose enough structure to support surprisingly flexible constraint injection after pretraining.

A more explicit guidance route is taken by \citet{unknown2023cheabettdiff}, who use plug-and-play classifier guidance in continuous latent space for controllable text generation. 
Their experiments span semantic content, part-of-speech patterns, syntax trees, syntax spans, and length. 
Although the implementation is different from Plaid's token guidance, both works illustrate the same broader principle: the reverse process can be steered by an auxiliary signal that is external to the unconditional denoiser, provided that the denoiser remains differentiable and interpretable enough for that signal to act on.

An analogue of classifier-free guidance is explored by \citet{unknown2025simpdenodiff}. 
Their unsupervised variant constructs an unconditional branch by replacing conditioning inputs with random tokens, making it possible to recover CFG-like behavior without paired conditional data. 
This line is especially relevant for discrete diffusion because it turns "dropout conditioning" into a training-time mechanism whose purpose is realized mainly at inference: by learning both conditional and weakened-condition behaviors within one model, the sampler gains a simple control knob for trading off fidelity against adherence to the requested condition.

More broadly, the conditioning mechanisms used in this literature can be grouped by where the condition enters. 
In some cases it is encoded as observed tokens or spans, in some it is supplied through an external classifier or gradient signal, and in others it is represented implicitly through conditional versus weakened-conditioning branches. 
The common feature is iterative intervention. 
Since diffusion refines samples over multiple denoising steps, guidance is not a one-shot adjustment but a repeated modulation of the reverse trajectory. 
That iterative character is one reason discrete diffusion can accommodate diverse forms of control even when the underlying denoiser is trained in a comparatively generic way.

\subsection{Practical Training Recipes} \label{subsec:train_recipes}
In discrete diffusion, training recipes matter nearly as much as the nominal objective. 
The first recurring theme is that the noise schedule acts as a curriculum. 
It determines how corruption accumulates and the order in which information tends to be reconstructed during generation. 
MD4 shows that schedule choice affects optimization variance and discretization quality even when the underlying continuous-time objective is invariant to the schedule shape~\citep{shi2024simplified}. 
\citet{unknown2025maskdiffmode} make this point especially concrete by optimizing the sampling schedule through an energy-minimization view grounded in discrete optimal transport, using a Beta-CDF parameterization that can be tuned post hoc without retraining. 
This is a particularly elegant result because it turns schedule design into a lightweight yet principled control interface.

Several papers go further and design the schedule to induce a specific generation order. 
\citet{wang2023improving} mask common token categories first, which yields a corresponding common-first reverse process. 
\citet{unknown2023cheabettdiff} propose a linguistically informed soft-masking policy based on TF-IDF and information entropy, so that more informative words are corrupted earlier and therefore reconstructed later. 
These choices are not just cosmetic. 
They encode beliefs about which parts of the sample should be fixed early and which should remain unresolved until richer context is available.

A second major theme is variance control and optimization stability. 
\citet{shi2024simplified} use antithetic time sampling, cosine discretization, an $\epsilon$-shifted schedule, and EMA with decay $0.9999$. 
\citet{unknown2025brinstabdiff} provide the most explicit variance decomposition, separating masking-pattern noise, masking-rate noise, and data noise, and then reducing them with P-POTS timestep sampling and MIRROR antithetic masking. 
\citet{unknown2025demydiffobje} show that a simple constant weighting across noise levels can itself serve as a strong and stable recipe. 
\citet{unknown2025repadiscdiff} combine EMA, checkpoint averaging, label smoothing, and adaptive top-$k$ routing, demonstrating that strong results can coexist with extremely short sampling chains. 
Read together, these works suggest that the practical bottleneck in masked diffusion is often not expressivity but the stochasticity of the training estimator.

Masking policy is another practical lever that has received increasingly principled treatment. 
\citet{unknown2025tuniimplregu} argue that the masked-diffusion objective implicitly contains a regularizing effect arising from unidentifiable masked inputs, and use this analysis to recommend a narrower, signal-favorable masking-rate window. 
This is a useful example of how seemingly low-level training heuristics can be justified by decomposing the objective into distinct signal and noise regimes.

Numerical precision also matters more than one might expect. 
\citet{shi2025maskeddiffus} show that float32 Gumbel-based categorical sampling induces a truncated distribution and thereby lowers the effective temperature in masked diffusion generation. 
Their recommendation to use float64 for this operation is practically important, because it reveals that standard mixed-precision recipes may interact with discrete sampling in nontrivial ways. 
In other words, discrete diffusion inherits many large-scale training habits from contemporary language modeling, but not all of them transfer unchanged.

Plaid contributes a complementary set of large-scale lessons~\citep{shi2024likelihoodba}. 
It learns the noise schedule jointly with the model, explicitly tying schedule optimization to variance reduction, and its scaling analysis suggests that compute-optimal diffusion language models should be smaller and trained for longer than their autoregressive counterparts. 
This is one of the clearest cases where the optimal training recipe is not simply "copy the AR recipe and change the objective"; the diffusion setting induces its own allocation of model size, training length, and denoising difficulty.

Finally, alignment and post-training remain comparatively underexplored, but the existing results already suggest a plausible roadmap. 
At a high level, supervised fine-tuning can be cast as denoising under task-specific partial observations, while preference optimization can be applied to completed samples or, more ambitiously, to intermediate refinement trajectories. 
Guidance-based mechanisms provide an additional lightweight option: some forms of alignment may be easier to implement by steering the sampler than by re-optimizing the denoiser from scratch. 
For now, the literature offers more ingredients than settled recipes, but those ingredients already indicate that post-training for diffusion models will likely involve both objective design and sampler design, rather than treating the denoiser alone as the whole system.

%% file: tables/05-training-objectives.tex
\begin{table}[t]
\centering
\caption{Families of training objectives for discrete diffusion. All can be paired with the predict-$x_0$, predict-logits, or predict-ratio parameterizations of Section~\ref{subsec:train_param}; the table lists the most common pairing.}
\label{tab:training_objectives}
\small
\setlength{\tabcolsep}{4pt}
\renewcommand{\arraystretch}{1.15}
\begin{tabular}{@{}p{3.5cm}p{4cm}p{4.5cm}p{3.4cm}@{}}
\toprule
\textbf{Objective} & \textbf{What it optimizes} & \textbf{Typical parameterization} & \textbf{Trade-off} \\
\midrule
Discrete ELBO & Variational bound on $\log p_\theta(\bm{x}_0)$ & Predict $x_0$ & Principled likelihood; compute-hungry \\
\midrule
Simplified denoising & Reweighted cross-entropy on corrupted positions & Predict $x_0$ & Easy to scale; bound can be loose \\
\midrule
Score / ratio matching & Concrete score / probability ratios & Predict ratios & Tight CTMC likelihood; fragile if unconstrained \\
\midrule
Flow matching & Interpolant velocity between noise and data & Predict velocity / $x_0$ & Flexible step count; integration error \\
\bottomrule
\end{tabular}
\end{table}

%% file: sections/07-inference.tex
\section{Inference Algorithms and Efficiency} \label{sec:inference}
\noindent
Discrete diffusion inference can be viewed as a policy layer on top of the learned denoiser. 
In autoregressive models, generation follows a fixed append-and-cache order. 
In masked and other discrete diffusion models, the reverse process exposes several additional control variables. 
The sampler can choose which time points to visit, which tokens to reveal or revisit, which positions are grouped into a block, how external rewards or constraints steer the path, and which parts of the Transformer computation are reused or skipped. 
These choices expand the design space, but they also introduce characteristic failure modes: aggressive parallel updates may break dependencies, confidence scores can be miscalibrated, guidance can alter the reverse dynamics, and nominal NFE reductions may not translate into wall-clock speed.

The discussion follows the main degrees of freedom exposed by the reverse process. 
It begins with ancestral sampling and step schedules (Section~\ref{subsec:inf_ancestral}), where the central issue is how the chain is discretized, corrected, or shortened. 
It then moves from time to positions: confidence-based remasking (Section~\ref{subsec:inf_confidence}) decides which tokens to trust, reopen, or advance in parallel. 
Semi-autoregressive and block updates (Section~\ref{subsec:inf_block}) change the dependency structure itself, while guidance and constraints at inference (Section~\ref{subsec:inf_guidance}) add external objectives, from rewards and search to hard validity checks and posterior conditioning. 
The section closes with acceleration (Section~\ref{subsec:inf_accel}), where these algorithmic choices meet distillation, cache reuse, selective computation, speculative verification, and hardware-aware serving.

\input{tables/06-inference-methods}

\subsection{Ancestral Sampling and Step Schedules} \label{subsec:inf_ancestral}

\paragraph{Baseline reverse-time sampling.} 
A basic axis of discrete diffusion inference is how the reverse process is traversed in time. 
Even when the denoiser is fixed, a discrete diffusion sampler decides how many states to visit, where those states lie, and whether each reverse update is a simple local step or a corrected numerical approximation. 
Early discrete image-generation work illustrates this point. 
Discrete Predictor-Corrector samplers augment masked-token reverse updates with correction steps and show that sampler design can change sample quality at fixed model scale; for example, DPC variants improve ImageNet FID substantially under different NFE budgets~\citep{chang2023discrete}.
Informed correctors develop the same principle for discrete diffusion: the reverse chain can be improved by adding correction dynamics rather than simply replaying the base transition~\citep{unknown2025infocorrdisc}. 
These methods separate two questions that are often conflated: how good is the learned denoiser, and how well does the inference procedure use it?

Other works improve the local numerical update rather than only removing steps. 
High-order solvers for discrete diffusion treat sampling as numerical integration over discrete-state dynamics, yielding much better likelihood or perplexity at low NFE than Euler- or tau-leaping-style updates~\citep{unknown2025fastsolvdisc}. 
Neural sampler formulations such as MDNS cast masked diffusion sampling as stochastic optimal control, shifting the objective from matching a hand-designed reverse chain to learning or optimizing a path policy~\citep{unknown2025mdnsmaskdiff}. 
Reversible diffusion decoding and approximate joint sampling extend this path-centric view by asking whether the sampler can use additional reversible or joint structure to reduce wasted updates~\citep{unknown2025revediffdeco,unknown2025enabapprjoin}. 
These papers suggest that discrete diffusion inference is closer to controlled numerical simulation than to a single canonical decoding loop.

\begin{figure}[t]
  \centering
  \includegraphics[width=\linewidth]{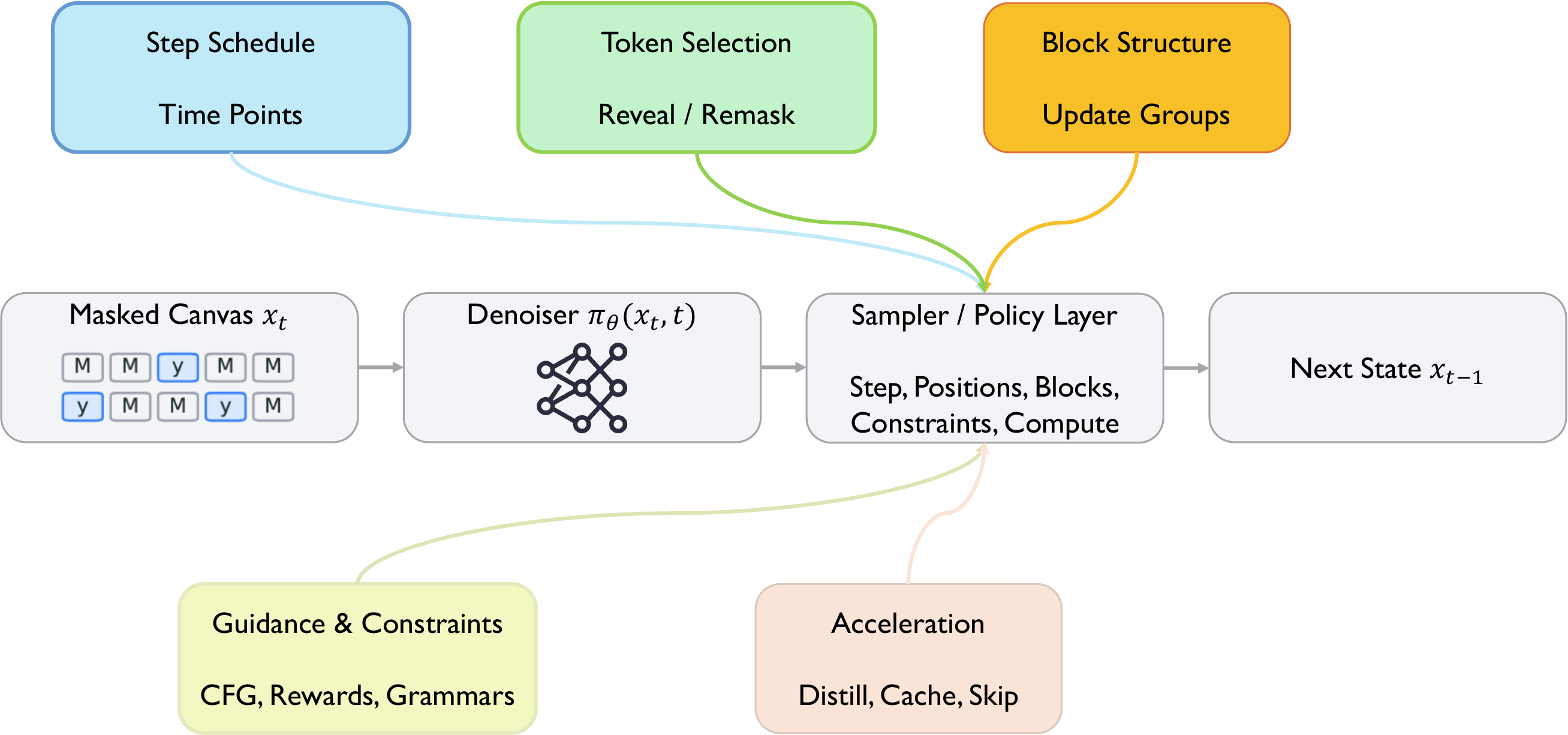}
  \caption{Inference as a policy layer over a fixed denoiser. Step schedules, token selection, block structure, guidance, and acceleration are composable sampler decisions that affect quality, validity, and runtime.}
  \label{fig:inference-design-space}
\end{figure}

%
\paragraph{Step schedules and effective computation.} 
A complementary line reduces the number of useful denoiser evaluations by changing the time representation. 
Discrete Non-Markov Diffusion Models pre-sample token transition times, so the denoiser is called only when a token can actually change; in translation experiments, this collapses thousand-step nominal chains into a few dozen effective denoiser calls while preserving BLEU~\citep{unknown2024fastsampdisc}. 
Few-shot temporal pruning reaches a similar conclusion empirically: many text-diffusion timesteps are redundant, and a small validation set can identify short schedules that outperform naive uniform pruning~\citep{unknown2024fewstempprun}. 
The older de-randomization formulation and its later Non-Markov version can be understood as part of the same general pattern: discrete trajectories often contain idle regions, and the inference algorithm need not treat all nominal steps uniformly~\citep{unknown2025fastsampdera}.

Schedule design also reappears in modern masked diffusion language models. 
Jump Your Steps optimizes where to place a fixed number of denoising steps, while Plan for Speed uses dilated schedules to reduce the number of decoding rounds under a structured unmasking plan~\citep{unknown2025jumpyourstep,unknown2025planspeedila}. 
Path Planning for masked diffusion generalizes this idea from time alone to a planner that decides which positions to unmask or remask along the reverse path, with applications ranging from language to biological sequences~\citep{unknown2025pathplanmask}. 
This creates a close connection between schedules and token-selection methods: schedules define the temporal scaffold, but effective schedules often encode assumptions about which positions will be resolved together.

This literature also clarifies that "fewer steps" can mean several different things. 
A method may use fewer nominal timesteps, fewer actual denoiser calls, a higher-order update with lower error per call, or a path planner that avoids unproductive parts of the trajectory. 
These notions have different implications for likelihood, sample quality, and serving latency. 
Comparisons are therefore more informative when they report not only NFE, but also whether the method changes the transition family, uses extra corrector networks or validators, and preserves quality under the same model and hardware.

\subsection{Confidence-Based Remasking (Iterative Parallel Decoding)} \label{subsec:inf_confidence}

\paragraph{Core confidence-based remasking.} 
Once the state is a partially masked sequence, a central inference axis becomes spatial: which positions are filled, reopened, or trusted at each step. 
This is one of the main freedoms that distinguishes masked diffusion from left-to-right generation. 
Train for the worst, plan for the best highlights this phenomenon: masked diffusion models are trained on hard arbitrary masks, but at inference time a sampler can choose easier subproblems first~\citep{unknown2025traiworsplan}. 
Simple token-ordering rules based on probability margins can substantially improve tasks such as Sudoku and code/math generation. 
These results suggest that confidence is useful because generation order itself is an inference-time resource.

This raises the question of how aggressively tokens can be decoded in parallel. 
Entropy-bounded unmasking links parallel-update risk to predictive entropy and dependence error, choosing more tokens only when the distribution is sufficiently certain~\citep{unknown2025accesampmask}. 
KLASS adds a stability signal: a token is more reliable when it has both high confidence and a stable predicted distribution across consecutive denoising steps~\citep{unknown2025klasklgufast}.
Conditional-independence testing attacks the same issue more directly by asking whether simultaneously decoded positions are conditionally safe to sample together~\citep{unknown2026parasampmask}. 
These methods point to a useful taxonomy of token selection criteria: confidence estimates whether the current top token is plausible, entropy estimates local uncertainty, KL or trajectory stability estimates whether the decision is settled, and independence tests estimate whether parallel acceptance itself is safe.
In multimodal diffusion LLMs, VISAGE extends this token-selection view to visual grounding, using the spatial entropy of cross-attention to downweight visually diffuse candidates and prioritize token commitments with localized image support~\citep{narnaware2026seeinggroundvisualattention}.

Another class of methods allows earlier decisions to be revised. 
Remasking Discrete Diffusion explicitly permits an unmasked token to return to the mask state, converting decoding from a monotone reveal process into an iterative refinement process~\citep{arriola2025remasking}. 
CORE identifies context-brittle tokens by probing their sensitivity to masked-context perturbations and prioritizes them for remasking, providing a context-robust alternative to static confidence~\citep{zhai2026core}.
Self-reflective remasking, backtracking-enhanced samplers, and head-only look-back correction add more targeted mechanisms for reopening suspicious positions or correcting earlier commitments~\citep{unknown2026dontsetttoo,unknown2025sabeeffisamp,unknown2025backpluglook}. 
Local determinism propagation takes a different route: high-confidence anchor tokens make nearby positions more reliable, so the sampler can locally relax thresholds around such anchors rather than globally lowering the bar~\citep{unknown2025accediffllm,unknown2025accedifflang}. 
Foreseeing-movement decoding and multimodal self-correction make a similar point in different settings: the sampler can use predicted future motion or cross-modal consistency to decide which positions deserve another refinement pass~\citep{nie2025decoding,unknown2025traiselfmult}. 
This separates two related controls: unmasking decides when to commit a still-masked position, while remasking decides whether a commitment remains final.

\paragraph{Design knobs for token selection.} 
Several recent papers turn token selection into an explicit learned or optimized policy. 
NI Sampling trains a lightweight neural indicator to identify tokens that can be advanced without changing a reference trajectory, reaching large step reductions when combined with cache~\citep{unknown2026nisampacce}. 
Learning Unmasking Policies formulates the decision as an MDP over confidence, mask status, and time~\citep{unknown2025learunmapoli}. 
Optimizing Decoding Paths scores entire candidate paths using denoising entropy and uses Best-of-N or SMC-style allocation to favor lower-uncertainty trajectories~\citep{unknown2025optidecopath}.
From Bits to Rounds argues that the most confident tokens are often the least informative; useful speedups require exploring medium-confidence, high-information positions rather than greedily resolving easy tokens first~\citep{unknown2025bitsrounpara}. 
This perspective helps explain why confidence-only decoding can reduce rounds while still missing important positions.

Position- and frequency-aware variants broaden the same theme. 
Position Contrastive Guidance corrects systematic positional biases in masked diffusion sampling, while FourierSampler and FreqTS use frequency-domain signals to decide which positions are worth updating~\citep{unknown2025imprsampmask,unknown2025fourunlonona,unknown2026freqfreqtoke}. 
MC-DiT shows the same issue in masked image diffusion, where clean-to-clean reconstruction can improve contextual consistency during masked sampling~\citep{unknown2024mcdicontenha}. 
Sink-token and token-level cross-validation papers show that local token dynamics can also be regularized or verified at inference time~\citep{unknown2025onetokeenou,unknown2025finifirsperf}.
A recurring challenge across this subsection is calibration. 
Token policies are most effective when their scores predict future correctness under the chosen decoding path; model confidence alone often measures ease, not importance, and can therefore produce fast but brittle generation.

\subsection{Semi-Autoregressive and Block Updates} \label{subsec:inf_block}

\paragraph{Block sampling as dependency management.} 
Token-selection policies decide which individual positions to update, but block and semi-autoregressive methods change the dependency topology of generation. 
Block Diffusion provides a representative starting point: it interpolates between autoregressive generation and fully parallel diffusion by making blocks causal across time while allowing bidirectional denoising within the active block~\citep{arriola2025blockdiffusi}. 
This gives a tunable structural compromise. 
Smaller blocks preserve more left-to-right dependency and make the model closer to an AR LM; larger blocks expose more parallelism but increase the burden on the denoiser to model within-block dependencies.

This blockwise perspective has become a practical adaptation path for diffusion LLMs. 
From Next-Token to Next-Block treats an AR LM as the block-size-one limit of block diffusion, then gradually expands the active block while preserving causal context outside it~\citep{unknown2025nextnextprin}. 
ReFusion and Fast-dLLM v2 similarly use block or semi-autoregressive structure to combine diffusion-style local refinement with AR-like progression and cacheable context~\citep{unknown2026refudifflarg,unknown2026fastv2effi}. 
In this view, block diffusion is more than a reduction in parallelism. 
It creates a different conditional factorization, one that can preserve instruction-following and long-context behavior while still permitting multiple tokens to be refined per forward pass.

\paragraph{When block structure helps.} 
Adaptive block methods make the topology state-dependent. 
AdaBlock-dLLM changes block size based on semantic or confidence signals, and Swordsman partitions sequences according to entropy so that easy regions can be decoded more coarsely than hard ones~\citep{unknown2025adabsemadiff,unknown2025sworentradap}. 
Hierarchy Decoding recursively splits masked spans as confident positions are resolved, keeping remaining unresolved positions more sparsely distributed~\citep{unknown2026hierdecotrai}. 
WavefrontDiffusion uses a dynamic schedule that propagates decoding from easier or more reliable regions, improving reasoning-oriented generation by shaping how information moves through the sequence~\citep{unknown2026wavedynadeco}. 
HEX exploits diversity across semi-autoregressive decoding schedules by treating heterogeneous block schedules as latent experts and ensembling their generation paths at test time, showing that committing to a single fixed schedule can leave substantial performance gains untapped~\citep{lee2026testtime}.
Reviving any-subset autoregressive models connects this blockwise view back to older order-agnostic AR formulations, showing that principled parallel sampling and speculation can be interpreted as another point in the same topology design space~\citep{unknown2025revianysauto}.
These methods connect block topology back to token policies, but with a stronger structural prior: the sampler is choosing regions, not only positions.

Length control is another place where semi-autoregressive structure matters. 
Any-order flexible length diffusion, flexible-length text infilling, implicit infilling-length estimation, and bidirectional end-of-sequence control all try to avoid fixing a rigid output length before generation~\citep{arriola2025anyorder,unknown2025flextextinfi,unknown2025difflmscan,unknown2025traibidivari}.
This is particularly important for dLLMs, where a fixed canvas can waste computation on padding or make early end-of-sequence decisions unstable. 
Blockwise generation gives the model a way to expand, stop, or refine local regions without requiring the entire output to be predetermined.

A recurring limitation is dependency error. 
Generation-order theory and ParallelBench show that decoding multiple mutually dependent tokens in the same step can violate the conditional assumptions that make parallel generation attractive~\citep{unknown2025geneordepara,unknown2025parageneorde,unknown2026paraundetrad}. 
Multiverse and related parallelize-and-merge methods show that models sometimes contain latent preferences about parallelization, but exploiting those preferences requires careful merging and verification~\citep{unknown2025multyourlang}. 
Fast and Fluent decoding further warns that overly large windows or poorly structured future context can degrade open-ended fluency~\citep{unknown2025fastfluediff}. 
Thus, blockwise inference can be interpreted as dependency management: it chooses which positions may interact bidirectionally, which positions are committed, and how much future uncertainty the current update is exposed to.

\subsection{Guidance and Constraints at Inference} \label{subsec:inf_guidance}

\paragraph{Logit shaping and soft guidance.} 
Guidance adds external objectives to the reverse process. 
The simplest family adapts classifier-free or classifier-based guidance to discrete states. 
Simple Guidance Mechanisms derives discrete classifier-free and classifier-based guidance rules and shows how the choice of noise process affects editability and repeated refinement~\citep{nisonoff2025simpleguidan}. 
Unlocking Guidance generalizes guidance to continuous-time Markov processes over discrete spaces, using process structure to make rate normalizers tractable across molecules, DNAs, and proteins~\citep{rectorbrooks2025unlocking}. 
Discriminator and training-free molecular guidance works provide related mechanisms for steering autoregressive or discrete diffusion transitions without retraining the base generator~\citep{hoogeboom2022discriminato,unknown2025traiguiddisc}. 
These papers establish that guidance can operate on logits, rates, transition matrices, or proposal distributions, and that tractability depends heavily on the chosen discrete process.

Recent analysis shows that discrete guidance is not only a scalar preference adjustment. 
What Exactly Does Guidance Do in Masked Discrete Diffusion Models analyzes low-dimensional cases and shows how classifier-free guidance can concentrate probability on class-specific support while suppressing overlap regions, with strong guidance producing unstable concentration effects~\citep{unknown2026whatexacdoes}. 
Improving Classifier-Free Guidance identifies a more concrete issue: unnormalized discrete CFG can change the total unmasking rate, not just the relative token preference~\citep{rojas2026imprclasguid}. 
Position-contrastive guidance similarly modifies token scores in ways that affect both order and content~\citep{unknown2025imprsampmask}. 
Guidance in discrete diffusion is therefore better described as modifying the reverse dynamics, including the timing of state changes, rather than only pushing samples toward a condition.

Reward and search-based methods form a second cluster. 
Soft Value-Based Decoding evaluates candidate states under a value function and selects higher-reward denoising paths without requiring differentiable rewards~\citep{unknown2025deriguidcont}.
TreeG extends this idea into explicit tree search over diffusion or flow trajectories, enabling non-differentiable objectives in domains such as symbolic music, molecules, and biological sequences~\citep{unknown2025traiguidbeyo}. 
Importance-weighted SMC provides a more formal inference-time-scaling view: intermediate reward-tilted targets define weights, while proposal design controls variance and efficiency~\citep{unknown2026infescaldisc}. 
These approaches make extra inference compute productive, but their benefits depend on reward reliability, proposal quality, and the cost of evaluating branches.

For dLLMs, guidance increasingly uses internal confidence, verifier feedback, or reward-free references rather than a single external classifier. 
Self-Rewarding SMC uses trajectory-level confidence as an implicit reward~\citep{unknown2025selfsequmont}. 
RFG constructs reward-free guidance from policy-reference logit differences~\citep{unknown2025rfgtestscal}. 
Reward-Weighted Sampling changes token selection order with an external reward model, and IterRef combines reward-guided noising-denoising refinement with a Multiple-Try Metropolis mechanism~\citep{unknown2025rewasampenha,unknown2025effetestscal}. 
Feedback guidance and Prism-style hierarchical search/self-verification further blur the line between guidance and test-time reasoning~\citep{unknown2025feedguiddiff,unknown2025priseffitest}.
The common theme is inference-time allocation: the sampler spends additional computation where reward, confidence, or verification suggests the current trajectory can be improved.

\paragraph{Validity constraints and posterior conditioning.} 
Hard constraints are usually treated separately because the goal is validity, not merely higher reward. 
Constrained Discrete Diffusion projects each denoising distribution toward the KL-nearest distribution satisfying explicit constraints, achieving zero violations in several constrained generation settings~\citep{li2025constrained}. 
Grammar-constrained dLLM decoding checks whether a partial sequence can be completed to a CFG-valid string, while Lookahead-then-Verify samples possible continuations before accepting a token~\citep{unknown2026consdecodiff,unknown2025lookrelicons}. 
DINGO compiles regular constraints into token-level automata and dynamic programs over block positions and automaton states~\citep{suresh2025dingconsinfe}. 
These works illustrate why discrete diffusion is attractive for constrained generation: the model repeatedly exposes a full partially specified sequence, giving the sampler opportunities to enforce global structure before the final output is fixed.

Finally, guidance connects discrete diffusion to inverse problems and posterior sampling. 
G2D2 uses categorical variational distributions and Gumbel-Softmax relaxation to apply gradient guidance with discrete diffusion priors~\citep{unknown2025g2d2graddisc}. 
Split Gibbs, DICE, test-time anchoring, and DDPS adapt discrete diffusion to posterior sampling, controllable editing, and graph/path inverse problems~\citep{unknown2025spligibbdisc,nie2024dicediscrete,unknown2025testanchdisc,unknown2025ddpsdiscdiff}. 
These methods motivate a three-way distinction: soft guidance improves preference or reward; hard constraints aim for guaranteed validity; posterior-control methods condition generation on observations or inverse-problem structure. 
All three exploit the same inference-time freedom, but they are evaluated with different metrics.

\subsection{Acceleration: Distillation, Caching, and Parallelism} \label{subsec:inf_accel}

\paragraph{Distillation for fewer denoising steps.} 
Practical acceleration begins by distinguishing algorithmic step reduction from actual serving speed. Distillation methods reduce the number of refinement steps by training a new student, sampler, or consistency structure. 
Di4C distills dimensional correlations so low-step discrete samplers better preserve joint structure~\citep{unknown2025distdiscdiff}. 
Learnable Sampler Distillation keeps the denoiser fixed but learns sampler coefficients and timesteps~\citep{unknown2025learsampdist}. 
One-step and consistency models go further: Di[M]O distills masked image generation into a one-step generator, DLM-One is an early one-step language diffusion system, d3LLM uses pseudo-trajectory distillation, and CDLM/CD4LM train consistency or adaptive-decoding students for faster language generation~\citep{unknown2025dimodistmask,unknown2025dlmodifflang,unknown2025d3llultrdiff,unknown2025cdlmconsdiff,unknown2025cd4lconsdist}. 
These methods can yield large NFE reductions, but they are training-time interventions rather than purely training-free decoding policies.

\paragraph{Parallel and selective inference.} 
Per-step computation can also be reduced without reducing the number of denoising iterations. 
ReCAP reuses context features in masked generation~\citep{unknown2025plugcontfeat}; DiCache learns sample-adaptive cache schedules for visual diffusion~\citep{unknown2026dicaletdiff}; SparseD exploits stable sparse attention patterns across denoising steps~\citep{unknown2026sparsparatte}.
ES-dLLM skips low-importance tokens after early layers using confidence and hidden-state variation, while SureLock stops computation for locally converged tokens but preserves their K/V states for the remaining active tokens~\citep{unknown2026esdleffiinfe,unknown2026stopcompconv}.
FOCUS evicts non-decodable tokens after early-layer importance scoring, and D3ToM/DToM applies selective compute to diffusion MLLMs by dynamically merging visual tokens according to decider-token salience~\citep{unknown2025focudllmknow,unknown2025dtomdecidyna}. 
These methods share a systems observation: a denoising step need not be implemented as a full-sequence Transformer call.

Suffix and window methods expose another source of waste. 
DPad treats far-future suffix masks as a scratchpad and drops distant suffix tokens with sliding windows or distance-decay dropout~\citep{unknown2026dpadeffidiff}. 
Streaming-dLLM formalizes suffix pruning and dynamic thresholding for long generation, showing substantial throughput gains when much of the canvas is not locally useful~\citep{unknown2025streaccediff}. 
Fast and Fluent diffusion decoding shows that long windows can also hurt fluency, motivating convolutional decoding and rejective fine-tuning~\citep{unknown2025fastfluediff}. 
Loopholing Discrete Diffusion addresses a different bottleneck: categorical sampling collapses soft information at each step, so deterministic latent bypasses can reduce idle or oscillating steps~\citep{unknown2026loopdiscdiff}. 
These works complement cache methods by asking not only what to reuse, but which regions belong in the computation at all.

Speculative decoding is another acceleration path. 
Speculative Diffusion Decoding uses diffusion to draft multiple tokens for an AR verifier~\citep{arriola2025speculative}. 
DiffuSpec improves the diffusion-as-drafter approach with causal-consistency path search and adaptive draft length~\citep{unknown2025diffunlodiff}. 
FailFast argues that dLLMs are useful as adaptive drafters: difficult regions can be rejected quickly, while easy regions can produce long accepted drafts~\citep{unknown2025failfastwin}. 
DART is diffusion-inspired rather than a full diffusion LM; it trains a lightweight parallel drafter conditioned on AR target hidden states and verifies with the target model~\citep{unknown2025dartdiffspec}. 
A separate line makes draft-and-verify native to dLLMs: FreeDave verifies future diffusion states, Spiffy builds directed draft graphs over future block states, and Self-Speculative Masked Diffusions split drafting and verification inside the architecture~\citep{unknown2025freedraftowa,unknown2025spifmultdiff,unknown2026selfmaskdiff}. 
It is useful to distinguish AR-target speculation from dLLM-target speculation, since their correctness and latency accounting differ.

\paragraph{Caching in bidirectional denoisers.} 
A prominent systems line for dLLMs is cache and feature reuse. 
The difficulty is that bidirectional diffusion invalidates the simple append-only KV cache used by AR LMs. 
dLLM-Cache observes different temporal stability for prompt and response tokens and reuses cached states on different schedules~\citep{unknown2025dllmaccediff}. 
dKV-Cache caches already decoded tokens with delayed updates, showing that partial KV reuse is possible even in bidirectional denoising~\citep{unknown2025dkvccachdiff}. 
Fast-dLLM combines approximate block-wise KV cache with confidence-aware parallel decoding~\citep{unknown2026fasttraiacce}; FlashDLM adds efficient KV caching and a guided diffusion mechanism~\citep{unknown2026flasaccediff}. 
SlowFast sampling and learnable/adaptive parallel decoding show that cache reuse is often paired with token-acceptance policies rather than deployed alone~\citep{unknown2026accedifflarg,unknown2025learparaacce,unknown2026learparaacce}.
dCache/d$^2$Cache, Dynamic-dLLM, and Attention-Is-All-You-Need-style cache methods refine the policy with token-, attention-, or budget-aware refresh~\citep{unknown2026dcacaccediff,unknown2026dynadynacach,unknown2026atteallyou}. 
SPA-Cache pushes this trend further by using singular-value proxies of value-state drift and layer-wise adaptive budgets~\citep{unknown2025spacsingprox}. 
The progression is from coarse cache reuse toward state-aware cache control. 
Open problems around KV-cache analogues and streaming generation are discussed further in Section~\ref{sec:kv_cache_future}.

\paragraph{Systems metrics and wall-clock evaluation.} 
A final consideration is evaluation discipline. 
Acceleration comparisons are more informative when they report the backbone, generation length, batch size, prompt length, hardware, cache refresh policy, memory overhead, number of real model calls, wall-clock latency or throughput, and quality under matched prompts. 
Hardware-aware work makes this especially visible: arithmetic-intensity-inspired scheduling and NPU-oriented designs show that the practical bottleneck may be memory movement, irregular sparsity, cache overhead, or verification cost rather than nominal NFE~\citep{unknown2025orchdualarit,unknown2025beyogemmnpus}. 
This connects Section 6 to the broader efficient-inference literature for reasoning models, but dLLMs add their own complications because every denoising pass touches a mutable canvas rather than appending one token~\citep{unknown2025effiinfelarg}. 
Without such details, a reported "10x speedup" may mean fewer denoising rounds, more tokens accepted per round, fewer active tokens per layer, more favorable batching, or hardware-specific throughput. 
Speed can therefore be treated as a systems property produced by the interaction of sampler policy, model architecture, and execution substrate.

%% file: tables/06-inference-methods.tex
\begin{table}[t]
\centering
\caption{Degrees of freedom exposed by the discrete diffusion reverse process, with representative mechanisms and the failure mode each must guard against.}
\label{tab:inference_methods}
\small
\setlength{\tabcolsep}{4pt}
\renewcommand{\arraystretch}{1.15}
\begin{tabular}{@{}p{3.5cm}p{3.5cm}p{4.5cm}p{4cm}@{}}
\toprule
\textbf{Control axis} & \textbf{What it decides} & \textbf{Representative mechanisms} & \textbf{Main failure mode} \\
\midrule
Step schedule & When to visit reverse states & Adaptive / dilated schedules, high-order solvers, predictor--corrector & Quality loss at low NFE \\
\midrule
Token selection & Which positions to commit or reopen & Confidence / entropy remasking, planned unmasking & Miscalibrated confidence \\
\midrule
Block structure & Dependency topology of updates & Block / semi-autoregressive decoding & Parallel dependency error \\
\midrule
Guidance \& constraints & External objectives and validity & CFG, reward / search, projection, grammars & Altered reverse dynamics \\
\midrule
Acceleration & Reused or skipped computation & Distillation, KV / feature caching, speculation & NFE $\ne$ wall-clock \\
\bottomrule
\end{tabular}
\end{table}

%% file: sections/08-scaling.tex
\section{Scaling and System Considerations} \label{sec:scaling}

%
This section turns from formulations and algorithms to systems-level questions that govern whether discrete diffusion is practical at scale. 
Section~\ref{subsec:scaling_backbone} reviews backbone architectures and how bidirectionality reshapes positional encoding and length generalization. 
Section~\ref{subsec:scaling_compute_optimality} examines scaling laws and compute-optimality. 
Section~\ref{subsec:training_pipeline} covers the training pipeline of pretraining, supervised fine-tuning, and alignment, and Section~\ref{subsec:scaling_deploy} discusses deployment and product constraints such as latency, throughput, and hybrid serving.

\noindent





\subsection{Backbone Architectures} \label{subsec:scaling_backbone}
The fundamental shift from autoregressive (AR) generation to iterative denoising necessitates reevaluating the underlying neural backbones. 
While Transformers currently dominate the landscape, the discrete diffusion framework itself is agnostic to the specific neural architecture, opening avenues for diverse sequence modeling paradigms.

\paragraph{Bidirectional transformers.} 
Unlike AR models, which rely on left-to-right next-token prediction, discrete diffusion models learn a reverse denoising process that reconstructs clean text from corrupted or masked sequences. 
This paradigm naturally requires progressive full-context prediction, making bidirectional Transformers a highly intuitive backbone~\citep{dream2025dream7bdiffu, nie2025largelanguag}. 
By leveraging parallel, global attention, the model can simultaneously incorporate contextual information from all positions at once to denoise the text. 
However, utilizing full-context attention can sometimes introduce a "training-inference gap" if the model is later decoded in a semi-autoregressive, block-by-block manner. 
To address this, recent systems have proposed block-wise attention patterns that apply bidirectional modeling within individual blocks while enforcing causal masking across blocks~\citep{unknown2025blocsftdiff, unknown2025effiautodiff}.

Crucially, the fully bidirectional nature of these Transformers profoundly impacts positional encodings and length generalization. 
Defining the relative position offset as $\Delta = i-j$, where $i$ is the current position and $j$ is the attended position, standard AR models only expose the model to non-negative offsets during training, i.e., $\Delta \in [0,\, T_{\text{train}}-1]$. 
By contrast, global bidirectional attention exposes diffusion models to both negative and positive offsets, yielding a substantially more symmetric range, $\Delta \in [-(T_{\text{train}}-1),\, T_{\text{train}}-1]$~\citep{unknown2025longunlolong, unknown2026ultrscalcont}.
By effectively learning both negative and positive relative distances, diffusion models capture a complete period of the sine and cosine components in Rotary Position Embeddings (RoPE). 
This symmetric coverage drastically reduces out-of-distribution embedding spaces, allowing diffusion language models to maintain remarkably stable perplexity during direct length extrapolation—avoiding the catastrophic performance collapse typical of AR models~\citep{unknown2025longunlolong}. 
Recognizing this, researchers have successfully adapted Neural Tangent Kernel (NTK)-based RoPE scaling into "Diffusion-aware NTK," extending context windows up to 128K tokens via lightweight post-training~\citep{unknown2026ultrscalcont}.

\paragraph{State space models (SSMs).}
Although bidirectional Transformers provide a strong foundation, the diffusion training objective does not impose strict constraints on the choice of the denoising network~\citep{inception2025mercuryultra}. 
State Space Models (SSMs) and recurrent architectures, such as Mamba, present a compelling alternative backbone for diffusion language models~\citep{inception2025mercuryultra, unknown2025seriscalhypo}. 
SSMs are particularly notable for their ability to perform linear-time sequence modeling using selective state spaces~\citep{inception2025mercuryultra}. 
While full-attention Transformers scale quadratically with sequence length—often bottlenecking the generation of massive contexts—Mamba-like backbones could alleviate this computational burden.
Consequently, integrating SSMs into the diffusion framework holds significant promise for supporting long sequences and enabling efficient linear-time inference, potentially marrying the arbitrary-order generation capabilities of diffusion with the scalable context mechanics of modern recurrent models~\citep{unknown2025seriscalhypo}.

\subsection{Scaling Laws and Compute-Optimality}\label{subsec:scaling_compute_optimality}

\paragraph{Scaling questions.}
A central question for discrete diffusion language models (DLMs) is whether they follow scaling trends comparable to those that have underpinned the success of autoregressive (AR) models.
Current empirical evidence suggests that DLMs can scale competitively both when trained from scratch and when adapted from pre-trained AR models. 
For instance, LLaDA reports that, at the 8B scale, masked diffusion models can achieve performance competitive with strong AR counterparts under matched compute budgets~\citep{nie2025largelanguag}; this competitiveness is as self-reported by the originating work and has not been independently audited here.
Similarly, DiffuLLaMA shows that increasing the size of adapted DLMs, up to the 7B scale, is associated with clear gains on downstream tasks~\citep{ye2025scalingdiffu}.

However, recent analyses suggest that compute-optimal token-to-parameter ratios and training dynamics of DLMs may differ from those of AR models. 
According to \citet{unknown2026scalbehadisc}, different diffusion noise types behave similarly in strictly compute-bound settings, whereas in data-bound regimes uniform diffusion appears to favour a more parameter-heavy and data-light scaling profile. 
If this pattern hold at larger scale, it could make DLMs particularly relevant in settings where high-quality training data become a limiting factor. 
Furthermore, Efficient-DLM suggests that training choices affect the downstream efficiency frontier: stronger AR-to-DLM conversion, together with blockwise attention and position-dependent masking, can improve the trade-off between generation quality, throughput, and the number of function evaluations (NFE) at inference time~\citep{unknown2025effiautodiff}. 
Finally, recent work on overparameterised diffusion models suggests that generalisation and memorisation compete over training time: across image and language diffusion settings, the onset of memorisation is empirically observed to scale approximately with dataset size, providing a motivation for early stopping when generalisation is the priority~\citep{unknown2025biggisntalwa}.

\paragraph{Empirical evidence.}
Beyond training-time scaling, DLMs also exhibit distinctive inference-time trade-offs, among generation quality, decoding steps, and test-time compute. 
Unlike AR decoding, diffusion generation does not require a strict one-step-per-token correspondence between output length and inference depth. 
Reported trends indicate that more denoising steps can improve generation quality on complex logical and planning tasks, as illustrated by Dream 7B on Countdown and Sudoku~\citep{dream2025dream7bdiffu}.

Another distinctive inference-time issue in DLMs concerns the initial generation length budget.
Unlike AR models, where output length is typically determined dynamically through \texttt{<eos>}, masked DLMs are often initialised with a fixed budget of mask tokens. 
Rather than treating longer budgets as universally beneficial, recent work frames the denoising trajectory itself as a source of reasoning-time computation, where intermediate reverse-diffusion states can act as latent reasoning actions optimised by outcome-based reinforcement learning~\citep{unknown2025reindiffchai}. 
However, this form of test-time scaling also has practical limits. 
Naively allocating an excessively long generation budget can trigger an empirical failure mode known as \texttt{<eos>} overflow, in which the model terminates prematurely or degenerates into repeated \texttt{<eos>}-like padding outputs~\citep{unknown2026rainpaddmiti}.

A related issue concerns the limits of step reduction. 
Naively compressing inference steps with uniform schedules can degrade generation quality, since early denoising stages are typically lower-confidence and more uncertainty-sensitive. 
To mitigate this problem, recent work has proposed more adaptive scheduling strategies, such as the Ascending Step-Size (ASS) scheduler. 
Under its proposed decoding design, ASS is reported to reduce the required number of inference steps from a linear bound $O(L)$ to a logarithmic scale $O(\log L)$ while maintaining competitive generation quality~\citep{unknown2025tamimaskdiff}; this $O(\log L)$ claim is the originating paper's analysis under its own assumptions and decoding design, not a worst-case guarantee verified here.

Finally, early evidence suggests that DLM scaling behaviour is sensitive to the chosen corruption strategy and may also vary across task regimes. 
Current evidence does not yet support a universal best corruption strategy; instead, different diffusion variants may behave differently across reasoning-heavy and knowledge-heavy task regimes~\citep{unknown2026scalbehadisc}. 
Taken together, these observations suggest that scaling choices for DLMs should likely be matched to the downstream task regime rather than treated as universally optimal.

\subsection{Training Pipeline: Pretraining, SFT, and Alignment}\label{subsec:training_pipeline}

\paragraph{Pretraining.}
Pretraining masked discrete diffusion language models can be viewed as a non-autoregressive analogue of next-token prediction. 
Rather than generating tokens sequentially, these models corrupt an input sequence and learn to recover the clean text through a parallel denoising objective related to the ELBO~\citep{nie2025largelanguag}. 
Because training such models from scratch is expensive, many recent studies instead adapt pre-trained autoregressive (AR) models into diffusion language models (DLMs). 
For example, DiffuLLaMA~\citep{ye2025scalingdiffu} and Dream \citet{dream2025dream7bdiffu} introduce a shift operation to preserve aspects of the AR model's next-token dependency structure during diffusion training, while LAD~\citep{unknown2025ladloradiff} adopts a parameter-efficient LoRA-based adaptation strategy together with structured perturbations such as token swaps, duplication, and span shifts.

Pretraining also involves diffusion-specific choices in corruption schedules and representation design. 
Dream \citet{dream2025dream7bdiffu} proposes Context-Adaptive Token-Level Noise Rescheduling (CART), which assigns noise levels according to contextual informativeness. 
Efficient-DLM~\citep{unknown2025effiautodiff} further introduces position-dependent masking to better reflect the left-to-right structure of language and reduce training--inference mismatch. 
To enrich intermediate representations, Soft-Masked Diffusion~\citep{unknown2026softdifflang} replaces hard masking with a soft mixture over the top-$k$ predicted embeddings from earlier denoising steps.

\paragraph{Instruction tuning.}
Supervised fine-tuning (SFT) adapts diffusion denoisers to follow natural-language instructions and prompts. 
In a standard dLLM SFT setup, the prompt is typically kept clean while the response tokens are randomly masked. 
However, this still mismatches the semi-autoregressive blockwise decoding strategy often used at inference time, where the model conditions on a clean prefix and a fully hidden future~\citep{unknown2025blocsftdiff}. 
Blockwise SFT~\citep{unknown2025blocsftdiff} addresses this mismatch by hiding all future tokens and computing the loss only over the currently decoded block. 
Similarly, Planner Aware Path Learning (PAPL)~\citep{unknown2026planawarpath} incorporates planned denoising trajectories into the training objective.

To better support later reasoning-oriented post-training, IGPO \citet{unknown2026inpapoliopti} introduces Length-Aligned SFT, which synthetically rewrites verbose reasoning traces into more concise forms that better match the generation profile of full-attention dLLMs during RL sampling. 
In addition, Rainbow Padding~\citep{unknown2026rainpaddmiti} mitigates early termination under large decoding budgets by decoupling padding from termination and replacing repeated \texttt{<eos>} placeholders with a cyclic set of distinct padding tokens. 
Beyond pure text generation, Co-GRPO~\citep{unknown2025cogrcoopgrou} reformulates masked diffusion generation as a Markov Decision Process and jointly optimises model parameters and inference schedules, with reported gains in generation quality and alignment in broader masked-diffusion settings.

\paragraph{Preference optimization.}
Preference optimization and reinforcement learning have become an increasingly active area of DLM post-training. 
Relative to autoregressive models, the main complication is that rewards and preferences must be assigned over an iterative denoising trajectory rather than a single left-to-right generation path. 
This makes both credit assignment and likelihood estimation more difficult, and has led to several distinct but related lines of work.

One line of research extends DPO-style offline preference alignment to diffusion processes. 
Here, the core challenge is how to construct tractable and stable preference objectives over denoising trajectories or their variational surrogates. 
LLaDA 1.5~\citep{nie2025llada15varia} addresses this issue with Variance-Reduced Preference Optimization (VRPO). 
SDPO~\citep{unknown2026discdifftraj} further decomposes sequence-level trajectory alignment into stepwise posterior-matching objectives, making optimization more tractable. 
D2-DPO~\citep{unknown2025prefaligdisc} also adapts DPO to continuous-time diffusion dynamics, although its empirical validation is mainly conducted on structured binary sequences rather than full language generation. 
It is also important to distinguish these methods from superficially similar work such as DiffPO~\citep{unknown2025diffdiffpref}, which applies diffusion-inspired iterative refinement to inference-time alignment for AR LLMs rather than performing preference optimization for native dLLMs.

A second line of work studies online RL and policy-gradient training for dLLMs. 
The main bottleneck here is that exact per-step likelihoods are typically unavailable or expensive to compute, so most methods rely on approximations or alternative objectives. 
d1~\citep{zhao2025d1scaling} estimates likelihood ratios through a single-step mean-field approximation for GRPO-style updates, while d2 \citet{unknown2025d2imprtech} develops trajectory-likelihood estimators such as d2-AnyOrder and d2-StepMerge to improve policy-gradient training for masked DLMs. 
Other methods instead seek to reduce or avoid the instability introduced by ratio estimation. 
wd1 \citet{unknown2026wd1weigpoli} proposes a ratio-free weighted policy optimization objective, while SPG \citet{unknown2026spgsandpoli} introduces sandwiched variational bounds to reduce bias in policy-gradient estimation. 
At a broader theoretical level, SEPO \citet{unknown2025finediscdiff} provides a general policy-gradient framework for discrete diffusion models based on an explicit score-entropy formulation.

Beyond stepwise optimization, recent work increasingly treats the denoising trajectory itself as a central object of alignment. 
This shift is motivated in part by the possibility that useful intermediate predictions may later be degraded or overwritten by subsequent denoising steps~\citep{unknown2025mdpoovertrai}. 
MDPO~\citep{unknown2025mdpoovertrai} and TraceRL~\citep{unknown2026revoreinlear} therefore optimise trajectory-level behaviour rather than treating each denoising step in isolation. 
d-TreeRPO~\citep{unknown2025dtretowamore} further improves trajectory evaluation through tree-structured rollouts, enabling verifiable step-wise reward propagation and more precise transition-probability estimation. 
At inference time, EntRGi \citet{unknown2025entrentrawar} shows that generation can also be steered without additional training through entropy-aware reward guidance. 
At the same time, the trajectory-centric and any-order nature of DLM decoding introduces new safety concerns. 
A2D \citet{unknown2026a2danyoanys} shows that arbitrary-order decoding can enlarge the attack surface for jailbreaks, and proposes token-level safety alignment as a corresponding defence.
Taken together, these studies suggest that post-training for DLMs is gradually shifting from token-level preference matching toward broader control over the denoising trajectory itself.

\subsection{Deployment and Product Constraints} \label{subsec:scaling_deploy}

\paragraph{Latency and throughput targets.} 
Discrete diffusion models (DDMs) have a different deployment trade-off from autoregressive (AR) models. 
AR models generate text one token at a time, while DDMs update many tokens in parallel through iterative denoising. 
Because of this, DDMs can make better use of highly parallel hardware such as GPUs and TPUs in some serving settings. 
For example, \textsc{Mercury} \citet{inception2025mercuryultra} reports that diffusion-based coding models can reach very high throughput, with self-reported figures exceeding 1,000 tokens per second on its own serving stack and hardware, claimed to outperform optimized AR baselines.
As with the other headline numbers in this section, these are figures reported by the originating work rather than independently reproduced, and the hardware, batch size, and baselines differ across papers; they should be read as indicative rather than directly comparable. 
\textsc{Efficient-DLM}~\citep{unknown2025effiautodiff} further shows that converting pre-trained AR models into diffusion models with block-wise attention can improve throughput while keeping competitive accuracy. 
In addition, systems such as \textsc{DiRL}~\citep{unknown2025dirleffipost} combine practical inference engines and efficient attention implementations to reduce the cost of iterative diffusion decoding. 
However, DDMs still face an important product constraint: Time-To-First-Token (TTFT). 
Since bidirectional diffusion needs several denoising steps before producing readable text, it may be less suitable than AR models for interactive streaming applications where fast first response matters.

\paragraph{Hybrid models.} 
To balance the cost of full diffusion decoding with the benefit of parallel generation, recent work has explored hybrid and semi-autoregressive designs. 
In this line of work, sequences can be processed block by block, while tokens inside each block are refined in parallel. 
Blockwise SFT~\citep{unknown2025blocsftdiff} improves this setting by reducing the mismatch between standard supervised fine-tuning and blockwise diffusion inference, which leads to better performance on complex reasoning tasks. 
Beyond hybrid decoding, DDMs are also naturally useful for refinement. 
Their bidirectional structure makes them well suited for editing, inpainting, and self-correction.
For example, \textsc{PRISM}~\citep{unknown2025priseffitest} adds an inference-time self-correction module that identifies low-quality tokens and re-masks them for another round of revision. 
In reinforcement learning, \textsc{IGPO} \citet{unknown2026inpapoliopti} uses this inpainting ability to insert partial ground-truth reasoning traces during exploration, which can improve sample efficiency in sparse-reward settings. 
These results suggest that DDMs can be used not only as standalone generators, but also as refinement and editing modules on top of existing language modeling pipelines.

%% file: sections/09-applications.tex
\section{Applications} \label{sec:applications}

%
This section instantiates the framework across application domains, showing how the same corruption-denoising machinery specializes to very different discrete state spaces. 
The first four subsections cover modalities with engineered tokenizations: language and text (Section~\ref{subsec:app_text}), code (Section~\ref{subsec:app_code}), multimodal foundation models (Section~\ref{subsec:app_multimodal}), and tokenized media (Section~\ref{subsec:app_media}). 
The next four cover domains with natural discrete structure: proteins (Section~\ref{subsec:app_protein}), genomics (Section~\ref{subsec:app_genomics}), molecules and graphs (Section~\ref{subsec:app_molecule}), and planning and agents (Section~\ref{sec:agents_planning}). 
We then treat tabular data and imputation (Section~\ref{subsec:app_tabular}) and close with recommended case studies and a reporting checklist for new domains (Section~\ref{subsec:app_casestudies}).
The domains differ in maturity: text/code and tokenized media are well established with large-scale systems and standard benchmarks; proteins, genomics, and molecules/graphs are active but more specialized; and planning/agents and tabular generation are comparatively emergent, with results that are promising but preliminary. 
We try to flag this evidence strength as we go rather than presenting all domains at a uniform cadence.
Table~\ref{tab:applications_overview} summarizes these domains and explicitly separates well-established settings from active but specialized or still-emerging ones.

\input{tables/07-applications}

\subsection{Language Modeling and Text Generation} \label{subsec:app_text}
Text generation is among the earliest and most active applications of discrete diffusion.
Existing work spans unconditional generation, sequence-to-sequence generation, controllable generation, rewriting and paraphrasing, summarization, data augmentation, instruction following, and emerging reasoning-oriented applications. 
Early continuous or simplex-based approaches such as Diffusion-LM and SSD-LM diffuse in continuous embedding or simplex spaces and demonstrate the potential of diffusion for controllable text generation and modular control~\citep{li2022diffusionlm,han2023ssdlmsemiaut}. 
Sequence-to-sequence diffusion models extend this idea to conditional generation, where the source sequence is kept as conditioning context and the target sequence is generated through denoising~\citep{gong2023diffuseq,unknown2023diffbriddisc,unknown2024textdiffmode}. 
Other works use diffusion for non-autoregressive generation and machine translation, often combining iterative refinement with masked prediction to improve parallel decoding quality~\citep{unknown2024diffselfdisc,unknown2024diffglantran,unknown2025diffdireacyc}.

A second major direction is controlled generation. 
Because diffusion generation proceeds through iterative refinement rather than one-pass left-to-right decoding, control signals can be injected at multiple stages of generation. 
This has been used for attribute control, conversation-structure control, commonsense knowledge generation, controllable synthetic data generation, and text editing~\citep{unknown2023contconvgene,unknown2024diffcontcomm,unknown2024diffguidlang,unknown2025diffcontsynt,lee2025editcontcoar}. 
Diffusion language models have also been applied to summarization and document-oriented tasks, including perspective-preserving summarization, efficient text summarization, and generative document retrieval~\citep{unknown2024p3supresauth,unknown2025discdifflang,unknown2025diffgenedocu}. 
More recently, large-scale diffusion language models have been explored for instruction following and general-purpose language modeling, suggesting that diffusion is moving from task-specific text generation toward broader LLM-style applications~\citep{mahabadi2025tess2largesc,sahoo2025scalingup,nie2025largelanguag,dream2025dream7bdiffu,unknown2025difflangmode}.

Why diffusion is attractive here (global revision, bidirectional conditioning, and repeatedly applicable guidance) and when it is not, follows the general analysis of Section~\ref{subsec:disc_refinement}; we do not repeat it. 
The way these systems instantiate the four-component pipeline (corruption operator, denoiser parameterization, training objective, sampler) likewise follows the unified view of Section~\ref{sec:formulations} and Table~\ref{tab:formulation_map}, with text-specific choices being absorbing-state masking over a subword or byte vocabulary, a bidirectional Transformer denoiser, a reweighted denoising objective, and a confidence-based or block sampler.

Language modeling and text generation provide a central testbed for discrete diffusion. 
The area has progressed from early embedding-space and simplex diffusion models to masked diffusion LMs and large-scale diffusion LLMs. 
Across tasks, the main advantages are global revision, bidirectional conditioning, controllable refinement, and robustness to local generation errors. 
The main challenges are multi-step latency, weaker streaming behavior, difficulty matching AR models in in-context learning and reasoning, and the need for standardized evaluation. 
These trade-offs make text generation a useful lens for understanding the broader design space of discrete diffusion models.

\subsection{Code: Infilling, Repair, and Structured Constraints} \label{subsec:app_code}

Code generation has become an active testbed for discrete diffusion \citep{gong2026diffucoder, unknown2025codacodilm, unknown2025diffacceunit, unknown2025discmathequa, unknown2025drea7bopen, unknown2026difflangmode}, and in four-component terms these systems share a common instantiation: an absorbing (masking) corruption over a code-subword vocabulary, a bidirectional Transformer denoiser, a reweighted denoising objective, and a sampler augmented with syntactic or test-based constraints, differing mainly in whether the denoiser is trained from scratch or adapted from an AR code model, and in how constraints enter the sampler.
Software development frequently involves editing existing codebases through tasks such as fill-in-the-middle completion and bug fixing. 
Autoregressive models face challenges in these scenarios due to their left-to-right decoding constraints. 
Discrete diffusion models address this by treating the target edit region as an absorbing masked state, allowing the reverse denoising process to simultaneously condition on both preceding and succeeding code segments. 
Recent masked diffusion models indicate that this bidirectional awareness facilitates non-causal reasoning and flexible generation orders, such as drafting the structural outline before filling in specific details~\citep{gong2026diffucoder, unknown2025drea7bopen}. 
This bidirectional capability can also be implemented by adapting existing autoregressive models into diffusion models via specialized masking strategies~\citep{unknown2025codacodilm}.

Beyond context modeling, the iterative denoising trajectory of diffusion supports the injection of structural constraints during generation to enforce the validity of the outputs. 
These constraints are applicable to code synthesis as well as tasks requiring strict syntactic correctness, such as discovering mathematical equations~\citep{unknown2025discmathequa}. 
The iterative framework also accommodates dynamic validation techniques. 
For instance, unit test generation can be accelerated by mining structural patterns from abstract syntax trees and incorporating them into intermediate decoding steps to guide the model toward functionally correct solutions~\citep{unknown2025diffacceunit}. 
Finally, applying these infilling techniques to repository-level contexts requires handling varying edit lengths. 
Recent work addresses this through dynamic length adjustment mechanisms, allowing the diffusion process to expand or contract the target sequence while processing localized blocks within long files~\citep{unknown2026difflangmode}.

\subsection{Multimodal Foundation Models (Text + Vision/Audio)} \label{subsec:app_multimodal}
\noindent

%
A growing line of research generalizes discrete diffusion from text-only dLLMs to \emph{diffusion multimodal large language models} (dMLLMs), which jointly handle text and additional modalities, including images, video, and audio. 
These methods share a core objective: substituting the conventional autoregressive (AR) backbone (used in standard MLLMs) with a bidirectional, masked-denoising backbone, thereby enabling understanding and generation to be performed by one unified decoder. 
We categorize this emerging area into four primary dimensions: the objectives of dMLLMs, the way different modalities are embedded and organized in the token space, how corruption schedules are synchronized across multiple streams, and the distinctive failure modes that stem from employing a single shared denoiser.

\paragraph{Objectives of diffusion MLLMs.}
Early multimodal masked generative transformers, such as MAGVLT~\citep{unknown2023magvmaskgene}, together with unified discrete diffusion methods~\citep{hu2023unifieddiscr}, introduced a key shift: they demonstrated that vision-language tasks can be solved by predicting masked tokens across multiple modalities, instead of relying on strictly left-to-right token decoding. 
Show-o~\citep{xie2024showoone} operationalized this idea by employing a single transformer backbone for both understanding and generation, a strategy that was subsequently adopted in instruction-tuned frameworks. 
MMaDA~\citep{kou2025mmadamultimo} and LLaDA-V~\citep{nie2025lladavlarge} were subsequently proposed as multimodal analogues of text-only masked dLLMs. 
Collectively, these models demonstrate that pairing bidirectional attention with iterative remasking allows the system to repeatedly revisit and incrementally refine its predictions. 
More recent approaches, such as Lavida-O~\citep{unknown2026lavielaslarg}, Lumina-DiMOO~\citep{unknown2025lumiomnidiff}, and FUDOKI~\citep{unknown2025fudodiscflow}, instead emphasize "omni" backbones, which perform both generation and understanding within a single denoising step.
Meanwhile, Muddit~\citep{unknown2025muddlibegene} and \citet{unknown2026beyotextlibe} extend generative modeling beyond basic text-to-image applications. 
Overall, the twofold advantage of dMLLMs compared with AR MLLMs becomes clear: they support the simultaneous decoding of multiple modalities, and they enable globally consistent revision of multimodal outputs, rather than committing to an irreversible left-to-right generation process.

\paragraph{Unified tokens vs.\ cross-attention.}
The central idea concerns how to fuse multiple modalities. 
One strategy converts each data stream into a shared set of discrete tokens that a single denoiser processes; the other keeps separate encoders per modality and connects them using cross-attention.
Models in the first category—including Show-o~\citep{xie2024showoone}, Unified Multimodal Discrete Diffusion~\citep{unknown2025unifmultdisc}, MMaDA~\citep{kou2025mmadamultimo}, Muddit~\citep{unknown2025muddlibegene}, and Lumina-DiMOO~\citep{unknown2025lumiomnidiff}—rely on a VQ tokenizer or neural codec to convert non-text inputs into discrete tokens, which are then concatenated with textual tokens. 
A bidirectional self-attention mechanism is subsequently applied to capture dependencies both within each modality and across different modalities. 
This approach keeps the denoiser straightforward, but it introduces design decisions around codebook resolution, vocabulary fusion, and positional encoding schemes for mixed-modal streams. 
In contrast, cross-attention-based dMLLMs like LLaDA-V~\citep{nie2025lladavlarge} and LaDiC~\citep{unknown2024ladidiffmode} maintain image representations in a continuous encoder and apply diffusion only to the text tokens; this sacrifices a single, unified generative process in exchange for tighter alignment with existing visual-instruction-tuning workflows. 
Hybrid approaches fall between these extremes: SDAR-VL~\citep{unknown2025sdarstabeffi} applies block-wise denoising to improve stability, Sparse-LaViDa~\citep{unknown2025sparsparmult} prunes the multimodal context to cut computation, and Lavida-O~\citep{unknown2026lavielaslarg} adds flexible token budgets so a single backbone can process everything from brief captions to long videos.

\paragraph{Corruption schedules across streams.}
When several token streams rely on a single denoiser, their corruption schedules must be carefully aligned. 
Because text and image tokens inherently carry different amounts of information—with a masked image patch typically holding less semantic meaning than a masked word—applying the same masking ratio across modalities is suboptimal. 
It can squander model capacity on easy image inpainting while degrading the text stream so severely that learning becomes ineffective. 
To mitigate this, recent approaches introduce modality-specific schedules. 
Prominent instances include distinct text and image mask ratios in MMaDA~\citep{kou2025mmadamultimo} and its parallel thinking–aware extensions~\citep{unknown2026paramultdiff, unknown2025mmadmultlarg}; block-wise frame masking for video dMLLMs such as VidLaDA~\citep{unknown2025vidlbididiff} and VDT-style transformers~\citep{lu2024vdtgeneralpu}; and flow-based velocity scheduling in FUDOKI~\citep{unknown2025fudodiscflow}. 
How these schedules are coordinated forms an important dimension of model design. 
Some architectures draw a single diffusion timestep and apply it across all modalities via separate noise functions, while others assign distinct timesteps to each stream, training the model to reconstruct one modality from the partially corrupted information in another. 
In addition, as illustrated by masked diffusion captioning~\citep{unknown2025maskdiffcapt}, the masking distribution itself plays a decisive role in shaping which visual features the model ultimately learns to emphasize, even when the objective is purely generative.

\paragraph{Failure modes and open issues.}
Despite rapid advances, dMLLMs display distinctive failure patterns that diverge from those of AR MLLMs. 
First, because later iterations are conditioned on partially denoised latents, early mistakes can cascade through subsequent steps; corrective methods~\citep{unknown2025denoreficorr} address this by reintroducing masks mid-process so the model can revise earlier errors. 
Second, context tokens may act as anchors, leading to repetitive or degenerate generations in dMLLMs~\citep{unknown2025conttokeanch, unknown2026conttokeanch}, a phenomenon that becomes more pronounced as sequence length increases. 
Third, the quality of decoded outputs is fundamentally capped by the tokenizer: limited codebooks can cause visual or auditory artifacts that cannot be fixed by any amount of denoising in code space. 
Finally, because diffusion-based sampling distributes computation across many iterations, how test-time computation is allocated becomes critical; for instance, self-verified test-time scaling for dMLLM-based TTS \citet{unknown2025dmllselfeffi} shows that inference-time remasking and verification can be as influential as the underlying architecture. 
Overall, these observations indicate that the advantages of parallel, globally consistent multimodal generation introduce new tuning burdens related to scheduling, repetition control, and verification, and that dMLLMs should be regarded as complementary to, rather than direct drop-in replacements for, AR multimodal foundations.%

\subsection{Tokenized Media Generation (Images, Audio, Video)} \label{subsec:app_media}

%
Beyond text, a substantial body of work applies discrete diffusion to continuous-valued media such as images, audio, and video. 
The basic recipe first compresses the raw signal into a sequence of discrete codes using a vector-quantized autoencoder or neural codec, then trains a categorical denoiser over those codes, and finally decodes the predicted codes back into the original signal through the tokenizer's decoder.
We organize this area along four themes: the shared pipeline, how it is instantiated for each modality, the strengths of working in token space, and the limitations inherited from the codec.

\paragraph{The shared pipeline.}
The canonical pipeline for discrete diffusion in media is continuous data $\to$ tokenizer or codec $\to$ discrete diffusion in token space $\to$ decoder. 
This template was established by VQ-Diffusion~\citep{gu2022vectorquanti} and its improved version~\citep{unknown2025imprvectquan}, which replace the Gaussian transition with a categorical one over VQ-VAE codes; the resulting model achieves competitive text-to-image quality and supports natural parallel infilling. 
Muse~\citep{chang2023musetexttoim} and Unleashing Transformers~\citep{unknown2025unletranpara} push the parallel-decoding view further, using more iterations only at sampling time, and Effective and Efficient Masked Image Generation~\citep{unknown2025effeeffimask} reports that careful schedule design can narrow the gap between discrete token diffusion and continuous latent diffusion at fewer sampling steps. 
Subsequent variants add guidance and conditioning machinery on top of this backbone (text-conditioned sampling~\citep{lee2023textsampfram}, global context modeling~\citep{unknown2022globcontdisc}, self-guided sampling~\citep{unknown2024unlocapamask}, and layout- or typography-conditioned generation such as TextDiffuser~\citep{chen2023textdiffuser}) without changing the basic "tokenize then denoise" structure. 
A more recent thread~\citep{unknown2026unifdiscdiff} studies how to unify several such denoisers under a single discrete backbone across modalities, connecting the tokenized view to the dMLLMs discussed above.

\paragraph{Images and videos.}
For images, tokenized discrete diffusion tends to be most effective in settings where the main goal is editing, infilling, or otherwise constrained generation. 
Masked generative transformers~\citep{chang2023musetexttoim, unknown2025effeeffimask} enable rapid editing by allowing arbitrary subsets of tokens to be remasked, and text-conditioned variants~\citep{chen2023textdiffuser, lee2023textsampfram} employ parallel decoding to jointly manage global composition and fine-grained details. 
For video, the basic approach remains similar: spatio-temporal tokens produced by a VQ codec are denoised in parallel, while block- or frame-level corruption is applied to preserve temporal structure. 
VDT~\citep{lu2024vdtgeneralpu} employs masked modeling on video transformers for general-purpose prediction, CineTrans~\citep{unknown2026cineleargene} tailors this pipeline to handle cinematic transitions, and Lumos-1~\citep{unknown2026lumoautovide} integrates AR and discrete-diffusion video generation into a unified framework. 
Discrete diffusion has similarly been leveraged for structured visual outputs that are inherently discrete—including scene text recognition~\citep{unknown2025mdifmaskdiff}, human mesh recovery~\citep{unknown2025megamaskgene}, 3D pose estimation~\citep{unknown2024di2pdiscdiff}, motion generation~\citep{unknown2025divetowadive, unknown2025dimodiscdiff, unknown2025m2d2multgene}, sign language~\citep{unknown2025maditamimask, unknown2025g2pdgenesign}, co-speech gesture~\citep{unknown2025discdiffcosp}, and talking-face synthesis~\citep{unknown2025emotface}, contexts in which a categorical formulation of the outputs naturally matches the label structure.

\paragraph{Audio and speech.}
For audio, tokenization is usually performed by a neural codec, commonly using residual vector quantization (RVQ), which generates multiple parallel streams of discrete tokens for each frame. 
Applying discrete diffusion to these tokens offers the same advantages seen in image-token diffusion—such as parallel decoding, infilling, and editing—while aligning naturally with the architecture of contemporary codec-based speech and music models. 
Representative work includes Diffsound~\citep{unknown2025diffdiscdiff} and MDSGen~\citep{pham2025mdsgfasteffi} for text-to-sound generation; VampNet~\citep{unknown2025vampmusigene} and diffusion-based symbolic music models~\citep{unknown2025diffsymbmusi, unknown2025diffmasklang, unknown2023discdiffprob} for music synthesis; as well as Metis~\citep{unknown2025metifounspee}, an efficient high-fidelity speech system~\citep{unknown2025styleffihigh}, BAD~\citep{unknown2025badbidiauto}, wavelet-domain speech diffusion~\citep{unknown2024speawavedoma}, and VibeVoice~\citep{unknown2026vibeexprpodc} for expressive, long-form speech generation. 
The same framework naturally enables token-level audio inpainting~\citep{unknown2026tokeaudiinpa}, and has further been adapted to discriminative settings such as automatic speech recognition—Drax~\citep{campbell2024draxspeech}, dLLM-ASR~\citep{unknown2025dllmfastdiff}, audio-conditioned diffusion LLMs~\citep{unknown2025audidiffllms}, and cross-modal speech diffusion~\citep{unknown2025crosdiffmode}, in which iterative refinement over hypothesis tokens replaces standard greedy or beam-search decoding. 
In parallel, multiple studies repurpose vector-quantization mechanisms for diffusion operating directly in the acoustic space~\citep{unknown2025vectquandiff, unknown2023diffsemapres, unknown2025discresidiff}.

\paragraph{Strengths and limitations.}
The main advantages of tokenized discrete diffusion for media are its parallel decoding, straightforward editing via remasking chosen tokens, and the ability to reuse backbones from text and multimodal models. 
Together, these features support inpainting, outpainting, and localized editing workflows without any retraining: one simply re-masks and re-decodes a token region under the desired conditioning.
However, two recurring limitations are reported in the literature. 
First, the tokenizer fundamentally limits generation quality: once the codec throws away a detail, it can never be reconstructed, and small codebooks typically introduce recognizable artifacts, including blocky images, metallic or tinny speech, and temporal flickering in video. 
Improved tokenizers and residual codecs~\citep{unknown2025imprvectquan, unknown2025discresidiff} can mitigate these issues, but the fundamental upper limit remains unchanged. 
Second, artifact patterns are linked to the topology of the codebook: since transitions operate over a fixed set of code indices, the model’s effective "neighbors" during denoising are determined by which codes lie nearby in the codebook. 
If the codebook is not smooth with respect to the underlying signal space, this can lead to unnatural transitions. 
Overall, these findings indicate that tokenized discrete diffusion is best understood as a complement to continuous latent diffusion, excelling in scenarios where editing, infilling, integration with other discrete modalities, or highly parallel decoding are the primary goals, and underperforming in cases where ultimate output quality depends on perceptual fidelity that the tokenizer is not yet able to maintain.

\subsection{Proteins and Biological Sequences} \label{subsec:app_protein}
Proteins are perhaps the most natural fit for discrete diffusion: a protein is a sequence over the twenty-letter amino-acid alphabet, so generation reduces to denoising in a small, well-defined categorical space whose "tokens" already carry biochemical meaning. 
The dominant approach adapts masked and absorbing-state diffusion to this alphabet, treating protein design as conditional sequence modeling under stability, motif, and functional constraints. 
DPLM established that a discrete diffusion objective yields a versatile protein language model whose representations transfer across generative and predictive tasks~\citep{wang2024diffusion}, and its successor DPLM-2 extends this to a multimodal model that jointly generates sequence and structure tokens~\citep{wang2025dplm2multimo}. 
Guided discrete diffusion provides a general mechanism for steering generation toward desired properties by combining a sequence-space denoiser with a property predictor~\citep{gruver2023proteindesig}, an idea instantiated across antibody design~\citep{luo2022antigenspeci, unknown2024antidesiusin, unknown2025antidesiopti, unknown2025genefounmode}, therapeutic-peptide generation under multiple objectives~\citep{unknown2025peptdenovo, unknown2025dflomultflow}, combinatorial functional design~\citep{unknown2025cfpgcombfunc}, membrane proteins~\citep{unknown2025tokeguiddisc}, and even the problem of shrinking a protein while preserving its function~\citep{unknown2025diffmodeshri}.

A second major thread conditions generation on three-dimensional structure. 
Inverse folding, recovering a sequence that folds to a target backbone, has been cast as discrete flow matching over the amino-acid alphabet~\citep{campbell2024allatominver, unknown2025allaprotsequ} and improved through mask-prior and denoising strategies that inject structural priors into the corruption process~\citep{unknown2025maskdenodiff}. 
Other work conditions on geometric constraints for backbone inpainting~\citep{unknown2025difflangdiff}, generates protein conformational ensembles via structure language models~\citep{unknown2025strulangmode}, and frames property optimization itself as a denoising problem over pretrained protein language models~\citep{unknown2025depldenoprot}. 
Robustness to distribution shift, which is critical when designs are pushed beyond the training distribution, has been addressed through context-guided diffusion~\citep{unknown2025contdiffouto}.

Evaluation in this domain is distinctive because the ground truth is ultimately physical. 
In-silico proxies (sequence perplexity, recovery rate against native sequences, motif accuracy, and predicted structural metrics such as scTM and pLDDT from folding models) are standard for rapid iteration, but the most meaningful signals come from downstream structure predictors and, where available, wet-lab measurements of stability, binding, and function. 
We refer readers to a recent survey dedicated to controllable protein sequence design for a broader taxonomy of conditioning mechanisms and benchmarks~\citep{unknown2025geneaicont}.

\subsection{Genomics (DNA/RNA) and Regulatory Sequences} \label{subsec:app_genomics}
Nucleic-acid sequences present a different regime from proteins: the alphabet is even smaller (four nucleotides), but the functionally relevant signals (regulatory grammar, splicing motifs, secondary structure) depend on long-range interactions that span hundreds or thousands of positions. 
This is precisely where global, bidirectional refinement is attractive, since regulatory elements can be coordinated across the whole sequence rather than committed left to right~\citep{sun2026training, yuan2026diffusionlargelanguagemodels}. 
Dirichlet flow matching adapts continuous flow-matching machinery to the probability simplex over nucleotides and demonstrates controllable DNA-sequence design, including generation conditioned on target regulatory activity~\citep{stark2024dirichlet}.

Beyond DNA, discrete diffusion has been applied to RNA secondary-structure prediction, where the reverse process refines a discrete base-pairing representation~\citep{unknown2025rnadgenerna}, and to three-dimensional RNA inverse folding, where a hyperbolic discrete diffusion model captures the hierarchical geometry of functional RNA~\citep{unknown2025hypediscdiff}. 
SE(3)-equivariant discrete diffusion has been used to jointly generate sequence and structure for nucleic-acid and protein complexes~\citep{unknown2025towajoinsequ}. 
At the single-cell and functional-genomics scale, masked discrete diffusion jointly models cell identity and expression~\citep{unknown2025scdimaskdisc}, while unifying multimodal masking frameworks~\citep{unknown2025nonaunifmult} and large-scale RNA foundation models~\citep{unknown2024largfounmode} point toward general-purpose generative models of regulatory sequence. 
A recurring practical consideration across these applications is measurement noise: experimental readouts of expression, accessibility, or structure are themselves noisy, so conditioning signals must be treated as soft targets rather than hard constraints.

\subsection{Molecules and Graph Generation} \label{subsec:app_molecule}
Molecules and, more generally, graphs are an intrinsically discrete and combinatorial domain: a molecular graph is defined by categorical node attributes (atom types) and categorical edge attributes (bond types, including the absence of a bond). 
Discrete diffusion is therefore a natural fit, and DiGress, which corrupts node and edge categories with independent transition matrices and learns a graph denoiser, is the canonical instantiation~\citep{vignac2023digressdiscr}. 
In four-component terms, the graph setting fixes the state space (categorical node/edge attributes with hard validity) and a joint node-edge corruption, and most subsequent work varies one component at a time; the recurring design lesson is that the binding constraint here is not denoiser capacity but \emph{validity}, which is enforced through the sampler (guidance, projection, constrained decoding) far more than through the corruption operator.
Organized by which component they modify, a large body of work refines this setup: on the \emph{corruption} side, continuous-time variants \citep{unknown2025comecontdisc, unknown2024disccontdiff} and Schr\"odinger-bridge or auto-encoder reformulations \citep{unknown2025discdiffschr, unknown2025discgrapauto}; on the \emph{sampler/ordering} side, autoregressive and layerwise orderings that improve scalability \citep{kong2023autoregressi, unknown2024pardpermauto, unknown2025layelayeauto, unknown2025learautomode, unknown2025genediregrap}; on the \emph{state space} itself, higher-order and hypergraph generation \citep{unknown2026hogdhighguid, unknown2026hygediffhype, unknown2026hypegrapdiff}; on the \emph{denoiser/representation} side, sparse and degree-guided formulations for large graphs \citep{chen2023efficient, unknown2025sparspardisc, unknown2025spartraidisc, unknown2025grapcandiff, unknown2025larggenegrap, unknown2025crosgrapdata} and stability-oriented recastings of the denoising process \citep{unknown2025simpcrititer, unknown2025grapgenediff, unknown2025advagrapgene}; and on the \emph{systems} side, post-training quantization for efficiency \citep{unknown2025outlpostquan}.

A central challenge specific to 2D molecular graph generation is \emph{validity}: a randomly denoised graph need not correspond to a chemically valid molecule.
Much of the 2D literature therefore focuses on enforcing valence rules, connectivity, and property constraints, either through guidance or through constrained sampling.
MolDiff explicitly addresses the atom--bond inconsistency that arises when atoms and bonds are generated independently~\citep{peng2023moldiffaddre}.
For 3D structure-based drug design, the modeling bottleneck is different: many methods generate atom types and coordinates while chemical bonds are inferred or reconstructed afterward, so the main challenges include geometric equivariance, target-aware conditioning, steric and interaction constraints, and the limited flexibility caused by fixing the molecular size or atom count during generation.
A range of methods incorporate three-dimensional and target-aware structure for structure-based drug design~\citep{guan20233dequivarian, guan2024aligning, unknown2025bindbondinte, unknown2025genemoleevol}, fragment- and motif-level tokenizations that build in chemical validity by construction~\citep{campbell2024fragfmeffici, unknown2025molefragdiff, unknown2025genmdrugdisc}, and linker design for fragment joining~\citep{igashov2024equivariant}.
Conditional and multi-conditional generation has been pursued through Bernoulli and discrete-graph guidance~\citep{unknown2025grapintecont, unknown2026conddiffbase}, graph diffusion transformers~\citep{peng2024graphdiffusi, unknown2026grapdifftran}, composable score-based conditioning~\citep{unknown2025compscorgrap}, and unified frameworks for dual-target molecule generation~\citep{unknown2025unifconddiff}.
Language-to-molecule translation connects text prompts to molecular generation~\citep{unknown2024langdiffgene, unknown2026textmolegene}, and plug-and-play projection enables controllable generation without retraining~\citep{qin2024diffusesampl}.

The same machinery extends naturally to adjacent problems where the object of interest is a discrete structure scored by an external objective. 
Retrosynthesis, predicting reactants from products, has been framed as categorical diffusion and discrete flow matching~\citep{unknown2024diffcatediff, unknown2025retrsynfdisc, unknown2025advaretrretr}; mass-spectrum-conditioned generation recovers molecules from spectra~\citep{unknown2025diffdiffgenea}; and inorganic-materials and crystal generation incorporate symmetry and electronic-structure constraints~\citep{zeni2024mattergen, unknown2025wyckgenediff, unknown2025inteelecstru, unknown2025perimategene}. 
Discrete graph diffusion has further been applied to protein co-design~\citep{campbell2024generative}, neural-architecture search~\citep{unknown2025multgrapdiff}, graph-edit-distance computation~\citep{unknown2025diffcompgrap}, end-to-end path planning~\citep{unknown2025grapdiffendt}, social-graph generation~\citep{unknown2025priodiscdiff}, node classification~\citep{unknown2025redirepamask}, GNN explanation~\citep{unknown2023d4exindiexpl}, and de novo drug design more broadly~\citep{unknown2025genedeeplear, unknown2025advagrapgene}.
Viewed through the optimization lens of Section~\ref{sec:discussion}, property-guided molecular generation is naturally read as search over a discrete configuration space for high-scoring structures under a learned predictor. 
Finally, as graph diffusion models mature, questions of fairness~\citep{unknown2024fairfairgrap} and security, including watermarking~\citep{unknown2026checwategrap} and backdoor attacks~\citep{unknown2025backattadisc}, have begun to receive attention.

\subsection{Planning, Agents, and Decision Making}
\label{sec:agents_planning}

%
Planning, agents, and decision-making applications of discrete diffusion have recently emerged across language agents, embodied control, multimodal reasoning, and combinatorial optimization settings~\citep{deng2025thinkwhile, kaplan2025implicit, unknown2025discdiffvla, unknown2025plandiffmode, ye2025beyondautore, zhang2024copilot4d, sanokowski2025scaldiscdiff, sun2023difuscograph}. 
DiscreteRTC further exploits the native inpainting capability of discrete diffusion policies for asynchronous embodied control, carrying over committed actions while iteratively generating the remaining action chunk so that planning and execution can proceed concurrently~\citep{wang2026discretertc}.
These works explore how iterative refinement in discrete spaces can be leveraged for planning, structured action generation, and constraint-aware decision-making. We organize this emerging area along four themes.

\paragraph{Diffusion agents as planners.}
A central distinction between diffusion-based agents and conventional autoregressive (AR) agents lies in how plans or reasoning traces are constructed. 
AR agents generate sequences left-to-right, committing to earlier decisions without the ability to revise them, which can lead to error accumulation in long-horizon reasoning or planning tasks. 
In contrast, discrete diffusion models generate plans through iterative global refinement, allowing later context to influence and revise earlier tokens. 
Recent works explore this idea in both language-based and multimodal agents~\citep{deng2025thinkwhile, kaplan2025implicit, ye2025beyondautore, unknown2026discdiffrefl, unknown2025lladvisilang}. 
For example, diffusion-based reasoning frameworks generate intermediate plans that are repeatedly refined, enabling implicit backtracking and correction. 
In embodied or vision-language settings, diffusion planners have been used to iteratively refine action sequences or trajectories conditioned on observations~\citep{unknown2025discdiffvla, unknown2025dvladiffvisi, unknown2025vilalargvisi, unknown2025dvlmenhadiff}. 
It is important to emphasize that this “global revision” capability is a \emph{potential advantage} rather than a guaranteed outcome: its effectiveness depends on the training objective, noise schedule, and inference strategy (e.g., remasking policies or guidance). 
Without appropriate design, diffusion models may still exhibit local inconsistencies or converge slowly.

\paragraph{Offline RL \& trajectory generation.}
Discrete diffusion provides a natural framework for modeling trajectories in offline reinforcement learning (RL), where sequences of states, actions, and optionally rewards are tokenized and treated as structured discrete objects. 
Instead of learning a policy directly, diffusion models learn a generative distribution over trajectories, which can then be sampled or guided toward high-reward behaviors. 
Several works demonstrate that diffusion over trajectories can act as a stochastic planner that improves upon suboptimal datasets~\citep{unknown2025reinleardisc, unknown2026dichdiffpoli, unknown2026unifdiffvla, unknown2025hybrdiffsimu}. 
By conditioning on reward signals or value estimates, the model can generate trajectories that are unlikely under the behavior policy but achieve higher returns. 
This connects to broader perspectives of diffusion as an optimization process in discrete spaces, where denoising corresponds to iteratively improving candidate solutions. 
In robotics and control settings, diffusion-based trajectory planners have also been explored for structured action generation and robustness~\citep{unknown2025plandiffmode, unknown2026plandiff, unknown2025grapdiffrobu}. 
However, practical performance depends critically on how trajectories are tokenized and how constraints (e.g., dynamics feasibility) are enforced during sampling.

\paragraph{Tool use \& API calls.}
Modern language agents frequently rely on structured tool use, such as generating JSON outputs or function calls that must satisfy strict schemas. 
Discrete diffusion is well-suited for such settings because it operates directly over categorical tokens and supports constraint-aware decoding. 
Diffusion-based approaches to tool use and structured generation highlight two key advantages.
First, global context can be used to jointly refine all fields in a structured output, rather than committing to arguments sequentially as in AR decoding. 
Second, constraint enforcement (e.g., valid JSON syntax or schema compliance) can be integrated into the denoising process via masking, projection, or guided sampling~\citep{zhang2024copilot4d, unknown2025effidiffnona, unknown2025wamdmaskdiff}. 
Recent systems explore diffusion-based language agents that interleave reasoning and tool invocation, enabling iterative correction of tool arguments and outputs~\citep{unknown2025dreadreaopen, unknown2025unsutraidiff}. 
Nevertheless, reliability in practice depends not only on the generative model, but also on external constraint mechanisms such as validators, execution feedback, and verification loops.
Diffusion alone does not guarantee correctness without such auxiliary components.

\paragraph{Combinatorial optimization.}
Many planning problems can be formulated as combinatorial optimization over discrete configurations, such as traveling salesman problems (TSP), scheduling, or graph construction.
Discrete diffusion offers an alternative to classical search methods by treating solutions as tokens that are iteratively refined. 
In this view, diffusion acts as a stochastic refinement process over candidate solutions, analogous to local search but with global context and learned transition dynamics. 
Applications include graph optimization, routing, and structured design problems~\citep{sun2023difuscograph, sanokowski2025scaldiscdiff, campbell2024wamflowparal}. 
Compared to traditional approaches such as beam search, evolutionary algorithms, or simulated annealing, diffusion models can learn problem-specific priors and produce diverse candidate solutions in parallel. 
However, feasibility constraints remain a central challenge. 
Many works incorporate repair operators, constraint-aware transitions, or projection steps to ensure validity during denoising~\citep{unknown2024diffmodefram, unknown2025plandiffmode}. 
The effectiveness of diffusion in combinatorial settings depends on how constraints are encoded, either implicitly through training data or explicitly through inference-time mechanisms.

Across these domains, discrete diffusion reframes planning and decision making as iterative refinement in structured discrete spaces. 
Its main promise lies in global consistency, flexibility in conditioning, and compatibility with constraints. 
At the same time, its advantages are contingent on careful design of tokenization, corruption processes, and inference strategies, and should be viewed as complementary to, rather than replacements for, autoregressive and classical planning methods.

\subsection{Tabular Data and Imputation} \label{subsec:app_tabular}
Tabular data sit at an interesting boundary for discrete diffusion: a single row mixes categorical fields with numerical fields, so a generative model must handle both modalities coherently. 
Categorical columns are a natural fit for the discrete machinery developed throughout this paper, each column is simply a small vocabulary, and the forward process corrupts category labels exactly as it does tokens. 
Numerical columns require an additional decision: they can be discretized into bins and absorbed into the discrete state space, or modeled with a continuous diffusion process running alongside the discrete one in a hybrid design. 
Early work such as TabDDPM took the latter route, combining Gaussian diffusion for numerical features with multinomial diffusion for categorical features under a shared schedule~\citep{kotelnikov2023tabddpmmodel}. 
More recent models pursue tighter integration of the two modalities: TabDiff formulates a mixed-type, multi-modal diffusion process that jointly generates numerical and categorical columns~\citep{unknown2025tabdmixediff, unknown2025tabdunifdiff}, TabNAT couples continuous and discrete generation in a single non-autoregressive framework~\citep{unknown2025tabncontjoin}, and Unmasking Trees casts tabular generation as a masked-prediction problem solved with tree-based predictors~\citep{unknown2025unmatreetabu}.

Missing-value imputation is a particularly natural application, because it maps directly onto the masking view of discrete diffusion: the observed entries of a row play the role of unmasked context, the missing entries play the role of masked positions, and the reverse process fills them in conditioned on the rest of the row. 
This framing requires no special-purpose architecture, imputation and generation are the same task with different masking patterns, and it accommodates arbitrary, non-contiguous missingness, since any subset of cells can be designated as the target. 
Evaluation in this domain attends to three distinct criteria that are sometimes in tension. 
\emph{Fidelity} measures how closely generated columns and their joint correlations match the real data distribution; \emph{privacy} asks whether synthetic rows leak information about real records (e.g., via membership-inference or distance-to-closest-record analyses), which is a primary motivation for synthetic tabular data in the first place; and \emph{downstream utility} measures whether models trained on synthetic or imputed data perform comparably to those trained on real data. 
A recurring theme is that optimizing fidelity alone can degrade privacy, so the most useful evaluations report all three together rather than a single aggregate score.

\subsection{Other Case Studies} \label{subsec:app_casestudies}

%
Beyond the applications outlined above, discrete diffusion model is also employed in settings where data are inherently discrete or can be represented via compact codebooks. 
We treat these as case studies rather than applications. 
While they frequently reuse modeling strategies from text or media generation, their unique constraints, evaluation criteria, and typical failure patterns warrant separate treatment. 
We close by providing a standardized checklist to guide the reporting of design choices in future case studies.

\paragraph{Layout and document generation.}
Document, slide, GUI, and scene layouts are an obvious early application area for discrete diffusion, because layout components have categorical types and quantized coordinates. 
LayoutDM \citet{inoue2023layoutdm} introduces the standard approach of modeling element classes and bounding-box bins jointly as categorical variables, evolved under masked or absorbing transitions. 
Subsequent research develops this template along three main directions: \emph{conditioning} on partial layouts or predefined graph structures~\citep{unknown2023dltcondlayo, unknown2023layoimprgrap}; \emph{correction} of layout-decoder-specific failure modes, including layout sticking and pattern collapse~\citep{unknown2025layoallelayo, unknown2025difflayopatt}; and \emph{specialization} to particular application domains such as floor plans, vector graphics, and landmark-based layouts~\citep{unknown2025consvectfloo, unknown2025vqcacompdesi, unknown2026latolanddiff}. 
A common finding is that layout-oriented metrics—like overlap, alignment, and validity—pose challenges for greedy AR decoders, while iterative remasking enables the model to jointly manage both global alignment and fine-grained local placement.

\paragraph{3D assets generation.}
Representing 3D content as sequences of discrete primitives—such as part labels, voxel or cube indices, or VQ shape tokens—allows 3D generation to leverage the same masked denoising architectures. 
3DQD~\citep{unknown2023genedeep3d, unknown20253dqdgenedeep} discretizes shapes at the part level.
Cube-Absorb discrete diffusion~\citep{unknown2025largscengene} employs absorbing transitions to cope with the very long token chains found in large scenes. 
Mixed Diffusion for 3D Indoor Scenes~\citep{unknown2025mixediff3d} combines continuous and discrete variables to model object placement, while PartDiffuser~\citep{unknown2025partpart3d} synthesizes 3D meshes part by part. 
Complementary research investigates 2D-to-3D alignment~\citep{unknown2023unal2d3d}, scan-based generative modeling~\citep{unknown2023scandiffmode}, and instance-level 2D-to-3D representation learning~\citep{unknown2025instinst2d}. 
A key design trade-off across these methods lies between having an expressive codebook rich enough to capture geometric detail and the resulting long token sequences, which in turn motivate block-wise or absorbing diffusion schedules.

\paragraph{Sketches, fonts, and stylized media.}
Discrete diffusion is also effective for structured creative tasks in which the components are small and well-defined. 
Representative applications include sketch-conditioned image inpainting~\citep{sharma2024sketimaginpa}, joint sketch-and-context generation~\citep{unknown2025sketjoincont}, glyph-style synthesis~\citep{unknown2026diffglypstyl}, high-quality font creation~\citep{unknown2025qtfohighfont}, Chinese calligraphy synthesis and recognition within a unified diffusion framework~\citep{unknown2026unicunifdiff}, and arbitrary style transfer~\citep{unknown2025dstyladva}. 
In these scenarios, the approach retains the parallel decoding and editability benefits of image-token diffusion, but operates in a smaller, more tightly constrained code space. 
This makes it easier to enforce properties such as legibility or stroke order compared to diffusion carried out directly in pixel space.

\paragraph{Recommendation and information retrieval.}
A growing body of work treats user–item interaction histories as discrete sequences, enabling the application of the same masked-denoising framework commonly used in NLP. 
Within this view, sequential recommendation has been reformulated as fuzzy modeling in a discrete state-space diffusion framework~\citep{unknown2024breadetefuzz}; generative recommendation has been implemented via LLaDA-Rec~\citep{unknown2025lladdiscdiff} and DiffGRM~\citep{unknown2025diffdiffgene}; and further extensions targeting reranking, bundle generation, and continuous-time recommendation continue to elaborate and refine this overarching pipeline~\citep{unknown2025discconddiff, unknown2026discdiffbund, unknown2025contdiscdiff}. 
Closely related approaches apply discrete diffusion at the retrieval stage in graph-based RAG through query-aware flow diffusion~\citep{unknown2026querflowdiff}, and for imputing graph features using fractional subgraph diffusion~\citep{unknown2026fsdcfracsubg}. 
These research avenues are closely related to the planning and decision-making subsection (Section~\ref{sec:agents_planning}). 
Consequently, we only summarize them briefly here and refer the reader to that subsection for planning-oriented applications of discrete diffusion.

\paragraph{Specialized scientific and structured domains.}
A final group of case studies highlights discrete diffusion beyond standard image and text modalities. 
Biomedical imaging is explored through LLaDA-MedV~\citep{nie2025lladamedv} and via discrete-diffusion methods for prostate MRI super-resolution~\citep{unknown2025prosmrisupe, unknown2025discresidiff}. 
In wireless communications, residual vector quantization is applied to channel symbols~\citep{unknown2025resivectquan}, and spiking neural networks have been integrated with VQ-based discrete diffusion to enable low-power generative models~\citep{unknown2025spikvectquan}.
Structured knowledge generation is addressed by DiSK~\citep{unknown2025diskdiffmode}; LEAF~\citep{unknown2025leaflarglang} reformulates time series forecasting as a diffusion-based language model; and object-centric representation learning is advanced through masked generative modeling~\citep{unknown2025effiobjerepr}. 
Vision–language applications that go beyond straightforward generation also fall into this category. 
These include image captioning~\citep{unknown2025expldiscdiff}, caption editing~\citep{unknown2025decatowagene}, segmentation refinement~\citep{unknown2023segrtowamode}, remote-sensing image captioning~\citep{unknown2025discdiffmodea}, defect detection~\citep{unknown2025towabettgene}, source-mask optimization for lithography~\citep{unknown2025ultrsourmask}, uncertainty-aware Bernoulli diffusion~\citep{unknown2025unceberndiff}, and guided discrete diffusion for scientific applications~\citep{unknown2025guiddiscdiff}. 
Although these domains differ substantially, prior work follows a similar template: a pretrained or manually crafted tokenizer converts the domain-specific signal into a discrete set of tokens, and a categorical denoiser then performs parallel, editable, and constraint-aware generation over this token vocabulary.

\paragraph{A checklist for new case studies.}
Based on these examples, we suggest that future case studies report at least the following elements, to enable consistent comparison of results across domains. 
This checklist is an authors' recommendation rather than an established community standard; we offer it to improve cross-domain comparability, not as a description of current reporting practice.
(1) The tokenization setup, including codebook size, level of granularity, and pretraining strategy. 
(2) The corruption process, specifying whether mask, uniform, or absorbing transitions are used, along with the noise schedule. 
(3) The training objective, for example cross-entropy, ELBO, score-entropy, or a flow-matching variant. 
(4) The model architecture, including how it incorporates auxiliary modalities. 
(5) The sampling procedure, including the number of sampling steps, the remasking strategy, and any guidance mechanisms. 
(6) Constraint mechanisms, such as validators, projection methods, or classifier-based guidance. 
(7) The evaluation protocol, including domain-specific metrics for cases where standard diffusion metrics are not applicable. 
(8) Typical failure modes, such as sticking, repetition, codebook collapse, or violations of validity constraints. 
Describing results along these dimensions facilitates cross-domain comparison and clarifies which design choices generalize across case studies and which remain specific to a particular domain.

%% file: tables/07-applications.tex
\begin{table}[t]
\centering
\caption{Application domains for discrete diffusion, organized by how the discrete state space arises. "Why diffusion" names the property that most motivates a diffusion approach in each domain. The evidence maturity column indicates whether the current literature is broad and benchmark-rich, active but domain-specialized, or still comparatively emergent.}
\label{tab:applications_overview}
\small
\setlength{\tabcolsep}{3pt}
\renewcommand{\arraystretch}{1.15}
\begin{tabular}{@{}p{3cm}p{4cm}p{3cm}p{2.65cm}p{3.0cm}@{}}
\toprule
\textbf{Domain} & \textbf{Token space} & \textbf{Why diffusion} & \textbf{Key challenge} & \textbf{Evidence maturity} \\
\midrule
Text \& code & Subword / byte & Infilling, editing, global revision & Streaming, ICL gap & Established: large-scale systems and standard benchmarks \\
\midrule
Multimodal \& media & VQ / codec codes & Unified decoder, parallel editing & Tokenizer fidelity ceiling & Established / active: mature media metrics, rapidly evolving systems \\
\midrule
Proteins \& genomics & Amino acids, nucleotides & Long-range co-design, constraints & Physical / wet-lab validity & Active but specialized: domain metrics and validation bottlenecks \\
\midrule
Molecules \& graphs & Atom/bond categories & Native combinatorial structure & Chemical validity & Active but specialized: strong validity metrics, task-specific benchmarks \\
\midrule
Planning \& agents & Action / state tokens & Revisable global plans & Constraint feasibility & Emerging: promising but preliminary evidence \\
\midrule
Tabular & Mixed categorical/numeric & Imputation as masking & Privacy vs.\ fidelity & Emerging: task- and dataset-specific evidence \\
\bottomrule
\end{tabular}
\end{table}

%% file: sections/10-evaluation.tex
\section{Evaluation and Benchmarks} \label{sec:evaluation}

%
This section reviews how discrete diffusion models are evaluated and why several autoregressive evaluation habits do not transfer unchanged. Section~\ref{subsec:eval_likelihood} discusses likelihood, perplexity, and calibration. 
Section~\ref{subsec:eval_quality} surveys quality metrics across text, media, science, and agentic domains. 
Section~\ref{subsec:eval_speed} addresses speed, cost, and efficiency metrics, arguing for quality--latency frontier reporting, and Section~\ref{subsec:eval_editing} treats editing, infilling, and conditional fidelity.

\subsection{Likelihood, Perplexity, and Calibration} \label{subsec:eval_likelihood}

Likelihood, perplexity, and calibration are central but subtle evaluation dimensions for discrete diffusion language models. 
Unlike autoregressive (AR) models, whose likelihood is given by a left-to-right product of next-token probabilities, diffusion models define generation through a corruption-denoising process. 
As a result, likelihood evaluation depends on the chosen formulation, parameterization, and sampler. 
Some models admit tractable or efficiently estimable variational bounds, while others are primarily evaluated by denoising losses, task metrics, or sample quality. 
This makes likelihood reporting useful but also easy to misinterpret, especially when comparing diffusion language models with AR baselines.

\paragraph{Likelihood reporting.}
For AR language models, evaluation commonly reports negative log-likelihood or perplexity under the factorization $p_{\theta}(x)=\prod_{i=1}^{L}p_{\theta}(x_i\mid x_{<i})$. 
This provides a direct token-level measure of predictive performance under teacher forcing. 
Discrete diffusion models generally do not use this factorization. 
Instead, they define a forward corruption process $q(x_t\mid x_0)$ and a learned reverse process $p_{\theta}(x_{t-1}\mid x_t)$, leading to likelihood objectives based on variational lower bounds or continuous-time analogues. 
In D3PM-style models, the evidence lower bound decomposes into reconstruction, prior, and timestep-wise KL terms, where the reverse model is trained to approximate the posterior $q(x_{t-1}\mid x_t,x_0)$~\citep{austin2021structured,hoogeboom2021argmaxflows}. 
In masked diffusion language models, the objective often simplifies to a reweighted masked-token prediction loss, which can be interpreted as a continuous-time or discrete-time variational bound under an absorbing-state corruption process~\citep{sahoo2024simpleeffect,shi2024simplified,lou2025yourabsorbin}.

Because of this difference, diffusion likelihoods and AR perplexities are not automatically apples-to-apples. 
First, many diffusion models report a variational bound rather than exact likelihood. 
The bound may be loose, and its tightness depends on the number of timesteps, the weighting scheme, the parameterization, and whether the reverse posterior is computed exactly or approximated. 
Second, likelihood units may differ across papers: some report bits per dimension, bits per token, negative log-likelihood, ELBO, or perplexity derived from a bound. 
Third, tokenization can change the apparent value of perplexity. 
A model evaluated with byte-level tokens, BPE tokens, or a different vocabulary size may have substantially different token-level perplexity even if the underlying text distribution is similar. 
For cross-tokenizer comparison, normalized quantities such as bits-per-byte or bits-per-character are often more meaningful than raw token perplexity.

A further complication is that likelihood and generation quality may diverge. 
In AR modeling, lower perplexity often correlates with stronger language modeling, although there is not a complete measure of instruction-following or human preference. 
For diffusion language models, the mismatch can be larger because the training objective measures denoising quality across corruption levels, whereas generation quality depends on the full reverse trajectory and the sampler. 
A model may achieve a strong denoising loss but still produce lower-quality samples if the sampler commits tokens too aggressively, accumulates errors across denoising steps, or uses a poor remasking schedule. 
Conversely, inference-time techniques such as confidence remasking, guidance, or test-time scaling can improve sample quality without changing the underlying likelihood bound~\citep{arriola2025remasking,nisonoff2025simpleguidan,unknown2025effetestscal}. 
Therefore, likelihood should be reported together with sample quality and decoding-budget metrics rather than used as the sole evaluation criterion.

For fair comparison with AR baselines, we recommend that papers state explicitly: (1) whether the reported value is an exact likelihood, an ELBO, an upper bound on perplexity, or a surrogate denoising loss; (2) the tokenization scheme and normalization unit; (3) the number of diffusion steps used for likelihood estimation if applicable; and (4) whether the same data preprocessing and context length are used for AR and diffusion models. 
When possible, diffusion models should report both likelihood-style metrics and task-level generation metrics, since these capture complementary aspects of model behavior. 
Recent work on likelihood-based diffusion language models and efficient perplexity bounds provides useful tools for making such comparisons more principled~\citep{shi2024likelihoodba,unknown2025effiperpboun,lou2024discrete}.

\paragraph{Calibration.}
Calibration measures whether a model's predicted probabilities correspond to empirical correctness. 
For discrete diffusion, calibration is important at two levels: the calibration of token posteriors at each denoising step, and the calibration of the final generated sequence. 
At a given timestep $t$, the model predicts a distribution over clean tokens, reverse states, or score/ratio values. 
If the model assigns high confidence to incorrect predictions, the sampler may prematurely freeze wrong tokens; if it is under-confident, the sampler may repeatedly remask correct tokens and waste computation. 
Thus, calibration directly affects the efficiency and reliability of iterative generation.

This connection is especially clear in confidence-based remasking. 
Many masked diffusion samplers generate candidate tokens for masked positions, estimate confidence from token probabilities or entropy, freeze high-confidence tokens, and remask low-confidence ones. 
The effectiveness of this strategy depends on whether confidence is well calibrated across positions and timesteps. 
Ideally, high-confidence predictions should be likely to be correct, and the confidence scale should remain comparable as the corruption level changes. 
In practice, calibration may vary substantially across timesteps: early denoising steps operate under high uncertainty and broad context ambiguity, while later steps condition on more fixed tokens and may become overconfident in locally fluent but globally inconsistent predictions.
Miscalibration can lead to error locking, where an incorrect token is frozen early and then treated as reliable context in later denoising rounds.

Calibration also interacts with masking schedules and selective refinement. 
A well-calibrated model can support adaptive computation: easy tokens can be fixed early, while uncertain spans receive more denoising steps. 
This is important for efficient generation because diffusion models trade quality for the number of refinement rounds. 
If token confidence accurately predicts correctness, the sampler can allocate computation non-uniformly across the sequence, focusing on difficult regions such as rare entities, syntactically constrained spans, or long-range dependencies. 
Recent methods that use remasking, adaptive block sizes, local determinism, or entropy-bounded unmasking are viewed as attempts to improve the calibration of intermediate predictions~\citep{arriola2025remasking,unknown2025adabsemadiff,unknown2025accediffllm,unknown2025accesampmask}.

Calibration should therefore be evaluated explicitly rather than assumed from likelihood or sample quality. 
Standard tools include reliability diagrams, expected calibration error (ECE), negative log-likelihood of token predictions at each timestep, entropy-accuracy curves, and selective prediction metrics. 
For diffusion language models, these metrics can be computed separately for different corruption levels, token types, and decoding iterations. 
Timestep-wise calibration curves are particularly informative: they reveal whether the model is overconfident early in generation, underconfident near convergence, or poorly calibrated only under certain masking ratios. 
In conditional generation, calibration can also be measured with respect to constraint satisfaction, e.g., whether high-confidence tokens are more likely to preserve required entities, attributes, or fixed spans.

Beyond token-level calibration, sequence-level uncertainty remains an open evaluation problem. 
Since diffusion models generate through multiple stochastic refinement steps, uncertainty arises not only from the final token distributions but also from trajectory-level choices such as masking order, resampling, and guidance strength. 
Two samples with similar final token probabilities may have very different denoising histories: one may converge steadily, while another may oscillate across several refinements. 
Such trajectory information can be useful for detecting unstable generations, hallucinations, or constraint violations. 
Reporting convergence diagnostics, remasking rates, entropy decay, and agreement across multiple denoising trajectories may therefore provide a richer picture of reliability than final likelihood alone.

In summary, likelihood and perplexity remain important for evaluating diffusion language models, but they must be interpreted carefully. 
Diffusion models often report variational bounds or denoising-based surrogates rather than AR-style likelihoods, and comparisons are sensitive to tokenization, normalization, and inference procedure. 
Calibration is equally important because intermediate confidence estimates determine which tokens are frozen, remasked, or refined. 
A comprehensive evaluation should therefore combine likelihood-style metrics, task-level generation quality, timestep-wise calibration, and sampler-dependent diagnostics.

\subsection{Quality Metrics Across Domains} \label{subsec:eval_quality}

\paragraph{Text.} 
Evaluating generation quality in dLLMs requires sophisticated and task-specific metrics. 
Early n-gram overlap statistics such as BLEU~\citep{papineni2002BlueMethAuto} and ROUGE~\citep{lin2004RougPackAuto} are systematically ill-suited to non-autoregressive generators: they reward lexical proximity to a single reference, thereby penalizing the diverse outputs that iterative stochastic denoising naturally produces. 
\citet{tarim2025canyoudete} illustrates that a generation can be semantically correct, factually grounded, and globally coherent yet score poorly simply because its wording diverges from the reference. 
They show that a dLLM can achieve equal or higher BLEU and ROUGE scores than a comparably-sized AR model while simultaneously exhibiting stylometric properties such as burstiness and perplexity under an external LM that cause standard AI-generated-text detectors to fail. 
This demonstrates that single surface metrics, applied without awareness of how diffusion generation differs architecturally from autoregressive generation, therefore risk producing misleading assessments of output naturalness.

Task-based evaluation on closed-form benchmarks, where correctness does not depend on reference-matching, is therefore the preferred primary quality signal for dLLMs. 
These benchmarks yield binary or ordinal correctness signals across diverse task families: mathematical reasoning, general knowledge, and code generation evaluated by execution. 
When closed-form correctness is insufficient, they should be complemented by instruction-following benchmarks, factuality-sensitive evaluations, and human preference comparisons. 
\citet{zhang2025diffautolang} open a further evaluative axis orthogonal to generation: they demonstrate that the bidirectional pretraining of dLLMs can be assessed through downstream text embedding tasks such as long-document and reasoning-intensive retrieval, where bidirectional context provides advantages that generation-focused benchmarks do not expose.

A further dLLM-specific evaluation concern is the attribution and detection of diffusion-generated text. 
Standard AR watermarking schemes fail in this setting because they condition each new token on previously generated tokens. 
This is an assumption that dLLMs' arbitrary generation order breaks. 
Three concurrent works address this from different angles: \citet{gloaguen2026watedifflang} compute watermark signals in expectation over undetermined context; \citet{wu2025dmarordewate} introduce order-agnostic watermarking that operates with both predictive and bidirectional context strategies; and \citet{raban2025lrdweffiwate} propose biasing tokens on both their available left and right neighbours to retain near-zero runtime overhead. 
That three independent proposals were needed to rehabilitate a single AR evaluation mechanism underscores a broader point: evaluation tools calibrated for autoregressive generation cannot be applied to dLLMs without first verifying that their underlying assumptions still hold.

\paragraph{Media.} 
For tokenized image generation, the Fr{\'e}chet Inception Distance (FID; \citealt{heusel2017GansTraiTwo}) measures the Wasserstein-2 distance between Inception-v3 feature distributions of real and generated images and is the standard quality metric for image generation. 
For token-based discrete diffusion, a key limitation is that FID conflates two distinct sources of error: tokenizer reconstruction quality and generative model quality.
\citet{gu2024rethobjevect} shows that optimizing a VQ tokenizer for reconstruction does not guarantee better downstream generation quality, motivating the practice of reporting reconstruction metrics separately from generation FID. 
Alongside FID, Inception Score and CLIP score are used as complementary metrics for sharpness and text-image alignment respectively~\citep{salimans2016ImprTechTrai, radford2021LearTranVisu}. 
Evaluating the full encode-generate-decode pipeline additionally requires reconstruction-consistency metrics, such as LPIPS, to detect artefacts introduced by the mismatch between the tokenizer's training distribution and the generative model's output distribution.

Similar to FID for images, Fr{\'e}chet Video Distance (FVD; \citealt{unterthiner2019AccuGeneMode}) and Fr{\'e}chet Audio Distance (FAD; \citealt{kilgour2019FrecAudiDist}) extend the same distributional summary principle to video and audio respectively, using domain-appropriate feature networks. 
For video, a known limitation is that FVD is insensitive to certain temporal distortions, such as frame-level flickering introduced by codebook-boundary transitions in discrete video generation~\citep{yu2023magvitmasked}. 
For audio, FAD is used alongside perceptual metrics such as word error rate and speaker similarity to capture distinct quality dimensions that a single distributional metric cannot summarize~\citep{yang2023diffusion, pham2025mdsgfasteffi}.

Across all three modalities, no single metric captures the full range of failure modes in token-based discrete generation. 
A principled evaluation protocol therefore requires three coordinated measurements: tokenizer reconstruction fidelity evaluated independently of the generative model; downstream generation quality measured by the appropriate distributional metric; and reconstruction-consistency evaluation of the full encode-generate-decode pipeline.

\paragraph{Science and graphs.} 
Protein and genomic sequence generation require evaluation protocols that combine sequence plausibility with downstream functional or structural criteria.
For protein design, sequence recovery and perplexity are useful for inverse-folding benchmarks, but they are insufficient on their own because many sequences with low recovery may still fold correctly, while high-recovery sequences need not satisfy the intended design objective.
Consequently, structure-aware evaluation commonly includes refoldability or self-consistency checks using structure-prediction models, RMSD or TM-score to the target backbone or motif, confidence metrics such as pLDDT, diversity and novelty relative to natural proteins, and, when available, experimental stability or binding measurements~\citep{dauparas2022robust, watson2023denovo, wang2023pdbstruct}.
For genomics, validity is usually less about syntactic correctness, since any string over the nucleotide alphabet is formally valid, and more about biological plausibility and target activity.
Evaluation therefore emphasizes sequence statistics such as GC content and k-mer or motif composition, novelty relative to training sequences, conservation of regulatory grammar, and predicted or measured functional activity, including chromatin accessibility, promoter strength, enhancer activity, or gene-expression targets~\citep{avdeyev2023dirichlet, stark2024dirichlet, dasilva2024dnadiffusion}.
Thus, as in molecular generation, biological-sequence evaluation is layered: low-level sequence realism must be paired with task-specific structural or functional validation.

Molecular generation evaluation follows a constraint-first hierarchy. 
The validity, uniqueness, and novelty must be satisfied before considering any other distributional metric, since a model generating chemically invalid or memorized structures provides little scientific value regardless of its distributional scores. 
The MOSES benchmark~\citep{polykovskiy2020MoleSetsMOSE} and GuacaMol~\citep{brown2019GuacMolBenc} operationalize this with complementary emphases, with MOSES centering on Fr{\'e}chet ChemNet Distance (FCD) and scaffold statistics, and GuacaMol focusing on KL divergence across physicochemical property distributions.
\citet{vignac2023digressdiscr} establish these benchmarks as the standard evaluations for molecular discrete diffusion, demonstrating that the validity-uniqueness-novelty hierarchy combined with FCD exposes failure modes that neither metric type reveals in isolation.

For graph generation, the standard suite uses maximum mean discrepancy (MMD), a kernel-based distributional distance, on degree distributions, clustering coefficients, and orbit statistics.
Discrete diffusion's sparse intermediate states keep structural features well-defined throughout denoising, a property \citet{haefeli2022diffmodegrap} show reduces average MMDs relative to continuous noise. 
For combinatorial optimization, optimality gap is the primary criterion~\citep{sun2023difuscograph, sanokowski2025scaldiscdiff}, and its monotone improvement with denoising steps provides a quality-compute tradeoff with no direct AR analogue. 
This illustrates that scientific evaluation for discrete diffusion is not a single metric problem, but a layered framework where constraint satisfaction and distributional fidelity serve complementary diagnostic roles.

\paragraph{Agents and tool use.} 
For agentic tool use and code generation, execution-based evaluation is the most dominant quality signal. 
Unlike surface-based metrics, executing the generated output is reference-free and directly measures functional correctness. 
Pass@$k$ on coding benchmarks such as HumanEval~\citep{chen2021EvalLargLang} and MBPP~\citep{austin2021ProgSyntLarg} is the standard execution-based metric, measuring the probability that at least one of $k$ generated samples passes all unit tests. 
A dLLM-specific concern is that non-AR generation can produce syntactically malformed outputs even when the intended logic is correct. 
Syntactic validity rate must therefore be reported alongside pass@$k$ for dLLMs, as left-to-right AR generation enforces local syntactic consistency by construction.

For function-calling and API use, pass@$k$ alone is insufficient. 
A model can produce fluent but structurally invalid function calls that fail downstream execution entirely~\citep{suresh2025dingconsinfe}. 
Structured outputs must additionally be evaluated on schema-level metrics, including schema validity, field completeness, and type compliance. 
Benchmarks such as BFCL \citet{patil2025bfcl} provide standardized harnesses for these metrics across diverse tool-use scenarios. 
End-to-end task success rate measures whether a full multi-step agent loop completes its goal, providing a coarser but more ecologically valid signal that aggregates individual tool-call quality into an outcome metric.

Iterative self-correction is another dLLM-specific evaluation dimension. 
The ability to re-mask and regenerate low-confidence or erroneous spans within a single generation is structurally unavailable to left-to-right AR models without explicit rollback. 
\citet{ye2025beyondautore} demonstrate that discrete diffusion's bidirectional context provides an advantage on tasks requiring global constraint satisfaction, where early positions can be revised once later context clarifies the requirements. 
Two metrics characterize this dimension: correction frequency, defined as the proportion of denoising steps that revise previously generated tokens, and pass@$k$ as a function of denoising steps, which traces how generation quality improves with additional compute. 
However, standardized protocols for reporting these quantities across dLLMs are currently lacking, making it difficult to attribute improvements in end-to-end task success to better initial generation or better self-correction. 
Establishing such protocols is an important open evaluation gap as the generate-verify-correct loop becomes the central operating mode for diffusion-based agents.

\subsection{Speed, Cost, and Efficiency Metrics} \label{subsec:eval_speed}

%
Speed and efficiency are central evaluation dimensions for discrete diffusion models because their main practical promise is to reduce the sequential bottleneck of autoregressive generation.
However, efficiency comparisons are often difficult to interpret: different papers report different numbers of denoising steps, different samplers, different hardware, different batch sizes, and different definitions of generated tokens. 
A diffusion model may use fewer serial decoding steps than an autoregressive model, but each step usually processes the full sequence with bidirectional attention; conversely, an autoregressive model may require many more decoding steps but benefits from KV caching and streaming. 
Therefore, speed evaluation should distinguish algorithmic step complexity, actual model forward passes, wall-clock latency, throughput, memory cost, and end-to-end system cost.

\paragraph{Standardizing speed.}
The first quantity to report is the number of \textbf{denoising steps}. 
For a masked diffusion language model, this is the number of refinement iterations used to transform an initially corrupted sequence into a final output. 
For discrete-time samplers, this may correspond directly to the reverse diffusion steps; for continuous-time models, it corresponds to the chosen discretization of the reverse process; for confidence-based or blockwise samplers, it may refer to the number of remasking and update rounds~\citep{sahoo2024simpleeffect,shi2024simplified,lou2024discrete,arriola2025blockdiffusi,arriola2025remasking}. 
Step count is useful because it captures the serial depth of generation, but it is not sufficient by itself: two samplers with the same number of steps may require different numbers of neural forward passes, different sequence lengths per pass, or different auxiliary computations.

A second metric is the number of \textbf{model forward passes}. 
Some samplers perform one denoiser call per step, while others require multiple calls for guidance, verification, rejection, speculative decoding, or self-correction. 
Classifier-free guidance, for instance, may require both conditional and unconditional predictions unless implemented with batching or approximation. 
Similarly, speculative or hybrid AR-diffusion methods may draft with one model and verify with another, so their cost cannot be summarized by diffusion step count alone~\citep{arriola2025speculative,unknown2025drafdiffveri,unknown2025diffunlodiff}. 
Reporting forward-pass count makes the computational budget more comparable across samplers and model families.

A third metric is \textbf{wall-clock latency}, including time-to-first-output and time-to-final-output. 
Autoregressive models are naturally streamable: they can emit tokens as soon as decoding begins, even if the full sequence takes many steps to complete. 
Diffusion models typically require several denoising rounds before producing a coherent output, so their time-to-first-readable-output may be worse even when final-output latency is competitive.
For interactive systems, this distinction matters. 
Papers should therefore report both end-to-end generation latency and, when relevant, the earliest point at which a usable partial output can be returned.

A fourth metric is \textbf{throughput}, measured as generated tokens per second or sequences per second. 
Throughput is strongly affected by batch size and sequence length. 
Diffusion models benefit from parallel updates across positions and from batching many sequences, making them attractive for offline or high-throughput settings. 
However, the per-step cost of bidirectional attention over the full sequence can be high, especially for long contexts. 
AR models, by contrast, have a sequential dependency over output length but exploit KV caching to reduce the marginal cost of each new token. 
Thus, throughput comparisons should specify whether the setting is single-query interactive decoding, batched offline generation, or long-context generation.

Hardware and implementation details should also be reported. 
Efficiency depends on GPU/TPU type, memory bandwidth, kernel implementation, precision, batch size, maximum sequence length, attention implementation, and whether caching or compilation is used. 
Recent work has begun to design diffusion-specific acceleration methods, including adaptive caching, local determinism propagation, token-wise feature caching, quantization, consistency distillation, adaptive block sizes, and one-step or few-step generation~\citep{unknown2025dkvccachdiff,unknown2025dllmaccediff,unknown2025accediffllm,unknown2025accedifftran,unknown2025dllmquandiff,unknown2025cdlmconsdiff,unknown2025cd4lconsdist,unknown2025adabsemadiff,unknown2025dlmodifflang,unknown2025d3llultrdiff}. 
These methods make it even more important to separate model-intrinsic efficiency from system-level engineering. 
At minimum, papers should report the model size, hardware, batch size, precision, context length, output length, number of denoising steps, number of forward passes, and whether caching or acceleration tricks are enabled.

\paragraph{Frontier plots.}
We note up front that this subsection advocates a reporting methodology rather than carrying out a new empirical comparison: we do not compile a normalized cross-paper results table, because the cited results use heterogeneous hardware, tokenizers, and budgets that we cannot re-run or fairly normalize here. 
The reporting recommendations below are therefore authors' proposals aimed at future work.
Because diffusion sampling exposes an explicit quality-speed trade-off, single-point efficiency numbers are often misleading. 
A model sampled for $4$ denoising steps may be fast but low quality; the same model sampled for $64$ steps may be much slower but substantially better. 
Similarly, confidence-based remasking, adaptive schedules, guidance strength, and block size can all shift the trade-off between latency and quality. 
Therefore, we recommend reporting \textbf{quality-latency frontier plots} rather than only a single speed number.

A useful frontier plot places a task-level quality metric on the vertical axis and latency or forward-pass budget on the horizontal axis. 
For language generation, quality may be measured by task-specific metrics such as ROUGE, BLEU, BERTScore, exact match, pass@$k$ for code, constraint satisfaction rate, human preference, or LLM-judge score. 
The horizontal axis may be wall-clock latency, number of forward passes, denoising steps, or FLOPs. 
Reporting multiple x-axes can be helpful: denoising steps reveal algorithmic serial depth, while wall-clock latency reflects implementation reality. For deployment-oriented papers, tokens per second and memory footprint should also be included.

Frontier plots are especially important when comparing diffusion models with AR baselines. 
AR models define a natural quality-latency curve through decoding parameters such as temperature, nucleus sampling, beam size, speculative decoding, and draft-model configuration. 
Diffusion models define a corresponding curve through denoising steps, remasking schedules, block sizes, guidance scales, and early stopping. 
Comparing one diffusion setting against one AR setting can obscure the true trade-off. 
A stronger comparison is to plot the best achievable quality at each latency budget for each family. 
This makes clear whether diffusion dominates AR in a particular regime, such as batched generation or fixed-length infilling, or whether AR remains preferable, such as low-latency streaming chat.

Frontier reporting also helps diagnose where efficiency gains come from. 
If a method improves latency without changing the quality-latency frontier, it is primarily an implementation or constant-factor improvement. 
If it shifts the frontier upward, it improves sample quality at a fixed budget. 
If it shifts the frontier leftward, it achieves the same quality with fewer steps or less wall-clock time. 
This distinction is important for evaluating acceleration methods such as distillation, caching, adaptive unmasking, or speculative diffusion decoding~\citep{arriola2025speculative,unknown2025accesampmask,unknown2025effiautodiff,unknown2025dimodistmask,unknown2025cdlmconsdiff}. 
In addition, frontier plots should include error bars or confidence intervals when results depend on stochastic sampling, because diffusion samplers can exhibit variance across random seeds, masking orders, and guidance settings.

\paragraph{End-to-end agent latency.}
For tool-using systems, planning systems, and agentic applications, model decoding latency is only one component of the total cost. 
A diffusion model may be used to propose a plan, generate a set of candidate actions, refine intermediate reasoning states, or produce structured tool calls. 
In such settings, the full system cost depends on the number of denoising steps, the number of generated candidates, the number of tool calls, and the number of verification or repair rounds.
Evaluating only raw token-generation speed can underestimate the true latency experienced by users or downstream systems.

We recommend reporting the full planning-cycle cost:
\begin{equation}
    \text{Total cost} \approx
    N_{\text{denoise}} C_{\text{forward}}
    + N_{\text{tool}} C_{\text{tool}} + N_{\text{verify}} C_{\text{verify}},
\end{equation}
where $N_{\text{denoise}}$, $N_{\text{tool}}$, and $N_{\text{verify}}$ denote the numbers of denoising steps, tool calls, and verification rounds, respectively, and $C_{\text{forward}}$, $C_{\text{tool}}$, and $C_{\text{verify}}$ denote their corresponding costs.

For systems that sample multiple candidate trajectories, the cost should also include the number of parallel or sequential candidates. 
If candidates are generated in parallel, wall-clock latency may be lower than total compute cost; if they are generated sequentially, both latency and compute grow with the number of samples. 
Tool calls may dominate the total runtime when they involve web search, code execution, theorem proving, database queries, or external simulators. 
Verification can also be expensive when it uses another large model, a compiler, a unit-test suite, or a domain-specific validator.

Diffusion can be attractive in agentic settings because it can revise a whole plan or structured output rather than committing to actions one at a time. 
For example, a model can generate a full action sequence, evaluate global consistency, remask invalid steps, and refine the plan. 
This global revision may reduce the number of failed tool calls or repair loops, even if the model decoding itself is slower. 
Therefore, the right efficiency metric is not always tokens per second, but \textbf{successful task completion per unit cost}. 
For tool-using agents, useful metrics include total wall-clock time to task completion, number of model calls, number of tool calls, number of failed tool calls, number of verification failures, total GPU time, and monetary cost.

This end-to-end perspective is also important for constrained generation and code generation. 
A diffusion model may generate candidate programs or structured outputs through multiple denoising rounds, after which an external compiler, parser, unit-test suite, or constraint checker filters or repairs the output. 
In such systems, decoding speed should be reported together with pass rate, repair rate, and validation cost. 
A model that produces more valid outputs may reduce downstream verification cost even if its raw decoding is slower; conversely, a faster sampler that produces many invalid candidates may be inefficient end-to-end. 
Thus, for agentic and tool-augmented applications, efficiency evaluation should couple generation latency with downstream execution and verification outcomes.

In summary, efficiency evaluation for discrete diffusion should go beyond denoising step count. 
A complete report should include model forward passes, wall-clock latency, throughput, memory, hardware, batch size, sequence length, and acceleration settings. 
Because diffusion exposes a continuous quality-speed trade-off, quality-latency frontier plots are preferable to single-point comparisons. 
Finally, for agents and tool-using systems, evaluation should measure the full planning cycle, including denoising, candidate generation, tool execution, and verification. 
These metrics make it possible to determine when diffusion provides a genuine systems-level advantage over AR generation and when its apparent parallelism is offset by multi-step refinement or downstream costs.

\subsection{Editing, Infilling, and Conditional Fidelity} \label{subsec:eval_editing}
\paragraph{Edit benchmarks.} 
Editing is a task that requires models to produce outputs that are simultaneously high-quality, consistent with their context, and minimally disruptive to unmodified regions. 
Benchmarks for this task typically measure infilling consistency, minimal-change editing, and constraint adherence. 
Several benchmarks already exist across modalities.
HumanEval-Infilling~\citep{bavarian2022EffiTraiLang} evaluates infilling consistency for code, measuring pass@$k$ on span-masked Python functions where the model must generate a middle segment conditioned on both a left prefix and a right suffix.
SARI~\citep{xu2016OptiStatMach} benchmarks text editing by comparing system outputs against both the source sentence and multiple human references. 
It computes F1 scores over tokens that are correctly added, kept, or deleted, making it the standard metric for minimal-change rewriting tasks such as simplification and paraphrasing. 
For media, image and audio inpainting benchmarks assess reconstruction quality within masked regions via pixel- or token-level fidelity metrics, along with the boundary coherence between edited and preserved regions~\citep{sharma2024sketimaginpa, unknown2026tokeaudiinpa}.
For constraint adherence, COLLIE \citet{yao2023COLLSystCons} provides a compositional benchmark measuring the success rate of outputs satisfying rich lexical and structural constraints across word, sentence, and paragraph levels. 
These existing benchmarks were designed and validated primarily on AR models. 
They remain applicable in discrete diffusion settings to the extent that the model is evaluated on span-level quality: HumanEval-Infilling's bidirectional conditioning is in fact a natural fit for masked denoising, and SARI's decomposition of adds, keeps, and deletes remains informative. 
However, a critical evaluation gap is that none of these benchmarks explicitly penalize out-of-span changes, which is a failure mode specific to discrete diffusion models where the denoising process may revise tokens that lie outside the targeted region. 
Within the discrete diffusion literature, EdiText~\citep{lee2025editcontcoar} takes a step toward closing this gap by proposing a controllable coarse-to-fine text editing evaluation that tracks attribute adherence and out-of-region token preservation jointly. 
Establishing out-of-span preservation rate as a first-class reported quantity alongside infilling quality and constraint satisfaction is a necessary step toward reliable editing evaluation for discrete diffusion models.
Concretely, given an input $\bm{x}$, an edit region $S\subseteq\{1,\dots,L\}$, and an output $\hat{\bm{x}}$, we suggest reporting the out-of-span preservation rate
\begin{equation}\label{eq:oosp}
\mathrm{OOSP} \;=\; \frac{1}{|\,\bar{S}\,|}\sum_{i\in \bar{S}} \mathbb{1}[\hat{x}_i = x_i], \qquad \bar{S}=\{1,\dots,L\}\setminus S,
\end{equation}
i.e.\ the fraction of tokens \emph{outside} the intended edit region that are left unchanged ($\mathrm{OOSP}=1$ means a perfectly localized edit). This is straightforward to compute, complements infilling quality on $S$, and exposes the diffusion-specific failure mode in which denoising rewrites tokens it was not asked to touch.

\paragraph{Conditional alignment.} 
Conditional generation should be evaluated in two quantities: conditioning fidelity, which measures how closely the output follows the signal, and unconditional quality, which measures whether the output remains coherent independent of that signal. 
These two quantities should be tracked simultaneously and neither alone is sufficient. 
Across modalities, standard practice has converged on pairing one metric from each axis.
For text, an externally trained attribute classifier provides fidelity scores on dimensions such as sentiment or topic, while perplexity or fluency serves as the quality control. 
For media, CLIP score~\citep{radford2021LearTranVisu} measures semantic alignment between a text prompt and a generated image or video, and is paired with FID~\citep{heusel2017GansTraiTwo} to jointly capture fidelity and distributional quality. 
\citet{lee2023textsampfram} adopt this dual-reporting protocol in the context of text-conditioned masked generative image models. 
For structured domains such as molecules, property satisfaction rates measures the conditioning fidelity, and validity and diversity measures global quality. 
Since all of these metrics evaluate the final output independently of how it was generated, they transfer to discrete diffusion settings without modification. 
However, the gap emerges not in what is measured, but in what is left unreported. 
In continuous diffusion, classifier-free guidance~\citep{ho2021ClasFreeDiff} interpolates scores linearly at inference. 
In masked discrete diffusion, the analogous operation acts on categorical logits and is sensitive when the denoising chain guidance is applied.
\citet{rojas2026imprclasguid} show that high guidance applied early in the denoising chain, when most tokens are still masked, causes excessive unmasking and degrades sample quality, whereas late-stage guidance improves fidelity. 
This trade-off is invisible to metrics that evaluate only the finished output. 
\citet{koh2025condmaskdisc} further show that fidelity must be measured by a held-out attribute classifier rather than the model used for conditioning, to avoid circular evaluation. 
For discrete diffusion models, the guidance weight used should therefore be reported as a first-class experimental variable alongside fidelity and quality scores, rather than treated as an implementation detail.

%% file: sections/11-discussion.tex
\section{Discussion: Optimization, Controllability, and Trustworthiness} \label{sec:discussion}

%
This section steps back from individual methods to synthesize cross-cutting themes.
Section~\ref{subsec:disc_refinement} draws together the recurring contrast between diffusion as global, iterative refinement and autoregressive generation, and reframes the reverse process as a form of discrete optimization. 
Section~\ref{subsec:disc_trust} then examines trustworthiness, distinguishing established findings from open hypotheses.

\subsection{Diffusion as Global Discrete Refinement vs.\ Autoregressive Generation} \label{subsec:disc_refinement}
Across the formulations, applications, and inference algorithms surveyed above, a single conceptual thread recurs: discrete diffusion generates by \emph{iterative global refinement} rather than by left-to-right commitment. 
This subsection draws that thread together and contrasts it with autoregressive (AR) generation, then reframes the reverse process as a form of discrete optimization.

\paragraph{Planning interpretation.} 
Each denoising step conditions on the entire current sequence and updates many positions at once.
The trajectory from a fully corrupted state to a clean sample can therefore be read as a coarse-to-fine planning process: early steps fix global structure under high uncertainty, while later steps resolve local detail once the surrounding context has stabilized. 
This stands in contrast to AR decoding, where the factorization $p_\theta(\bm{x})=\prod_i p_\theta(x_i\mid x_{<i})$ commits to each token before the next is considered. 
In agentic and reasoning settings, this difference is consequential: a diffusion planner can revise an earlier part of a plan or reasoning trace after inspecting a later part, whereas AR decoding must rely on explicit revision machinery (self-correction loops, external verification, or re-decoding) to change content it has already emitted~\citep{ye2025beyondautore, deng2025thinkwhile}.

\paragraph{Error correction across refinement steps.} 
Because positions are revisited over multiple steps, an early mistake need not propagate irreversibly the way it can under AR exposure bias. 
Remasking and confidence-based decoding (Section~\ref{sec:inference}) make this explicit: tokens committed under high uncertainty can be returned to the masked state and re-predicted once neighboring tokens supply more context. 
The practical upside is improved global coherence and constraint satisfaction; the practical caveat is that this benefit is realized only when the sampler actually budgets steps for revision rather than greedily freezing high-confidence tokens. 
The empirical picture is therefore nuanced, and we frame revision as a \emph{potential} advantage that depends on the decoding policy rather than a guaranteed property of the model class.

\paragraph{Optimization lens.} 
The following framing is an interpretive perspective we propose rather than an established result.
The reverse chain can be viewed as a stochastic optimizer over the discrete configuration space $\mathcal{V}^L$: it begins from a high-entropy prior and progressively concentrates probability mass on high-likelihood (or, under guidance, high-reward) regions. 
This perspective connects discrete diffusion to combinatorial search and offline black-box optimization, where the denoiser plays the role of a learned proposal distribution and guidance plays the role of an objective. 
Compared with classical heuristics such as beam search or evolutionary methods, diffusion offers learned, globally informed proposals and a natural mechanism for soft constraint handling via logit shaping; compared with exact combinatorial solvers, it trades optimality guarantees for amortized speed and the ability to exploit learned structure. 
We expand on these connections, including offline RL and combinatorial optimization, in Section~\ref{sec:applications}.

\paragraph{When AR or hybrid approaches win.} 
Diffusion is not uniformly preferable. 
AR models retain decisive advantages for token-by-token streaming, incremental user interaction, and settings where mature KV-cache infrastructure makes per-token latency negligible. 
Many of the most effective recent systems are therefore \emph{hybrids}: AR planning followed by diffusion refinement, diffusion used for editing or infilling on top of an AR backbone, block/semi-autoregressive schemes that interpolate between the two regimes, and two-stage pipelines that use diffusion for global drafting and AR for fluent local decoding. 
The choice between paradigms is best framed not as a winner-take-all question but as a matching of generation mechanism to task structure (latency profile, need for revision, constraint density, and streaming requirements).

\subsection{Trustworthiness: Hallucination, Uncertainty, and Safety Constraints} \label{subsec:disc_trust}
The trustworthiness properties of discrete diffusion differ from those of AR models in ways that are only beginning to be characterized. 
We summarize three threads, hallucination, uncertainty, and safety, and are careful to distinguish established findings from open hypotheses.

\paragraph{Hallucination.} 
It is sometimes hypothesized that bidirectional context and multi-step revision should reduce hallucination relative to AR generation, since the model can reconcile a claim against the full surrounding context before committing. 
This is plausible but not yet established: the evidence base is thin, confounded by differences in scale and training data, and complicated by the fact that parallel decoding can introduce its own failure mode in which jointly sampled tokens are individually plausible but mutually inconsistent. 
We therefore treat reduced hallucination as a research question requiring controlled, scale-matched comparisons rather than as a settled advantage.

\paragraph{Uncertainty-aware generation.} 
A more concrete opportunity is that discrete diffusion exposes per-token, per-step confidence signals as a native byproduct of decoding. 
These signals, already used to drive confidence-based remasking, can in principle support uncertainty-aware behaviors: abstaining on low-confidence spans, requesting clarification, or triggering retrieval or external verification before committing. 
Turning these raw confidences into calibrated, actionable uncertainty estimates is an open and practically valuable direction.

\paragraph{Safety constraints and adversarial robustness.} 
Guidance, constrained decoding, and policy shaping during denoising provide levers for steering generation toward safe outputs, and the ability to enforce hard constraints (valid schemas, grammars, valence rules) at each step is a genuine strength for structured domains. 
At the same time, a growing body of work shows that the same mechanisms that make diffusion flexible also create new attack surfaces and defense considerations: jailbreak and red-teaming studies targeting diffusion language models~\citep{nie2025jailbreaking, unknown2026textredtlarg, unknown2026devibehimask}, analyses of intrinsic safety and the role of the masking process~\citep{unknown2026diffintrsafe, unknown2025vulndefeunde, unknown2026towasafediff}, membership-inference and privacy analyses~\citep{unknown2026membinfeatta, unknown2025inheprivprop}, and adversarial prompting specific to the diffusion setting~\citep{unknown2025diffdiffprom}. 
A recurring theme is that safety properties cannot be inherited wholesale from the AR literature: bidirectional, multi-step generation changes both where interventions can be applied and where vulnerabilities arise. 
Establishing whether safety is fundamentally easier or harder to enforce in discrete diffusion than in AR models remains open.

%% file: sections/12-future-directions.tex
\section{Future Directions} \label{sec:future}

%
This section outlines open problems that we view as both consequential and tractable. 
Section~\ref{subsec:fut_scaling} concerns scaling laws and benchmark standardization.
Section~\ref{sec:kv_cache_future} discusses closing the gap with autoregressive models in in-context learning, caching, and streaming. 
Section~\ref{subsec:fut_unified} considers unified token spaces, and Section~\ref{subsec:fut_theory} surveys theoretical questions of expressivity, convergence, and identifiability.

\subsection{Scaling Laws, Compute Trade-offs, and Benchmark Standardization} \label{subsec:fut_scaling}
Despite rapid progress, the scaling behavior of discrete diffusion remains far less understood than that of AR models. 
Several first-order questions remain open: whether diffusion language models exhibit scaling laws comparable to those of AR models in the large-data limit, how the number of denoising steps trades off against parameter count and training tokens, and whether the reported compute-optimal frontier --- favoring  smaller models trained longer --- holds across domains and scales. 
Answering these requires standardized ablations that vary model size, step budget, noise schedule, and tokenization independently, rather than the single-point comparisons that dominate the current literature.

A closely related obstacle is the absence of shared evaluation standards. 
Because diffusion decoding exposes a quality-compute frontier rather than a single operating point, single-number comparisons are easy to misinterpret (see Section~\ref{sec:evaluation}). 
We see benchmark standardization as a prerequisite to, rather than a complement of, settling the scaling questions above: a minimum reporting set --- including the number of steps, wall-clock latency, hardware, decoding algorithm, batch and prompt lengths, and tokenization details --- together with reproducible suites spanning text, graphs, and biological sequences would let the field compare methods on a common footing and transform scaling claims into testable hypotheses.

\subsection{Closing the Gap with AR: In-Context Learning, Caching, and Streaming}
\label{sec:kv_cache_future}
Three capabilities that AR models provide almost for free remain difficult for discrete diffusion, and closing these gaps is among the most consequential open problems.

\paragraph{In-context learning.} 
Few-shot in-context learning appears to be weaker in current diffusion language models than in comparable AR models. 
The reasons for this gap remain poorly understood.
Possible explanations include a mismatch between the denoising objective and the next-token prediction that ICL appears to rely on, interference between the prompt and the iteratively refined response under bidirectional attention, and insufficient scale or instruction tuning. 
Promising directions include multi-pass prompting, retrieval-augmented denoising, step-wise conditioning schemes that protect the prompt from corruption, and hybrid AR-prompted diffusion decoding.

\paragraph{Caching and streaming.} 
As discussed in Section~\ref{subsec:inf_accel}, bidirectional attention and whole-canvas updates invalidate the append-only KV cache that underpins efficient AR serving, and they also complicate streaming output, since no token is final until the reverse process completes. 
Recent work on block-wise and delayed KV reuse, incremental unmasking caches, speculative refinement, and confidence-gated freezing shows that partial reuse is achievable, but a clean analogue of the AR KV cach, one that delivers comparable per-token serving efficiency without sacrificing the revision flexibility that motivates diffusion, has not yet emerged. 
Architectural changes that enable partial reuse while preserving bidirectional refinement, and decoding schemes that emit committed prefixes incrementally, are both worth pursuing.

\subsection{Unified Token Spaces and Cross-Modal/Scientific Integration} \label{subsec:fut_unified}
A recurring theme of this survey is that the discrete state space is a first-class design axis. 
A natural frontier is therefore the design of \emph{unified} token spaces: shared discrete vocabularies that span modalities (text, image, audio, video tokens) versus modality-specific codebooks coordinated through a common diffusion process. 
Unified spaces promise simpler architectures and cross-modal transfer, but raise hard questions about how to balance vocabulary granularity, codebook topology, and per-modality corruption schedules within a single model. 
In scientific domains, an analogous opportunity is the joint modeling of sequence and structure tokens --- for example, coupling amino-acid sequence with structural descriptors, or molecular graphs with 3D conformer tokens.
Another promising direction is to incorporate physics- and chemistry-based constraints directly into the corruption and reverse processes rather than enforcing them only at sampling time.

\subsection{Theory: Expressivity, Convergence, and Identifiability} \label{subsec:fut_theory}
The theoretical foundations of discrete diffusion are advancing but remain incomplete. 
Regarding convergence, recent analyses establish discrete-time and uniformization-based guarantees and error bounds for score-based and flow-matching discrete samplers~\citep{chen2025convergence, unknown2025convanaldisc, unknown2025absoconvprov, unknown2025statsizeinde, unknown2025erroanaldisc, campbell2024theoretical}, but a unified account of how schedule, transition design, and parameterization jointly govern sampling error is still missing. 
Regarding expressivity, an important question is what classes of distributions factorized reverse models can represent, and how the gap to fully joint models scales with sequence length and step count; results on the parallel-decoding and information-theoretic limits of diffusion language models~\citep{unknown2026difflangmodea, unknown2026infoestidisc, unknown2025forgbitit, unknown2025overdimefact} begin to address this. 
Regarding identifiability, the bias introduced by approximate objectives and the conditions under which different parameterizations recover the same reverse process deserve systematic study~\citep{unknown2025discdiffmode, unknown2025openprobhypo}. 
Beyond their intrinsic interest, such results would have practical payoff: a sharper theory would tell us how to choose schedules, design transition matrices, and anticipate failure modes rather than discovering them empirically.

%% file: sections/13-conclusion.tex
\section{Conclusion} \label{sec:conclusion}
%
This survey has presented a unified treatment of discrete diffusion models, organized around the thesis that the construction of the discrete state space, namely tokenization, is a first-class design axis rather than merely a preprocessing detail. 
We reviewed the major formulation families under a common four-component structure (corruption operator, denoiser parameterization, training objective, and sampler), surveyed training objectives and inference algorithms, examined scaling and systems considerations, and mapped applications spanning text and code, tokenized multimodal generation, proteins, genomics, molecules, and graphs, planning and agents, and tabular data.

The central message is that the state-space design, the corruption process, the learning objective, and the sampling algorithm form an inseparable system: a choice made for one component constrains the others, and the most effective designs are those in which these choices are co-adapted to the structure of the data and the demands of the task. 
Where discrete diffusion most clearly shines is in parallel refinement, infilling and editing, and constraint-aware generation across domains, capabilities that follow directly from its bidirectional, iterative nature and that are awkward to obtain from purely left-to-right models.

Looking ahead, we expect autoregressive and diffusion paradigms to coexist rather than one displacing the other, with hybrid systems (AR planning composed with diffusion refinement, or diffusion editing layered atop AR backbones) likely to dominate practical deployments. 
The open problems we have highlighted, from scaling laws and in-context learning to caching, unified token spaces, and theoretical foundations, are concrete and tractable, and progress on them will determine how far the paradigm can be pushed. 
We hope that the unified perspective and cross-domain map offered here will help researchers navigate this rapidly evolving field and establish discrete diffusion as a powerful framework for controllable discrete generation and structured discovery.

%% file: sections/14-broader-impact.tex
\section*{Broader Impact} \label{sec:broader_impact}

%
This survey covers generative modeling techniques that, while not introducing new capabilities themselves, map and make more accessible a fast-moving design space with real dual-use and deployment considerations. 
We summarize the principal concerns by domain and point to mitigations discussed in the technical sections, without claiming to resolve them.

\paragraph{Dual use in scientific generation.} 
The protein, genomics, and molecule/graph applications (Sections~\ref{subsec:app_protein}-\ref{subsec:app_molecule}) lower the barrier to designing biological and chemical sequences with targeted properties. 
The same conditioning and guidance machinery that steers a model toward a therapeutic peptide or a high-binding ligand can in principle be redirected toward harmful targets. 
The discrete, constraint-aware nature of these methods is relevant here in both directions: hard validity and property constraints (Section~\ref{subsec:inf_guidance}) can be used to \emph{exclude} hazardous regions of sequence space as readily as to include desirable ones. 
We encourage practitioners releasing scientific generators to pair them with property filters, access controls appropriate to the risk, and documentation of intended use, consistent with established responsible-research-in-the-life-sciences norms; we do not provide operational detail on any harmful application.

\paragraph{Safety of diffusion language models.}
Section~\ref{subsec:disc_trust} reviews evidence that the bidirectional, multi-step, any-order nature of discrete diffusion changes the safety surface relative to autoregressive models: arbitrary-order decoding can enlarge the jailbreak attack surface, and watermarking and membership-inference defenses calibrated for left-to-right generation do not transfer unchanged. 
Translating those findings into practice, we suggest that deployment-time evaluation of dLLMs report (i) jailbreak and adversarial-prompting results under diffusion-specific, order-varying attacks rather than only AR-style attacks; (ii) membership-inference and privacy tests appropriate to the training regime; and (iii) whether any watermarking scheme used is order-agnostic. 
Whether safety is ultimately easier or harder to enforce in discrete diffusion than in AR models remains, as noted in Section~\ref{subsec:disc_trust}, an open question; we flag it as a priority rather than a solved problem.

\paragraph{Synthetic-data privacy.} 
The tabular generation and imputation setting (Section~\ref{subsec:app_tabular}) is explicitly motivated by privacy-preserving data sharing, yet the same fidelity that makes synthetic tables useful can leak information about real records. 
As discussed there, fidelity, privacy, and downstream utility are in tension, and optimizing fidelity alone can degrade privacy. 
We recommend that work in this area report membership-inference or distance-to-closest-record analyses alongside fidelity and utility, rather than a single aggregate score, so that the privacy cost of a given fidelity level is visible.

\paragraph{Accessibility and concentration of capability.}
Finally, by consolidating scattered techniques into a single design reference, this survey is intended to broaden access to discrete diffusion methods. 
We view wider, better-documented access, together with the reporting and evaluation standards proposed throughout, as net positive for reproducibility and scrutiny, while recognizing that the same accessibility applies to the dual-use concerns above. 
We have tried to keep the level of detail at the level of design principles and citations to published work, and to avoid step-by-step guidance toward misuse.